\documentclass{article}
\usepackage{enumitem}

\PassOptionsToPackage{numbers,compress}{natbib}
\usepackage[preprint]{neurips_2026}

\usepackage[utf8]{inputenc}
\usepackage[T1]{fontenc}

\usepackage{pifont}
\usepackage[table]{xcolor}
\usepackage{hyperref}
\hypersetup{
    colorlinks=true,
    linkcolor=blue,
    citecolor=blue,
    urlcolor=blue
}
\newcommand{\bluelink}[2]{\hyperref[#1]{#2}}

\usepackage{url}
\usepackage{amsmath}
\usepackage{amssymb}
\usepackage{amsfonts}
\usepackage{amsthm}
\usepackage{nicefrac}

\usepackage{graphicx}
\usepackage{array}
\usepackage{float}
\usepackage{booktabs}
\usepackage{makecell}
\usepackage{tabularx}
\usepackage{caption}
\usepackage{subcaption}
\usepackage{colortbl}

\usepackage{algorithm}
\usepackage{algpseudocode}

\usepackage{tikz}
\usetikzlibrary{positioning,arrows.meta}

\usepackage{microtype}
\usepackage{pifont}
\usepackage{soul}
\usepackage{blindtext}
\usepackage{etoc}

\sethlcolor{blue!15}

\definecolor{myblue}{RGB}{0,70,190}
\definecolor{lightrow}{RGB}{236,240,248}
\definecolor{bestrow}{RGB}{210,225,248}
\definecolor{tableblue}{RGB}{220,235,250}
\definecolor{headerblue}{RGB}{180,210,240}
\definecolor{best}{RGB}{220,255,220}
\definecolor{alggray}{RGB}{235,235,235}

\newcommand{\cmark}{\textcolor{green!60!black}{\ding{51}}}
\newcommand{\xmark}{\textcolor{red!75!black}{\ding{55}}}
\newcommand{\pmark}{\textcolor{orange!85!black}{\textbf{P}}}
\newcommand{\contrib}[1]{\textcolor{myblue}{\textbf{#1}}}
\providecommand{\best}[1]{\textbf{#1}}
\providecommand{\Delta}{\mathit{\Delta}}
\providecommand{\uparrow}{\textuparrow}
\providecommand{\downarrow}{\textdownarrow}

\title{KathaTrace: Diagnosing Semantic Trajectory Collapse in Generated Visual Narratives}

\author{%
  Jamuna S.~Murthy \\
  Ramaiah Institute of Technology \\
  \And
  Amin Karimi Monsefi \\
  The Ohio State University \\
  \And
  Rajiv Ramnath \\
  The Ohio State University \\
}

\begin{document}

\maketitle

\begin{abstract}
Visual narratives are central to storyboards, comics, children's media and film previsualization, where viewers understand stories from images alone. Recent 
generators like StoryDiffusion produce coherent sequences, but visual coherence does 
not guarantee that source-story transition meaning remains recoverable. Existing 
benchmarks assess visual quality, content faithfulness, and scene coherence, but miss 
a critical failure mode: storyboards where scenes appear visually coherent while the 
semantic link between scenes disappears. We introduce \textbf{KathaTrace}, a generator-agnostic protocol for diagnosing 
\emph{semantic trajectory collapse}—loss of transition meaning needed to understand 
how one scene follows another. KathaTrace evaluates transitions under three evidence 
conditions (text-only, image-only, text+image) and filters ambiguous items. We contribute \textbf{KathaBench-25K}, with 5,000 narratives from classical collections (Aesop, Panchatantra, Kathasaritasagara), 20,000 transitions, and 28,712 recoverability questions. We define \textbf{Semantic Trajectory Gap (STG)} as text-only minus image-only recoverability, measuring transition meaning lost during visualization. Human validation yields Fleiss' $\kappa=0.845$. Experiments across state-of-the-art generators show  substantial STG (23.5±1.3). \textbf{Semantic Compass}, an actionability probe, uses  KathaTrace signals for post-generation repair, improving storyboard selection.
\url{https://jamunasmurthy.github.io/kathatrace/}
\end{abstract}

\vspace{-6pt}

\section{Introduction}

Visual narratives promise compact storytelling, but visual clarity alone is not enough. A generated storyboard may preserve characters, objects, and plausible scenes while losing the transition meaning that explains why one scene follows another. Recent story visualization systems improve visual memory, context modeling, identity preservation, layout control, and long-range consistency~\cite{maharana2022storydall,rahman2023make,ahn2023story,feng2023improved,liu2024intelligent,tao2024storyimager,zhou2024storydiffusion,zheng2025contextualstory,shen2025storygpt,he2025dreamstory,dong2026vista,liu2025one,singh2025storybooth,ma2025lay2story,dinkevich2025story2board,sarkar2026redistory,sreenivas2026attristory}, but these improvements do not ensure that source-story transition meaning survives generation.

\begin{figure*}[!t]
\centering
\includegraphics[width=\linewidth]{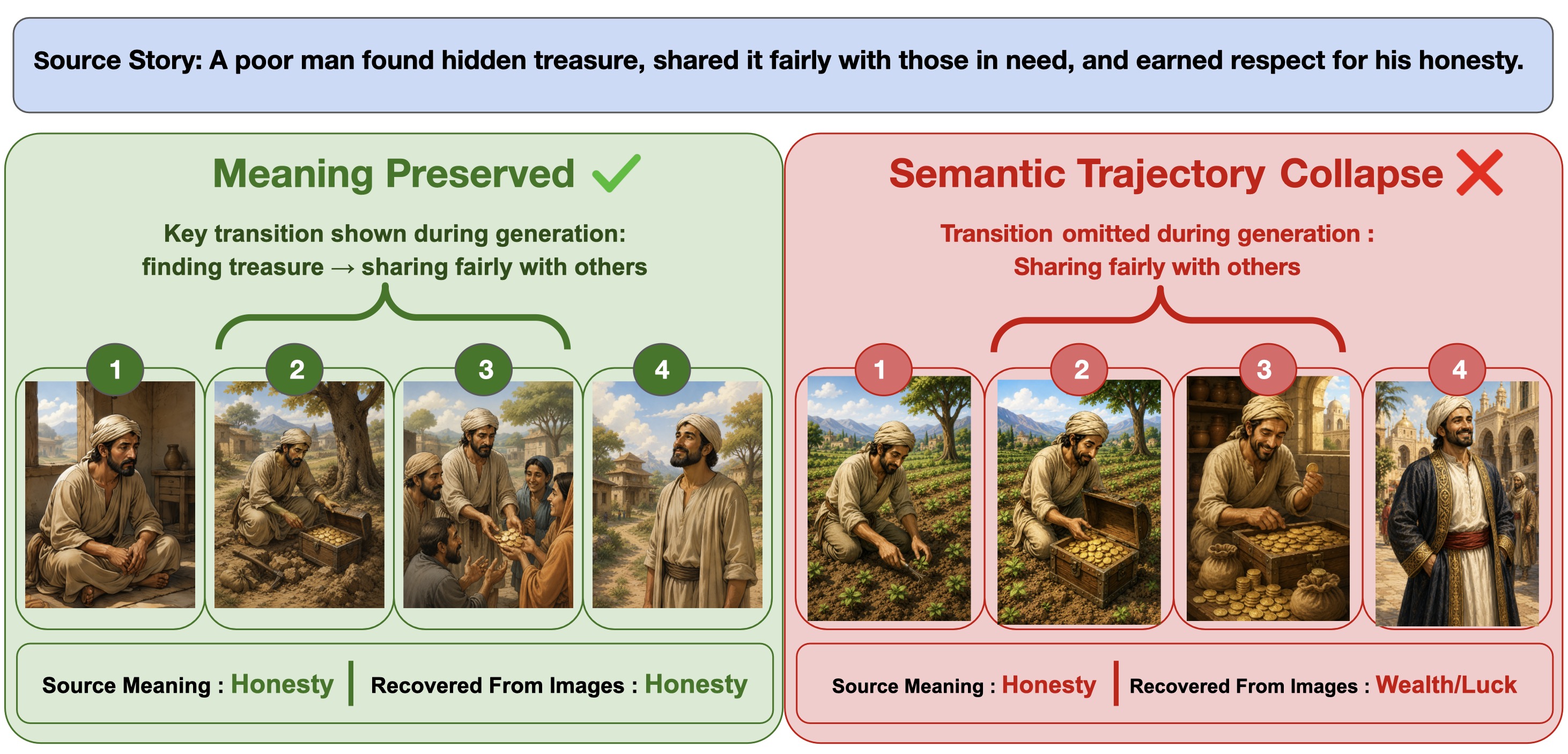}
\caption{
\textbf{Semantic trajectory collapse.}
A generated storyboard may preserve characters, setting, and local coherence while changing the transition meaning recoverable from images alone. KathaTrace measures this loss using image-only recoverability and the \emph{Semantic Trajectory Gap}.
}
\label{fig:semantic_trajectory_collapse}

\end{figure*}

Fig.~\ref{fig:semantic_trajectory_collapse} illustrates the failure. In the left panel, the poor man finds treasure, shares it fairly, and earns respect, so the recovered meaning remains \emph{honesty}. In the right panel, the storyboard keeps the man, treasure, and final respect scene but omits the sharing event; the sequence still looks coherent, yet the meaning shifts to wealth or luck. We call this transition-level loss \emph{semantic trajectory collapse}.

\emph{Semantic trajectory collapse} is the transition-level narrative information loss studied in this work. It occurs when source-supported transition meaning, such as sharing fairly before earning respect, is no longer recoverable from generated images alone; we diagnose this loss using the \emph{Semantic Trajectory Gap} (STG). A \emph{semantic trajectory} is the ordered chain of scene-to-scene transitions connecting actions, causes, emotions, consequences, temporal order, and moral-semantic targets. Here, a moral-semantic target is a source-defined recoverability target, such as honesty earning trust or greed causing loss, not a claim of objective moral truth as illustrated in Fig.~\ref{fig:semantic_trajectory_collapse}.

We define the \emph{Semantic Trajectory Gap} (STG) as text-only recoverability minus image-only recoverability. Image-only evaluation is not artificial: in storyboards, comics, wordless children's books, accessibility settings, low-literacy communication, and film previsualization, images are often the final medium and the source text may not be available to the viewer. If a transition is recoverable only when the source text is supplied again, then that meaning has not survived visualization.

Existing benchmarks and metrics evaluate coherence, discourse consistency, commonsense grounding, character continuity, visual logic, action-state consistency, and prompt faithfulness~\cite{bugliarello2023storybench,gao2025vinabench,zhuang2025vistorybench,ye2024openstory++,he2025dreamstory,meng2026logistory,hamilton2026narrabench,lin2026narratology}. Fine-grained text-to-image metrics such as TIFA, Davidsonian Scene Graphs, GenEval, GenAI-Bench, and VQAScore-style evaluation show that question-answering can provide interpretable image-text faithfulness signals~\cite{hu2023tifa,cho2024davidsonian,ghosh2023geneval,li2024genai,kamath2025geneval}. However, diagnosing semantic trajectory collapse requires a stricter set of protocol elements: source stories to define intended meaning, transition annotations to localize adjacent-scene targets, recoverability questions to test inference, contrastive variants to expose near-miss semantic drift, image-only evaluation to avoid source-side leakage, and paired text/image scoring to measure STG. Table~\ref{tab:kathatrace_overview} compares support for these diagnostic requirements. The comparison shows that prior benchmarks cover useful parts of story evaluation, such as coherence, visual logic, or prompt faithfulness, but none jointly provide source-story transition annotations, recoverability QA, contrastive variants, image-only testing, and paired text/image STG scoring.

We introduce \textbf{KathaTrace}, a generator-agnostic recoverability protocol for diagnosing semantic trajectory collapse in generated visual narratives. KathaTrace is multimodal in design but image-only in its primary diagnostic condition: text-only recoverability estimates source-side support, image-only recoverability tests whether the generated storyboard communicates transition meaning without source leakage, and text+image recoverability filters ambiguous or invalid items. We also construct \textbf{KathaBench-25K}, a transition-aware benchmark of 5,000 rights-compatible classical narratives from \emph{Kathasaritsagara}/\emph{The Ocean of Story}, \emph{Aesop's Fables}, and \emph{Panchatantra}, normalized into structured scenes, transition annotations, recoverability questions, accepted-answer sets, contrastive variants, evidence-isolated scoring, ambiguity filtering, and held-out splits.

Beyond the leaderboard, controls show that the recoverability gap is not explained by generic QA, renderer quality, or extra frames. In evaluator ablations, the strongest adapted DSG-style baseline reports $28.9$ STG and $0.58$ human correlation, while the full KathaTrace protocol reports $23.5$ STG and $0.71$, indicating stronger human alignment on causal gaps, hidden consequences, and moral-target drift. With FLUX fixed, the Gemma-ST planner reduces STG from $35.2$ to $24.0$, showing that transition-aware planning matters beyond rendering. Semantic Compass further reduces STG from $28.4$ to $21.4$ on held-out evaluation, while non-semantic extra frames yield only small gains.

\clearpage

\begin{table*}[!t]
\centering
\scriptsize
\renewcommand{\arraystretch}{0.92}
\setlength{\tabcolsep}{4.2pt}
\caption{
\textbf{Benchmark comparison for semantic trajectory diagnosis.}
Columns mark the minimal protocol elements needed to test whether source-story transition meaning remains recoverable from generated images alone; this does not assess overall benchmark quality.
}
\label{tab:kathatrace_overview}
\resizebox{\textwidth}{!}{%
\begin{tabular}{lcccccc}
\toprule
\textbf{Benchmark} &
\textbf{Story} &
\textbf{Trans.} &
\textbf{Recov. QA} &
\textbf{Contrast.} &
\textbf{Image-only} &
\textbf{STG} \\
\midrule
\rowcolor{gray!10}
StoryBench~\cite{bugliarello2023storybench}
& \cmark & \pmark & \xmark & \xmark & \xmark & \xmark \\
VinaBench~\cite{gao2025vinabench}
& \cmark & \pmark & \pmark & \xmark & \xmark & \xmark \\
\rowcolor{gray!10}
ViStoryBench~\cite{zhuang2025vistorybench}
& \cmark & \pmark & \pmark & \xmark & \xmark & \xmark \\
LogicTale / LogiStory~\cite{meng2026logistory}
& \cmark & \cmark & \pmark & \pmark & \pmark & \xmark \\
\rowcolor{gray!10}
OpenStory++ / Cohere-Bench~\cite{ye2024openstory++}
& \cmark & \pmark & \xmark & \xmark & \xmark & \xmark \\
DS-500 / DreamStory~\cite{he2025dreamstory}
& \cmark & \pmark & \xmark & \xmark & \xmark & \xmark \\
\midrule
\rowcolor{blue!15}
\textbf{KathaBench-25K}
& \textbf{\cmark} &
\textbf{\cmark} &
\textbf{\cmark} &
\textbf{\cmark} &
\textbf{\cmark} &
\textbf{\cmark} \\
\bottomrule
\end{tabular}%
}
\vspace{2pt}

\raggedright
\footnotesize
\textit{Story}: source narratives are included.
\textit{Trans.}: explicit adjacent-scene transition annotations.
\textit{Recov. QA}: recoverability questions for testing inferable transition meaning.
\textit{Contrast.}: semantic variants that expose near-miss transition drift.
\textit{Image-only}: evaluation without source text, prompts, captions, labels, or annotations.
\textit{STG}: paired text-side and image-only scoring for Semantic Trajectory Gap.
\cmark~= full support;
\pmark~= partial or proxy support;
\xmark~= absent.

\end{table*}

Our contributions are:
\begin{itemize}
\item We define \contrib{semantic trajectory collapse} as the failure where generated storyboards preserve visible content but lose the transition meaning needed to understand how one scene follows another.
\item We propose \contrib{KathaTrace}, a generator-agnostic image-only recoverability protocol for testing whether source-story transitions can be inferred from generated images alone.
\item We release \contrib{KathaBench-25K}, a transition-aware benchmark with structured scenes, transition annotations, accepted-answer sets, recoverability questions, contrastive variants, ambiguity filtering, and held-out splits.
\item We introduce \contrib{Semantic Trajectory Gap and actionability} to quantify text-to-image loss of recoverable narrative meaning, localize failures by transition type, and test whether these signals can guide reranking and bridge-scene repair.
\end{itemize}

\section{Related Work}

\paragraph{Story visualization, narrative benchmarks, and the dataset gap.}
Story visualization systems generate image sequences from narrative text using story continuation, visual memory, context modeling, bidirectional generation, diffusion rendering, multimodal history, identity preservation, layout control, and long-range consistency~\cite{maharana2022storydall,rahman2023make,ahn2023story,feng2023improved,liu2024intelligent,tao2024storyimager,zhou2024storydiffusion,zheng2025contextualstory,shen2025storygpt,he2025dreamstory,dong2026vista,yang2025seed,liu2025one,singh2025storybooth,ma2025lay2story,dinkevich2025story2board,sarkar2026redistory,sreenivas2026attristory,maostory}. These methods improve visual coherence, subject consistency, and generation control, but they do not directly test whether viewers can recover why one scene follows another from images alone. Multi-shot video and narrated-story systems add agents, memory, cinematic structure, and world consistency~\cite{hu2024storyagent,xu2025mm,shi2025animaker,zhang2025storymem,shi2026msvbench,zhang2026muss,elmoghany2026infinitystory,zhou2026videomemory}, yet visual continuity can still improve while causal, emotional, consequence, or moral-target meaning is omitted.

Existing benchmarks address related but different goals. As summarized in Table~\ref{tab:kathatrace_overview}, StoryBench, VinaBench, and ViStoryBench evaluate faithful and consistent visual narratives, but only partially support transition-localized recoverability and do not report paired STG~\cite{bugliarello2023storybench,gao2025vinabench,zhuang2025vistorybench}. OpenStory++ / Cohere-Bench and DS-500 / DreamStory support open-domain or consistent story generation, but do not isolate image-only recovery of adjacent-scene transition meaning~\cite{ye2024openstory++,he2025dreamstory}. LogicTale / LogiStory moves closer through visual logic and causal coherence, but lacks the full diagnostic chain of source transition annotations, fixed recoverability QA, contrastive variants, strict image-only testing, and paired text/image STG~\cite{meng2026logistory}. KathaBench-25K fills this gap by combining source meaning, transition targets, recoverability questions, contrastive drift, ambiguity filtering, and STG in one benchmark.

\paragraph{QA-based faithfulness evaluation and KathaTrace.}
QA-based evaluation provides interpretable signals for generated images. TIFA tests text-to-image faithfulness with generated QA pairs~\cite{hu2023tifa}; Davidsonian Scene Graphs evaluate atomic scene facts and dependencies~\cite{cho2024davidsonian}; GenEval and GenAI-Bench focus on object-level and compositional alignment~\cite{ghosh2023geneval,li2024genai}; GenEval 2 studies benchmark drift~\cite{kamath2025geneval}; and interleaved or open-ended benchmarks measure image-text structure, coherence, or human-aligned multimodal judgment~\cite{chen2025interleaved,chen2025comm,zhou2025opening}. These metrics provide useful QA-based diagnostics, but they do not define source-story transition targets and then test whether those targets remain recoverable from generated images without source text, prompts, captions, labels, or annotations.

KathaTrace keeps QA interpretability but changes the evidence design: each transition target is evaluated under text-only, image-only, and text+image conditions. Text-only scoring estimates whether the target is recoverable from the source story; image-only scoring tests whether the generated storyboard communicates the same target without source leakage; and text+image scoring filters items that remain ambiguous even with full evidence. This produces a paired estimate of text-to-image transition loss, localizes failures by dimension, and supports actionability analysis through Semantic Compass reranking and bridge-scene repair rather than only aggregate coherence.

\section{Method}
\label{sec:method}
\begin{figure*}[!t]
\centering
\includegraphics[width=\textwidth]{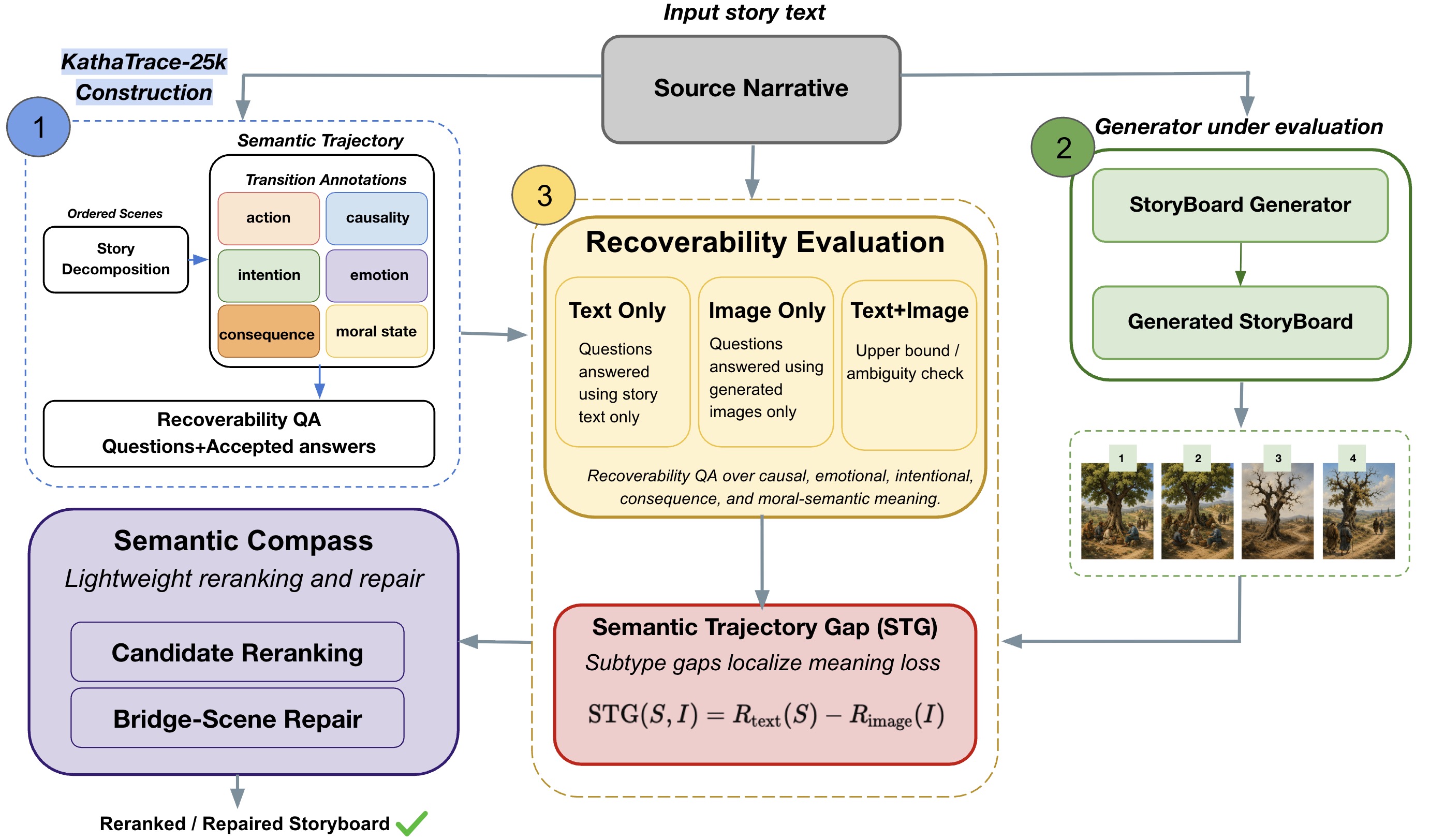}
\caption{
\textbf{KathaTrace framework.}
KathaTrace structures stories into scenes and transition-level recoverability questions, then tests them under text-only, image-only, and text+image evidence. It reports STG, dimension gaps, ambiguity rates, contrastive recoverability, and the optional Semantic Compass repair probe.
}
\label{fig:method_overview}
\end{figure*}
We introduce \textbf{KathaTrace}, a generator-agnostic protocol for testing whether source-story transition meaning remains recoverable after visual generation. Instead of scoring only visual plausibility, KathaTrace asks whether a viewer can infer why one scene follows another from images alone. Fig.~\ref{fig:method_overview} summarizes the framework. 
\textit{\textbf{See Appendix Secs. \hyperref[seca]{A}--\hyperref[secd]{D} for dataset construction, scoring, ambiguity filtering, judge aggregation, Semantic Compass, and metric details.}}

\subsection{Task Formulation}
\label{sec:task_formulation}

A source story is represented as an ordered scene sequence:
\begin{equation}
S=\{s_1,s_2,\ldots,s_T\}.
\label{eq:source_sequence}
\end{equation}
A storyboard generator $G$ produces ordered images:
\begin{equation}
\hat{X}=G(S)=\{\hat{x}_1,\hat{x}_2,\ldots,\hat{x}_T\}.
\label{eq:generated_storyboard}
\end{equation}
KathaTrace requires only three inputs: the source story $S$, the generated storyboard $\hat{X}$, and a fixed set of transition-level recoverability questions. It does not use generator internals, model weights, training data, hidden prompts, or intermediate plans. The goal is to test whether transition meaning that is recoverable from the source story is also recoverable from the generated image sequence. This includes causal, emotional, consequence-bearing, temporal, and moral-target transitions.

\subsection{KathaBench-25K Construction}
\label{sec:benchmark_construction}

We instantiate KathaTrace with \textbf{KathaBench-25K}, a transition-level recoverability benchmark built to test whether source-supported narrative meaning survives storyboard generation. The benchmark uses \textbf{5,000 rights-compatible source stories} from three traditions: 1,702 from \emph{Kathasaritsagara}/\emph{The Ocean of Story}, 1,676 from \emph{Aesop's Fables}, and 1,622 from \emph{Panchatantra}. This mixture is needed to avoid evaluating only one narrative style, moral pattern, or cultural source.

Each story is normalized into \textbf{five ordered scenes}, producing \textbf{25K structured scenes} and \textbf{20K adjacent transitions}. Five scenes are sufficient to represent setup, action, complication, consequence, and resolution, while keeping evaluation comparable across stories. The adjacent transitions are the core unit of KathaTrace because semantic trajectory collapse occurs when the viewer cannot recover why one scene follows another.

For each transition, we store \textbf{transition fields} for action, causality, emotion, consequence, and the benchmark-specific moral-semantic target. These fields are needed to localize failures rather than report only an aggregate score. We then derive \textbf{28,712 recoverability questions} with canonical answers and fixed accepted-answer sets, so evaluation checks recoverable meaning rather than free-form judge preference.

Finally, \textbf{10K contrastive variants} preserve major entities and settings while changing one meaning-bearing transition. These variants test whether an evaluator distinguishes true transition preservation from surface similarity. Human validation checks source-scene fidelity, scene order, transition validity, QA answerability, accepted-answer coverage, ambiguity flags, moral-target consistency, and contrastive validity. Together, the source diversity, transition annotations, recoverability questions, contrastives, and validation metadata make KathaBench-25K a benchmark for semantic recoverability, not just a collection of story prompts.

\subsection{Semantic Trajectory Representation}
\label{sec:semantic_trajectory_representation}

We define a \emph{semantic trajectory} as the ordered set of meaning-bearing transitions between adjacent scenes:
\begin{equation}
\tau=\{r_1,r_2,\ldots,r_{T-1}\},
\qquad
r_t=\Phi(s_t,s_{t+1}).
\label{eq:semantic_trajectory}
\end{equation}
Each transition stores six source-side fields:
\begin{equation}
r_t=(a_t,c_t,i_t,e_t,o_t,m_t),
\label{eq:transition_fields}
\end{equation}
where $a_t$ is visible action, $c_t$ is causality, $i_t$ is intention, $e_t$ is emotional shift, $o_t$ is consequence or outcome, and $m_t$ is the benchmark-specified moral-semantic target when applicable. These fields define what meaning should survive story-to-image translation. They are not all scored QA dimensions: intention is retained for annotation and planning, while temporal-order questions are derived from scene order.

\subsection{Recoverability Questions and Evidence Conditions}
\label{sec:recoverability_questions}

KathaTrace evaluates six scored QA dimensions:
\begin{equation}
\begin{aligned}
\mathcal{K}_{\mathrm{QA}}=\{&
\mathrm{action},\mathrm{causal},\mathrm{emotional},\\
&\mathrm{consequence},\mathrm{temporal},\mathrm{moral}
\}.
\end{aligned}
\label{eq:qa_dimensions}
\end{equation}
These correspond to action visibility, cause-effect recovery, emotional-state recovery, consequence recovery, event-order recovery, and benchmark-specified moral-target recovery. Moral labels are semantic recoverability targets, not claims of objective moral truth.

Each item is indexed by story, transition, QA dimension, and question instance:
\begin{equation}
q=(n,t,k,u),
\label{eq:question_index}
\end{equation}
where $n$ is the story, $t$ is the adjacent-scene transition, $k\in\mathcal{K}_{\mathrm{QA}}$ is the scored dimension, and $u$ indexes multiple questions for the same transition-dimension pair. Each question has a canonical answer $y_q$ and a frozen accepted-answer set $\mathcal{A}(q)$.

Each question is evaluated under three evidence conditions:
\begin{itemize}
\item \textbf{Text-only} $(R_{\mathrm{text}})$: source story $S$ only; estimates the source-side recoverability ceiling.
\item \textbf{Image-only} $(R_{\mathrm{image}})$: generated storyboard $\hat{X}$ only, without source text, prompts, captions, annotations, labels, or metadata; tests whether meaning survives in images.
\item \textbf{Text+image}: both $S$ and $\hat{X}$; used only for ambiguity, defect, and contradiction control.
\end{itemize}
Image-only is the primary diagnostic condition because KathaTrace measures whether transition meaning is recoverable from the generated visual narrative itself.

\subsection{Validity Filtering and Meaning Preservation}
\label{sec:validity_filtering_recoverability}

Meaning is treated as preserved when the accepted answer for a source-supported transition remains recoverable from generated images alone. Before STG computation, KathaTrace removes questions whose intended answer is not recoverable even with text+image evidence. For each scored dimension $k$, let $Q_k$ be the full question set. The valid set is:
\begin{equation}
\mathcal{V}_k=
\left\{
q\in Q_k:
\mathrm{match}\big(J(q,(S,\hat{X})),\mathcal{A}(q)\big)=1
\right\},
\label{eq:valid_question_set}
\end{equation}
where $J(q,z)$ is the judge answer under evidence condition $z$. Questions outside $\mathcal{V}_k$ are excluded from STG and counted in the ambiguity rate, preventing unclear questions or image-text contradictions from being treated as generator failures.

For evidence condition $z\in\{S,\hat{X},(S,\hat{X})\}$, dimension-specific recoverability is:
\begin{equation}
R^k_z =
\frac{1}{|\mathcal{V}_k|}
\sum_{q\in \mathcal{V}_k}
\mathbb{I}\!\left[
\mathrm{match}\big(J(q,z),\mathcal{A}(q)\big)=1
\right].
\label{eq:recoverability}
\end{equation}
Overall recoverability averages the six scored dimensions equally:
\begin{equation}
R_z=
\frac{1}{|\mathcal{K}_{\mathrm{QA}}|}
\sum_{k\in\mathcal{K}_{\mathrm{QA}}}R^k_z.
\label{eq:overall_recoverability}
\end{equation}
The ambiguity rate is:
\begin{equation}
\rho_{\mathrm{amb}}=
\frac{1}{|Q|}
\sum_{q\in Q}
\mathbb{I}\!\left[
q\notin \mathcal{V}_{k(q)}
\right],
\label{eq:ambiguity_rate}
\end{equation}
where $k(q)$ is the scored QA dimension of question $q$.

\subsection{Semantic Trajectory Gap: Diagnosing Meaning Collapse}
\label{sec:semantic_trajectory_gap}

The primary metric is the \emph{Semantic Trajectory Gap}:
\begin{equation}
\mathrm{STG}=R_{\mathrm{text}}-R_{\mathrm{image}},
\label{eq:stg}
\end{equation}
where $R_{\mathrm{text}}=R_S$ and $R_{\mathrm{image}}=R_{\hat{X}}$. $R_{\mathrm{text}}$ measures source-side recoverability, while $R_{\mathrm{image}}$ measures what remains recoverable after visual generation. Low STG indicates strong meaning preservation; high STG indicates semantic trajectory collapse, where source-supported transition meaning is missing from the image-only storyboard. The same definition applies to VLM and human judgments.

KathaTrace also reports dimension-specific gaps:
\begin{equation}
\mathrm{STG}_k=
R^k_{\mathrm{text}}-R^k_{\mathrm{image}},
\qquad
k\in\mathcal{K}_{\mathrm{gap}}.
\label{eq:dimension_stg}
\end{equation}
For latent-transition analysis, we use:
\begin{equation}
\mathcal{K}_{\mathrm{gap}}=
\{
\mathrm{causal},
\mathrm{emotional},
\mathrm{consequence},
\mathrm{moral}
\}.
\label{eq:gap_dimensions}
\end{equation}
Action visibility and temporal order remain part of overall recoverability but are reported separately because they test local depiction and ordering control rather than latent transition loss.

\subsection{Contrastive Semantic Variants: Isolating Meaning from Appearance}
\label{sec:contrastive_variants}

KathaBench-25K includes contrastive variants to test whether recoverability tracks transition meaning rather than shared objects, characters, or settings. Each variant preserves major entities where possible but changes one meaning-bearing transition, such as consequence, motivation, outcome, emotional shift, or causal link. Thus, visual appearance is held close while the semantic trajectory changes. Pairwise contrastive evaluation tests whether recovered meaning matches the source trajectory rather than a visually similar but semantically different alternative. The counterfactual minimal-pair category contains 400 source stories, while the 10,000 contrastive variants are constructed from all 5,000 narratives.

\subsection{Judge Aggregation and Calibration Validation}
\label{sec:judge_aggregation}

KathaTrace supports human or VLM judges. For scalable evaluation, it uses fixed evidence packets, fixed prompt templates, deterministic decoding where supported, structured output schemas, frozen accepted-answer sets, and majority-vote aggregation after answer normalization. Ties are scored conservatively as incorrect. Human calibration tests whether VLM recoverability follows human interpretation rather than a single-model artifact, using agreement, rank correlation, calibration error, confidence intervals, and inter-rater agreement. \textit{\textbf{See Appendix Secs. \hyperref[sece]{E}--\hyperref[secf]{F} for judge protocols and human calibration details}}. These checks validate STG as a recoverability measure rather than treating any one judge as ground truth.

\subsection{Semantic Compass: Validating STG as an Actionable Signal}
\label{sec:semantic_compass_method}

\textbf{Semantic Compass} tests whether KathaTrace diagnostics are actionable. It is not a new generator; it is a post-generation reranking and bridge-repair probe that asks whether localized unrecoverable transitions can guide better storyboard selection or minimal repair.

Given candidate storyboards $\{\hat{X}^{(1)},\ldots,\hat{X}^{(N_c)}\}$ for the same source story, Semantic Compass selects:
\begin{equation}
\begin{aligned}
\hat{X}^{*}
= \arg\max_{\hat{X}^{(j)}} \Big[
&\lambda_r R^{\mathrm{val}}_{\mathrm{image}}(\hat{X}^{(j)})
+ \lambda_t C_{\mathrm{trans}}(\hat{X}^{(j)},\tau) \\
&- \lambda_p P_{\mathrm{copy}}(\hat{X}^{(j)})
\Big],
\end{aligned}
\label{eq:semantic_compass}
\end{equation}
where $R^{\mathrm{val}}_{\mathrm{image}}$ rewards validation image-only recoverability, $C_{\mathrm{trans}}$ rewards visual support for annotated transitions, and $P_{\mathrm{copy}}$ penalizes repeated or near-duplicate frames.

If the weakest adjacent transition is localizable and falls below a validation-selected threshold, Semantic Compass may insert a bridge frame:
\begin{equation}
\hat{X}_{\mathrm{bridge}}
=
\{\hat{x}_1,\ldots,\hat{x}_t,\hat{x}_{t+\frac{1}{2}},
\hat{x}_{t+1},\ldots,\hat{x}_T\}.
\label{eq:bridge_sequence}
\end{equation}
All weights, thresholds, candidate counts, and repair decisions are selected on validation data and frozen before final evaluation. Final reporting uses held-out questions and held-out judge prompts, so results test whether STG exposes usable recoverability failures rather than whether Semantic Compass is a new generator.

\begin{figure*}[!t]
\centering
\includegraphics[width=\textwidth]{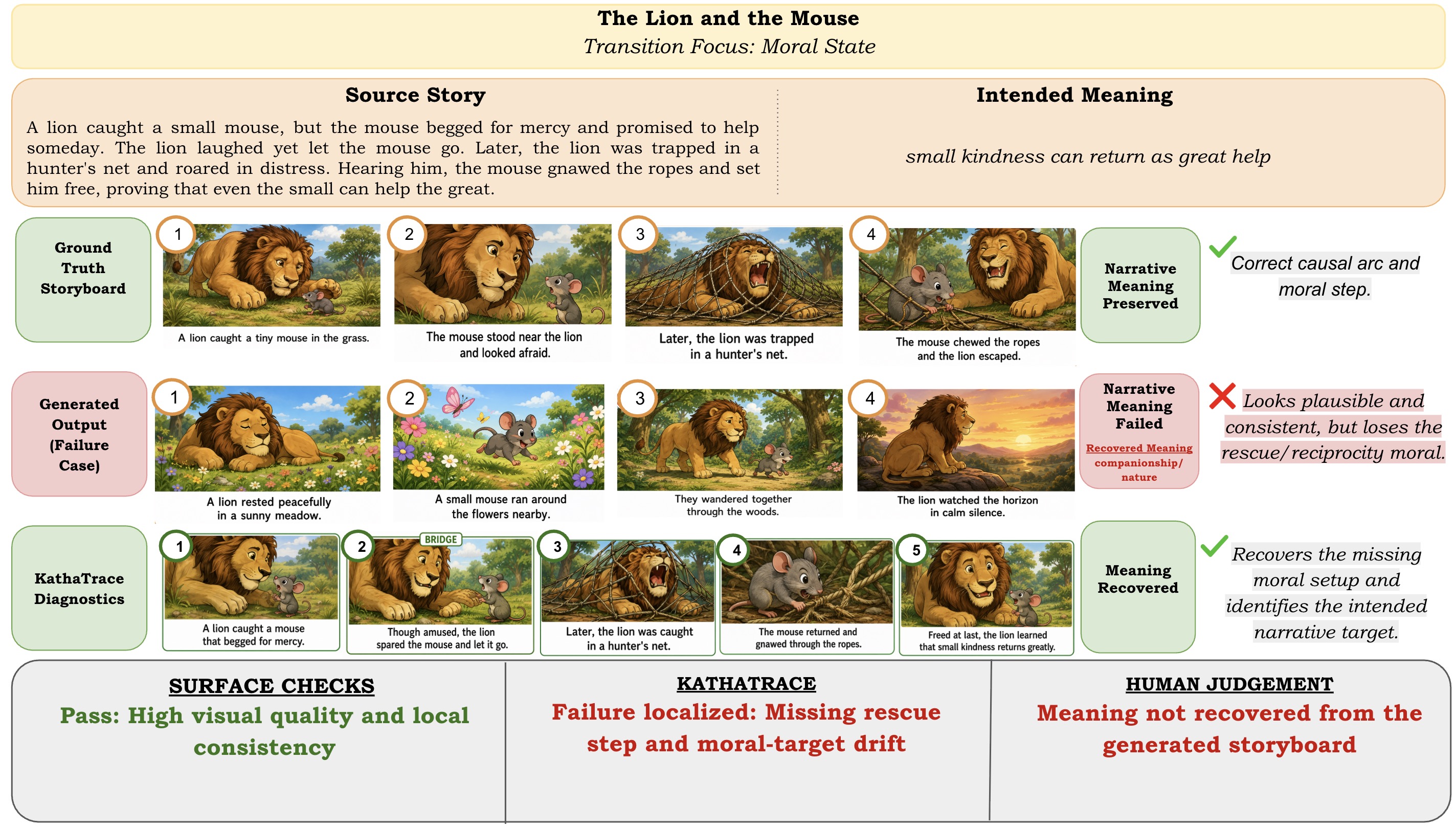}
\caption{
\textbf{Case study of a missed narrative failure.}
The generated storyboard remains visually plausible and locally consistent, but omits the rescue transition needed to recover the intended reciprocity moral. KathaTrace localizes this transition loss and moral-target drift, matching human judgment.
}
\label{fig:benchmark_failure_case}

\end{figure*}

\section{Experiments}
\label{sec:experiments}

We evaluate whether generated visual narratives preserve \emph{transition-level recoverable meaning} under image-only evidence. The experiments test three claims: semantic trajectory collapse is distinct from visual quality, transition-specific evaluation is necessary beyond generic QA, and localized recoverability failures provide actionable signals. All STG values are paired within the same valid examples after ambiguity filtering; raw, filtered, and common-valid audits test whether conclusions depend on filtering.

\subsection{Experimental Setup}

\noindent\textbf{Benchmark and split.}
We evaluate held-out KathaBench-25K examples with method-valid accounting. The release contains 5,000 narratives, 25,000 scenes, 20,000 transitions, 28,712 QA pairs, and 10,000 contrastive variants, with fixed train/validation/public-test/hidden-test splits of 3,500/750/500/250 stories. Validation is used for Semantic Compass selection, public-test for main reporting, and hidden-test for leaderboard evaluation. Common-valid results check whether rankings depend on method-specific usable-storyboard failures.

\noindent\textbf{Evaluated systems.}
We evaluate StoryDiffusion~\cite{zhou2024storydiffusion}, StoryGPT-V~\cite{shen2025storygpt}, DreamStory~\cite{he2025dreamstory}, Story-Iter~\cite{maostory}, ViSTA~\cite{dong2026vista}, LogiStory~\cite{meng2026logistory}, and Gemma-ST + Semantic Compass. Semantic Compass is an actionability probe, not a new generator. Extended experiments vary planners and renderers, including direct prompting, rule planning, caption planning, scene planning, Gemma-ST, SDXL, FLUX, ConSistory, and closed API references.

\noindent\textbf{Judges and reporting.}
Main results use an ensemble of Gemini 2.5 Flash, Qwen2.5-VL-7B/3B, and SmolVLM2-2.2B/500M with fixed evidence packets, frozen accepted answers, structured schemas, and deterministic decoding where supported. On the strict 400-story human-gold subset, the calibrated VLM ensemble reaches $77.1\%$ agreement, Spearman $\rho=0.74$, ECE $=0.081$, and $\kappa=0.77$. Human evaluation confirms the same pattern with $\mathrm{STG}^{\mathrm{human}}=15.8{\pm}2.7$.

\subsection{Main Result}

Table~\ref{tab:kathatrace_generation_methods} shows that visual quality does not imply transition recoverability. Prior methods cluster in a narrow STG range of $28.4$--$34.8$ despite different architectures, planners, and rendering strategies, ruling out a single-method artifact. Story-Iter has the highest controllable visual score, $4.18{\pm}0.05$, but $29.7{\pm}1.6$ STG. LogiStory has the strongest controllable image recoverability, $49.4{\pm}1.3$, but still retains $28.4{\pm}1.5$ STG. On semantic contrastive variants, LogiStory rises to $41.2$ STG versus $28.4$ on source trajectories, showing that KathaTrace is sensitive to meaning changes rather than shared appearance. Gemma-ST + Semantic Compass lowers STG to $21.4{\pm}1.3$ under held-out evaluation.

\subsection{Does Transition Recoverability Require More Than Generic QA?}
\label{sec:transition_recoverability_vs_generic_qa}

Table~\ref{tab:evaluator_baselines} shows that transition recoverability is not captured by generic image-level question answering alone. Higher values are better for image recoverability, localization, and human correlation, whereas lower values are better for STG because STG measures the gap between source-story transitions and generated visual transitions.

The strongest adapted generic baseline, multi-frame DSG-style QA, reaches
$48.0{\pm}1.4$ image recoverability, $28.9{\pm}1.5$ STG,
$0.50{\pm}0.03$ localization, and $0.58{\pm}0.04$ human correlation.
Full KathaTrace improves image recoverability to $54.6{\pm}1.2$,
reduces STG to $23.5{\pm}1.3$, improves localization to
$0.68{\pm}0.02$, and increases human correlation to $0.71{\pm}0.03$.
This corresponds to a $6.6$-point gain in image recoverability, a
$5.4$-point reduction in STG, a $0.18$ gain in localization, and a
$0.13$ gain in correlation with human judgments.

These results indicate that the evaluator is measuring more than whether
individual images contain answerable objects or events. Generic QA can
identify visible entities and some frame-level actions, but it does not
explicitly test whether a visual sequence preserves the causal, temporal,
motivational, and consequence-bearing transitions required by the source
story. The ablation results support this interpretation: transition
structure, dimension-specific recoverability questions, and contrastive
checks each contribute additional alignment with human recoverability
judgments. The improvement is therefore cumulative rather than attributable
to a single prompt-format change or a generic increase in QA coverage.

\begin{table*}[t]
\centering
\scriptsize
\setlength{\tabcolsep}{2.5pt}
\renewcommand{\arraystretch}{0.92}
\caption{
\textbf{KathaTrace evaluation of visual-story generation methods.}
Values are mean $\pm$ 95\% bootstrap CI\@. STG $=R_{\text{text}}-R_{\text{image}}$ on validity-filtered questions; best values are bolded.
$\ddagger$ marks a closed API reference whose internals are not fully controllable.
Semantic Compass is reported as a KathaTrace-guided actionability probe, not as a new generator.
$\dagger$ marks Bonferroni-corrected paired-bootstrap improvement over the strongest controllable baseline ($p_{\text{corr}}<0.01$).
}
\label{tab:kathatrace_generation_methods}
\resizebox{\textwidth}{!}{%
\begin{tabular}{lcccccc}
\toprule
\rowcolor{gray!20}
\textbf{Method} &
\textbf{Visual} $\uparrow$ &
\textbf{Text Rec.} $\uparrow$ &
\textbf{Image Rec.} $\uparrow$ &
\textbf{STG} $\downarrow$ &
\textbf{Moral-target Rec.} $\uparrow$ &
\textbf{Human Score} $\uparrow$ \\
\midrule
\multicolumn{7}{l}{\textit{Controllable generator baselines}} \\
StoryDiffusion~\cite{zhou2024storydiffusion}
  & $3.82{\pm}0.06$ & $74.6{\pm}1.0$ & $39.8{\pm}1.5$ & $34.8{\pm}1.7$ & $28.4{\pm}1.6$ & $3.21{\pm}0.07$ \\
\rowcolor{gray!5}
StoryGPT-V~\cite{shen2025storygpt}
  & $4.05{\pm}0.05$ & $76.9{\pm}0.9$ & $44.6{\pm}1.4$ & $32.3{\pm}1.6$ & $33.7{\pm}1.5$ & $3.46{\pm}0.06$ \\
DreamStory~\cite{he2025dreamstory}
  & $4.12{\pm}0.05$ & $76.4{\pm}0.9$ & $47.2{\pm}1.4$ & $29.2{\pm}1.5$ & $36.9{\pm}1.5$ & $3.58{\pm}0.06$ \\
\rowcolor{gray!5}
Story-Iter~\cite{maostory}
  & \textbf{$4.18{\pm}0.05$} & $75.8{\pm}1.0$ & $46.1{\pm}1.5$ & $29.7{\pm}1.6$ & $35.8{\pm}1.6$ & $3.51{\pm}0.06$ \\
ViSTA~\cite{dong2026vista}
  & $4.01{\pm}0.05$ & $75.9{\pm}1.0$ & $43.7{\pm}1.5$ & $32.2{\pm}1.7$ & $32.4{\pm}1.6$ & $3.44{\pm}0.06$ \\
\rowcolor{gray!5}
LogiStory~\cite{meng2026logistory}
  & $3.94{\pm}0.06$ & \textbf{$77.8{\pm}0.8$} & \textbf{$49.4{\pm}1.3$} & \textbf{$28.4{\pm}1.5$} & \textbf{$39.8{\pm}1.4$} & \textbf{$3.64{\pm}0.06$} \\
\midrule
\multicolumn{7}{l}{\textit{Closed-system reference}} \\
\rowcolor{gray!12}
GPT-4o + GPT-image-1$^{\ddagger}$
  & $4.35{\pm}0.05$ & $77.6{\pm}0.8$ & $52.4{\pm}1.3$ & $25.2{\pm}1.4$ & $41.7{\pm}1.4$ & $3.73{\pm}0.06$ \\
\midrule
\multicolumn{7}{l}{\textit{KathaTrace-guided actionability probe}} \\
\rowcolor{gray!5}
Gemma-ST + Semantic Compass
  & $4.16{\pm}0.05$
  & $77.2{\pm}0.8$
  & \textbf{$55.8{\pm}1.2$}$^{\dagger}$
  & \textbf{$21.4{\pm}1.3$}$^{\dagger}$
  & $39.1{\pm}1.3$
  & \textbf{$3.91{\pm}0.05$}$^{\dagger}$ \\
\bottomrule
\end{tabular}%
}
\vspace{3pt}
\raggedright\footnotesize
\textit{Visual}: 1--5 human visual-quality rating.\quad
\textit{Text\,/\,Image Rec.}: text-only\,/\,image-only recoverability.\quad
\textit{Moral-target Rec.}: benchmark-specific moral-semantic target recovery, not objective moral truth. $^{\ddagger}$Internal prompt rewriting, safety filtering, hardware, and model versioning are not fully controllable.

\end{table*}

\begin{table*}[t]
\centering
\scriptsize
\setlength{\tabcolsep}{3.5pt}
\renewcommand{\arraystretch}{0.92}
\caption{
\textbf{Evaluator baselines and KathaTrace component ablation.}
Values are mean $\pm$ 95\% bootstrap CI. $\dagger$ marks Bonferroni-corrected
improvement over the strongest adapted evaluator ($p_{\mathrm{corr}}<0.01$).
}
\label{tab:evaluator_baselines}

\begin{minipage}{0.92\textwidth}
\centering
\resizebox{\linewidth}{!}{%
\begin{tabular}{lcccc}
\toprule
\rowcolor{gray!20}
\textbf{Evaluator} &
\textbf{Image Rec.} $\uparrow$ &
\textbf{STG} $\downarrow$ &
\textbf{Loc.} $\uparrow$ &
\textbf{Human Corr.} $\uparrow$ \\
\midrule
Object-overlap QA
& $39.6{\pm}1.5$ & $36.2{\pm}1.7$ & $0.27{\pm}0.03$ & $0.39{\pm}0.04$ \\

\rowcolor{gray!5}
Generic story QA
& $41.2{\pm}1.5$ & $34.9{\pm}1.7$ & $0.31{\pm}0.03$ & $0.44{\pm}0.04$ \\

Multi-frame TIFA-style~\cite{hu2023tifa}
& $45.8{\pm}1.4$ & $31.1{\pm}1.6$ & $0.44{\pm}0.03$ & $0.52{\pm}0.04$ \\

\rowcolor{gray!5}
Multi-frame DSG-style~\cite{cho2024davidsonian}
& $48.0{\pm}1.4$ & $28.9{\pm}1.5$ & $0.50{\pm}0.03$ & $0.58{\pm}0.04$ \\

\midrule
KathaTrace: QA only
& $49.1{\pm}1.3$ & $28.6{\pm}1.5$ & $0.45{\pm}0.03$ & $0.61{\pm}0.04$ \\

\rowcolor{gray!5}
+ transition structure
& $51.0{\pm}1.3$ & $26.9{\pm}1.4$ & $0.55{\pm}0.03$ & $0.65{\pm}0.04$ \\

+ dimension-specific questions
& $52.8{\pm}1.3$ & $25.1{\pm}1.4$ & $0.63{\pm}0.02$ & $0.68{\pm}0.03$ \\

\rowcolor{blue!12}
\textbf{Full KathaTrace (+ contrastives)}
& \textbf{$54.6{\pm}1.2^{\dagger}$}
& \textbf{$23.5{\pm}1.3^{\dagger}$}
& \textbf{$0.68{\pm}0.02^{\dagger}$}
& \textbf{$0.71{\pm}0.03^{\dagger}$} \\
\bottomrule
\end{tabular}%
}

\vspace{2pt}
\footnotesize
\raggedright
\textit{Image Rec.}: image-only recoverability.
\textit{Loc.}: dimension-level failure localization.
\textit{Human Corr.}: correlation with human judgments.
The full protocol improves image-only recoverability by $5.5$ points over the QA-only KathaTrace ablation.
\end{minipage}

\end{table*}

\subsection{Planning, Rendering, and Failure Signatures}
\label{sec:planning_rendering_failure_signatures}

Planner--generator ablations show that planning quality affects narrative
recoverability beyond the choice of renderer. Holding the renderer fixed as
FLUX, replacing direct prompting with Gemma-ST increases image
recoverability from $40.8$ to $53.8$ and reduces STG from $35.2$ to
$24.0$. This is a $13.0$-point gain in image recoverability and an
$11.2$-point reduction in STG. The result shows that an explicit
story-transition plan helps the generator preserve the information needed
to recover the source narrative from the generated image sequence.

In contrast, when the planner is fixed as Gemma-ST, changing the renderer
reduces STG by only $4.1$ points on average. This does not mean renderer
quality is irrelevant; renderer choice still affects visual fidelity,
composition, and object depiction. However, the smaller STG change indicates
that renderer substitution alone does not solve the main failure mode. The
larger improvement comes from improving the intermediate narrative plan,
which controls what transition information is made available to the image
generator.

The failure signatures further show that different methods fail in
different transition dimensions. DreamStory exhibits larger consequence
gaps, meaning that generated sequences often fail to preserve what changes
after an event. StoryDiffusion exhibits larger causal and action gaps,
meaning that actions and their causal links are more often omitted,
weakened, or visually underspecified. LogiStory produces a more balanced
profile, but its nonzero gaps show that it still loses transition-level
information. These patterns are inconsistent with a purely generic
visual-quality explanation. If failures were only caused by low image
quality, the errors would be expected to affect all transition dimensions
more uniformly. Instead, the observed dimension-specific gaps support the need for transition-specific diagnosis.

\subsection{Generalization, Scope, and Hard Cases}

Table~\ref{tab:stg_validity_controls} summarizes controls for generalization and scope. Results are stable across source families, VLM judges, and dataset subsamples. Fig.~\ref{fig:benchmark_failure_case} illustrates the core failure mode: a storyboard can preserve visual plausibility and local character consistency while omitting the rescue transition needed to recover the intended reciprocity moral. KathaTrace localizes this missing transition and moral-target drift, matching the human judgment that the generated storyboard does not preserve the intended meaning. More generally, KathaTrace is most reliable when meaning is expressed through ordered scenes, transition-level questions, and fixed accepted answers. Symbolic, ironic, and long-range narratives show higher ambiguity, so STG should be reported with ambiguity rates and confidence intervals. 

\begin{table*}[!t]
\centering
\scriptsize
\setlength{\tabcolsep}{3.0pt}
\renewcommand{\arraystretch}{0.92}
\caption{
\textbf{Robustness controls for STG validity.}
Each control tests an alternative explanation for the observed text-to-image recoverability gap.
}
\label{tab:stg_validity_controls}
\resizebox{\textwidth}{!}{%
\begin{tabular}{p{3.1cm}p{5.2cm}p{6.2cm}}
\toprule
\rowcolor{gray!20}
\textbf{Threat tested} & \textbf{Audit result} & \textbf{Interpretation} \\
\midrule
Filtering artifact &
Raw, filtered, and common-valid STG rankings are highly correlated ($\rho=0.96$). &
The main ranking is not driven by ambiguity filtering or method-specific valid subsets. \\

\rowcolor{gray!5}
Source-family artifact &
Across source traditions, STG remains in a narrow range ($27.8$--$29.6$). &
The recoverability gap is not specific to one source collection or narrative tradition. \\

Single-judge artifact &
Gemini--Qwen STG correlation is high ($\rho=0.92$). &
The trend is stable across different VLM judge families. \\

\rowcolor{gray!5}
Dataset-size artifact &
Half-data subsampling preserves the STG ranking ($\rho=0.94$). &
The leaderboard is not an artifact of a small or unstable evaluation sample. \\

Surface-appearance artifact &
LogiStory STG increases from $28.4$ on source trajectories to $41.2$ on semantic contrastives. &
KathaTrace is sensitive to transition meaning, not only shared objects, characters, or settings. \\

\rowcolor{gray!5}
Renderer-only explanation &
Planner changes reduce STG by $11.2$ points, while renderer changes reduce STG by $4.1$ points. &
Transition-aware planning contributes more than rendering quality alone. \\

Scope and ambiguity &
Action-centric stories show STG $=18.4$ with $3\%$ ambiguity; symbolic stories show STG $=35.2$ with $22\%$ ambiguity. &
Harder narrative types are explicitly scoped using ambiguity rates rather than hidden in aggregate scores. \\
\bottomrule
\end{tabular}%
}
\end{table*}

\textit{\textbf{See Appendix Secs. \hyperref[secg]{G}--\hyperref[secm]{M} for implementation details, additional baselines, leaderboards, ablations, qualitative cases, extensibility examples, and robustness checks.}}
\vspace{-4pt}
\section{Conclusion and Future Work}
\label{sec:conclusion}
\vspace{-4pt}
We introduced \textbf{KathaTrace}, an image-only recoverability benchmark for evaluating whether generated storyboards preserved the transition-level meaning of source narratives. The benchmark was designed to test whether a viewer could recover not only visible entities and actions, but also the causal, temporal, emotional, consequence-bearing, and moral-target information that connected one story state to the next. This addressed a limitation of generic visual-quality and image-QA evaluations, which often treated frame-level correctness as sufficient evidence of narrative faithfulness. Experiments on \textbf{KathaBench-25K} showed that visually plausible storyboards still lost key transition information.
Generated sequences often preserved characters, settings, and isolated events, but weakened causal, emotional, consequence-bearing, and moral-target relations.
These failures showed that semantic trajectory collapse was distinct from poor visual quality, object omission, or generic QA failure.
Evaluator ablations showed that transition structure, dimension-specific questions, and contrastive checks aligned better with human judgments than generic multi-frame QA.
\textbf{Semantic Compass} localized the collapsed dimensions, making recoverability failures actionable for storyboard repair. Future work would need to address symbolic, ironic, and long-range transition recovery. These cases remained difficult because their meanings were often indirect, contrastive, or dependent on events separated across multiple scenes. Stronger planning modules, transition-aware generation objectives, and repair mechanisms would be needed to produce storyboards whose narrative meaning could be reliably recovered from images alone.

\bibliography{software}

%%%%%%%%%%%%%%%%%%%%%%%%%%%%%%%%%%%%%%%%%%%%%%%%%%%%%%%

\clearpage
\appendix
\part*{Appendix}
\addcontentsline{toc}{part}{Appendix}
\etocsettocstyle{\section*{Appendix Contents}}{}
\etocsetnexttocdepth{subsection}
\localtableofcontents

%%%%%%%%%%%%%%%%%%%%%%%%%%%%%%%%%%%%%%%%%%%%%%%%%%%%%

\section{Dataset Construction and Human Validation Details}
\label{seca}

This section specifies the construction, annotation schema, validation protocol, split design, provenance controls, release structure, and quality-control procedures for the proposed \textsc{KathaBench-25K} benchmark. \textsc{KathaBench-25K} is designed for transition-level recoverability evaluation in generated visual narratives: given a source story and a generated storyboard, the benchmark tests whether the story-transition meaning remains recoverable from images alone. We separate \emph{source-side transition fields}, which describe the intended semantic trajectory, from \emph{scored QA dimensions}, which define the frozen recoverability-question inventory used during evaluation.

\subsection{Construction Overview}
\label{app:construction_overview}

The \textsc{KathaBench-25K} construction pipeline has five stages: source narrative selection, scene structuring, transition annotation, recoverability-question construction, and contrastive-variant construction. Source narratives are drawn from three public-domain or rights-compatible classical narrative families: \emph{Kathasaritsagara} / \emph{The Ocean of Story} (1,702 stories), \emph{Aesop's Fables} (1,676 stories), and \emph{Panchatantra} (1,622 stories). These sources were selected because they contain compact, transition-dependent narratives in which causal action, consequence, emotional change, motivation, and moral-semantic interpretation are central to story understanding.

Each story is normalized into five ordered scenes. This fixed length supports controlled comparison across generators while exposing four adjacent transitions per story. The KathaTrace protocol is length-general, but the released \textsc{KathaBench-25K} instance fixes $T=5$ so that scene boundaries, transition targets, recoverability questions, and accepted-answer sets remain comparable across records. Each record stores provenance metadata, source-rights status, validation status, five scene records, four transition records, and recoverability questions with frozen accepted-answer sets.

Table~\ref{tab:appendix_dataset_summary} gives the released \textsc{KathaBench-25K} component counts used throughout the benchmark.

\subsection{Narrative Category Taxonomy}
\label{app:narrative_categories}

Each \textsc{KathaBench-25K} narrative receives one primary category for balancing, stratified sampling, and category-specific error analysis. Table~\ref{tab:appendix_narrative_categories} reports the eight-way category taxonomy and the semantic stress test associated with each category.

\subsection{Scene Representation Schema}
\label{app:scene_schema}

Each narrative in \textsc{KathaBench-25K} is decomposed into ordered scenes. A scene is treated as a local visual unit, but it also stores narrative-state fields needed to interpret the scene within the larger semantic trajectory. Table~\ref{tab:appendix_scene_schema} defines the scene-level schema released with each benchmark record.

\subsection{Transition Annotation Schema}
\label{app:transition_schema}

The transition between adjacent scenes is the central unit of \textsc{KathaBench-25K}. Each record stores a coarse \texttt{transition\_type} for provenance and balancing, together with source-side semantic fields that describe the intended relation between scenes. These fields are not identical to the scored QA inventory: \texttt{intention} is retained for annotation and planning, whereas temporal-order questions are derived from scene order. Table~\ref{tab:appendix_transition_schema} defines the transition annotation schema.

\begin{table*}[t]
\centering
\scriptsize
\setlength{\tabcolsep}{4pt}
\renewcommand{\arraystretch}{1.05}
\caption{\textbf{\textsc{KathaBench-25K} components.} Released counts by benchmark artifact.}
\label{tab:appendix_dataset_summary}
\resizebox{\textwidth}{!}{
\begin{tabular}{l r p{7.2cm}}
\toprule
\rowcolor{gray!20}
\textbf{Component} & \textbf{Count} & \textbf{Description} \\
\midrule
Narratives & 5,000 & Source stories used as semantic specifications. \\
\rowcolor{gray!5}
Structured scenes & 25,000 & Five ordered scene representations per story. \\
Scene transitions & 20,000 & Four adjacent scene-pair annotations per story. \\
\rowcolor{gray!5}
Recoverability QA pairs & 28,712 & Questions testing transition meaning under controlled evidence conditions. \\
Contrastive variants & 10,000 & Two semantic variants per story that alter causal, emotional, consequence-bearing, temporal, or moral-target structure. \\
\rowcolor{gray!5}
Narrative categories & 8 & High-level narrative types used for stratification and category-specific analysis. \\
Moral-target labels & 12 & Benchmark-specified moral-semantic targets used for moral-target recoverability analysis. \\
\bottomrule
\end{tabular}
}
\end{table*}

\begin{table*}[t]
\centering
\scriptsize
\setlength{\tabcolsep}{4pt}
\renewcommand{\arraystretch}{1.05}
\caption{\textbf{\textsc{KathaBench-25K} categories.} Narrative counts and stress tests.}
\label{tab:appendix_narrative_categories}
\resizebox{\textwidth}{!}{
\begin{tabular}{l r r p{6.2cm}}
\toprule
\rowcolor{gray!20}
\textbf{Narrative Type} & \textbf{Narratives} & \textbf{Scenes} & \textbf{Primary Stress Test} \\
\midrule
Moral-semantic stories & 1,000 & 5,000 & Moral-target state, ethical choice, and consequence preservation. \\
\rowcolor{gray!5}
Causal-transition stories & 800 & 4,000 & Cause-effect preservation across ordered scenes. \\
Emotional trajectory stories & 700 & 3,500 & Emotion evolution and affective state change. \\
\rowcolor{gray!5}
Procedural state-change stories & 600 & 3,000 & Ordered actions and object-state changes. \\
Social-interaction stories & 600 & 3,000 & Motivation, conflict, apology, reconciliation, and social repair. \\
\rowcolor{gray!5}
Hidden-consequence stories & 500 & 2,500 & Delayed outcomes and nonlocal narrative meaning. \\
Cultural / folk moral stories & 400 & 2,000 & Culturally grounded moral-target transfer. \\
\rowcolor{gray!5}
Counterfactual minimal-pair stories & 400 & 2,000 & Small semantic changes with large narrative consequences. \\
\midrule
\rowcolor{gray!12}
\textbf{Total} & \textbf{5,000} & \textbf{25,000} & \textbf{Semantic trajectory preservation.} \\
\bottomrule
\end{tabular}
}
\end{table*}

\begin{table*}[t]
\centering
\scriptsize
\setlength{\tabcolsep}{4pt}
\renewcommand{\arraystretch}{1.05}
\caption{\textbf{\textsc{KathaBench-25K} scene schema.} Fields stored for each scene.}
\label{tab:appendix_scene_schema}
\resizebox{\textwidth}{!}{
\begin{tabular}{l p{4.2cm} p{5.2cm}}
\toprule
\rowcolor{gray!20}
\textbf{Field} & \textbf{Type} & \textbf{Description} \\
\midrule
\texttt{story\_id} & string & Unique identifier for the narrative. \\
\rowcolor{gray!5}
\texttt{scene\_index} & integer & Temporal position of the scene. \\
\texttt{scene\_text} & string & Natural-language scene description. \\
\rowcolor{gray!5}
\texttt{characters} & list[string] & Characters or agents appearing in the scene. \\
\texttt{objects} & list[string] & Salient objects required to understand the scene. \\
\rowcolor{gray!5}
\texttt{action} & string & Main visible or narratively relevant action. \\
\texttt{state} & string & Local narrative state represented by the scene. \\
\rowcolor{gray!5}
\texttt{emotion} & string / optional & Dominant emotional state, when applicable. \\
\texttt{visual\_evidence} & string / optional & Visual evidence relevant to recoverability questions. \\
\rowcolor{gray!5}
\texttt{visual\_must\_show} & list[string] & Elements that should be visible for the scene to support transition recovery. \\
\texttt{generation\_prompt} & string & Prompt used for storyboard generation experiments. \\
\bottomrule
\end{tabular}
}
\end{table*}

\begin{table*}[t]
\centering
\scriptsize
\setlength{\tabcolsep}{4pt}
\renewcommand{\arraystretch}{1.05}
\caption{\textbf{\textsc{KathaBench-25K} transition schema.} Fields for adjacent scenes.}
\label{tab:appendix_transition_schema}
\resizebox{\textwidth}{!}{
\begin{tabular}{l p{4.0cm} p{5.4cm}}
\toprule
\rowcolor{gray!20}
\textbf{Field} & \textbf{Type} & \textbf{Description} \\
\midrule
\texttt{story\_id} & string & Unique narrative identifier. \\
\rowcolor{gray!5}
\texttt{from\_scene} & integer & Index of the source scene. \\
\texttt{to\_scene} & integer & Index of the target scene. \\
\rowcolor{gray!5}
\texttt{transition\_type} & categorical & Coarse category of semantic change. \\
\texttt{action} & string & Visible action or event change. \\
\rowcolor{gray!5}
\texttt{causality} & string & Why the later scene follows from the earlier scene. \\
\texttt{intention} & string / optional & Character goal, plan, or motivation; used for annotation and planning, not as a scored QA dimension. \\
\rowcolor{gray!5}
\texttt{emotion} & string / optional & Emotional state or emotional shift. \\
\texttt{consequence} & string & Outcome produced or revealed by the transition. \\
\rowcolor{gray!5}
\texttt{expected\_change} & string & Expected state change after the transition. \\
\texttt{moral\_target} & string / optional & Benchmark-specified moral-semantic target supported by the trajectory. \\
\rowcolor{gray!5}
\texttt{recoverability\_target} & string & Information expected to be recoverable from a storyboard. \\
\bottomrule
\end{tabular}
}
\end{table*}

Table~\ref{tab:appendix_transition_distribution} reports the coarse transition-type distribution in \textsc{KathaBench-25K}. These coarse types are used for provenance and balancing, not as scored QA dimensions.

\begin{table*}[t]
\centering
\scriptsize
\setlength{\tabcolsep}{4pt}
\renewcommand{\arraystretch}{1.05}
\caption{\textbf{\textsc{KathaBench-25K} transition types.} Coarse transition counts.}
\label{tab:appendix_transition_distribution}
\resizebox{\textwidth}{!}{
\begin{tabular}{l r p{6.8cm}}
\toprule
\rowcolor{gray!20}
\textbf{Transition Type} & \textbf{Count} & \textbf{Description} \\
\midrule
\texttt{cause\_effect} & 10,000 & A prior action or event causes a later event or state. \\
\rowcolor{gray!5}
\texttt{moral\_choice} & 5,000 & A character makes a benchmark-relevant moral or ethical decision. \\
\texttt{consequence\_reveal} & 5,000 & A later scene reveals the consequence of an earlier action. \\
\midrule
\rowcolor{gray!12}
\textbf{Total} & \textbf{20,000} & \textbf{Adjacent scene transitions.} \\
\bottomrule
\end{tabular}
}
\end{table*}

Table~\ref{tab:appendix_transition_field_coverage} reports the source-side field coverage for the 20,000 adjacent-scene transitions. All transitions store action, causality, intention, emotion, consequence, and expected-change fields, while moral-target fields are stored for the 5,000 story-level moral targets.

\begin{table*}[t]
\centering
\scriptsize
\setlength{\tabcolsep}{4pt}
\renewcommand{\arraystretch}{1.05}
\caption{\textbf{\textsc{KathaBench-25K} transition fields.} Coverage by semantic field.}
\label{tab:appendix_transition_field_coverage}
\resizebox{\textwidth}{!}{
\begin{tabular}{l r p{7.0cm}}
\toprule
\rowcolor{gray!20}
\textbf{Transition Field} & \textbf{Transition Records} & \textbf{Use} \\
\midrule
\texttt{action} & 20,000 & Records visible event or physical state change. \\
\rowcolor{gray!5}
\texttt{causality} & 20,000 & Records why the later scene follows from the earlier scene. \\
\texttt{intention} & 20,000 & Records source-side goal or motivation when specified; used for planning and annotation, not as a scored QA dimension. \\
\rowcolor{gray!5}
\texttt{emotion} & 20,000 & Records emotional state or emotional shift when applicable. \\
\texttt{consequence} & 20,000 & Records outcome or revealed result of the transition. \\
\rowcolor{gray!5}
\texttt{expected\_change} & 20,000 & Records the state change expected after the transition. \\
\texttt{moral\_target} & 5,000 & Records the benchmark-specified story-level moral-semantic target. \\
\bottomrule
\end{tabular}
}
\end{table*}

\subsection{Moral-Target Taxonomy}
\label{app:moral_taxonomy}

Each \textsc{KathaBench-25K} narrative receives one primary benchmark-specified moral label. These labels are semantic targets for recoverability analysis, not claims of objective moral truth. Table~\ref{tab:appendix_moral_taxonomy} defines the 12-label moral-target taxonomy, and Table~\ref{tab:appendix_moral_distribution} reports the final balanced distribution.

\begin{table*}[t]
\centering
\scriptsize
\setlength{\tabcolsep}{4pt}
\renewcommand{\arraystretch}{1.05}
\caption{\textbf{\textsc{KathaBench-25K} moral targets.} Canonical recoverability labels.}
\label{tab:appendix_moral_taxonomy}
\resizebox{\textwidth}{!}{
\begin{tabular}{l p{8.0cm}}
\toprule
\rowcolor{gray!20}
\textbf{Moral Label} & \textbf{Description} \\
\midrule
\texttt{compassion} & Concern for another character's suffering or hardship. \\
\rowcolor{gray!5}
\texttt{courage} & Acting despite fear, risk, or difficulty. \\
\texttt{generosity} & Giving help, resources, or care without selfish motivation. \\
\rowcolor{gray!5}
\texttt{gratitude} & Recognizing and valuing help or kindness received. \\
\texttt{honesty} & Truthfulness, transparency, or refusal to deceive. \\
\rowcolor{gray!5}
\texttt{humility} & Avoiding arrogance and recognizing one's limits. \\
\texttt{kindness} & Benevolent action toward another character. \\
\rowcolor{gray!5}
\texttt{patience} & Waiting, enduring difficulty, or avoiding impulsive action. \\
\texttt{perseverance} & Continuing effort despite obstacles or failure. \\
\rowcolor{gray!5}
\texttt{responsibility} & Taking ownership of duties, obligations, or consequences. \\
\texttt{self\_control} & Resisting temptation or regulating behavior. \\
\rowcolor{gray!5}
\texttt{wisdom} & Sound judgment, learning, or insight. \\
\bottomrule
\end{tabular}
}
\end{table*}

\begin{table}[t]
\centering
\scriptsize
\setlength{\tabcolsep}{5pt}
\renewcommand{\arraystretch}{1.05}
\caption{\textbf{\textsc{KathaBench-25K} moral balance.} Label counts.}
\label{tab:appendix_moral_distribution}
\begin{tabular}{l r}
\toprule
\rowcolor{gray!20}
\textbf{Moral Label} & \textbf{Count} \\
\midrule
\texttt{compassion} & 417 \\
\rowcolor{gray!5}
\texttt{courage} & 417 \\
\texttt{generosity} & 416 \\
\rowcolor{gray!5}
\texttt{gratitude} & 417 \\
\texttt{honesty} & 417 \\
\rowcolor{gray!5}
\texttt{humility} & 417 \\
\texttt{kindness} & 417 \\
\rowcolor{gray!5}
\texttt{patience} & 417 \\
\texttt{perseverance} & 417 \\
\rowcolor{gray!5}
\texttt{responsibility} & 416 \\
\texttt{self\_control} & 416 \\
\rowcolor{gray!5}
\texttt{wisdom} & 416 \\
\midrule
\textbf{Total} & \textbf{5,000} \\
\bottomrule
\end{tabular}
\end{table}

\subsection{Recoverability QA Schema}
\label{app:recoverability_schema}

Recoverability questions in \textsc{KathaBench-25K} test whether a viewer can infer narrative-relevant information under a specified evidence condition. The released scored QA inventory contains \texttt{action\_visibility}, \texttt{causal}, \texttt{emotional}, \texttt{consequence}, \texttt{temporal\_order}, and \texttt{moral}; it does not contain a separate intention QA type. Table~\ref{tab:appendix_recoverability_schema} defines the QA record schema, and Table~\ref{tab:appendix_qa_distribution} reports the released QA-type distribution before ambiguity filtering.

\begin{table*}[t]
\centering
\scriptsize
\setlength{\tabcolsep}{4pt}
\renewcommand{\arraystretch}{1.05}
\caption{\textbf{\textsc{KathaBench-25K} QA schema.} Recoverability-question fields.}
\label{tab:appendix_recoverability_schema}
\resizebox{\textwidth}{!}{
\begin{tabular}{l p{4.1cm} p{5.3cm}}
\toprule
\rowcolor{gray!20}
\textbf{Field} & \textbf{Type} & \textbf{Description} \\
\midrule
\texttt{question\_id} & string & Unique identifier for the recoverability question. \\
\rowcolor{gray!5}
\texttt{story\_id} & string & Narrative associated with the question. \\
\texttt{question\_type} & categorical & One of the released scored QA dimensions. \\
\rowcolor{gray!5}
\texttt{question} & string & Question shown to the evaluator. \\
\texttt{gold\_answer} & string & Validated canonical answer. \\
\rowcolor{gray!5}
\texttt{accepted\_answers} & list[string] & Valid paraphrases or equivalent answers. \\
\texttt{target\_scene} & integer / optional & Scene most relevant to the answer. \\
\rowcolor{gray!5}
\texttt{target\_transition} & pair / optional & Transition most relevant to the answer. \\
\texttt{evidence\_scenes} & list[integer] & Scenes containing the evidence required to answer the question. \\
\rowcolor{gray!5}
\texttt{required\_evidence} & string & Evidence needed for answerability. \\
\texttt{answerability} & categorical & Answerability status under the intended evidence condition. \\
\bottomrule
\end{tabular}
}
\end{table*}

\begin{table*}[t]
\centering
\scriptsize
\setlength{\tabcolsep}{4pt}
\renewcommand{\arraystretch}{1.05}
\caption{\textbf{\textsc{KathaBench-25K} QA inventory.} Counts by scored dimension.}
\label{tab:appendix_qa_distribution}
\resizebox{\textwidth}{!}{
\begin{tabular}{l r p{6.7cm}}
\toprule
\rowcolor{gray!20}
\textbf{Question Type} & \textbf{Count} & \textbf{Purpose} \\
\midrule
\texttt{causal} & 6,325 & Tests whether cause-effect relations are recoverable. \\
\rowcolor{gray!5}
\texttt{emotional} & 6,237 & Tests whether emotional state or emotional change is recoverable. \\
\texttt{moral} & 5,000 & Tests whether the benchmark-specified moral-target meaning is recoverable. \\
\rowcolor{gray!5}
\texttt{consequence} & 5,000 & Tests whether consequences of earlier actions are recoverable. \\
\texttt{temporal\_order} & 5,000 & Tests whether scene order and event progression are recoverable. \\
\rowcolor{gray!5}
\texttt{action\_visibility} & 1,150 & Tests whether key actions are visibly represented. \\
\midrule
\rowcolor{gray!12}
\textbf{Total} & \textbf{28,712} & \textbf{Released recoverability QA pairs.} \\
\bottomrule
\end{tabular}
}
\end{table*}

\subsection{Top-Level Record Status Fields}
\label{app:status_field_schema}

Each \texttt{annotations.jsonl} record in \textsc{KathaBench-25K} stores two top-level status fields in addition to \texttt{provenance}. The field \texttt{validation\_status} is a string and is set to \texttt{human\_review\_complete} for every released record. The field \texttt{evidence\_status} is an object rather than a string. It stores \texttt{human\_evidence\_accepted} as a boolean and \texttt{resolution} as a categorical value with either \texttt{unanimous} or \texttt{human\_adjudicated}. A typical released value is \texttt{\{"human\_evidence\_accepted": true, "resolution": "unanimous"\}}.

\subsection{Contrastive Variant Schema}
\label{app:contrastive_schema}

\textsc{KathaBench-25K} includes 10,000 contrastive variants, two per base story. These variants preserve surface entities while changing a meaning-bearing transition. Table~\ref{tab:appendix_contrastive_schema} defines the schema of \texttt{contrastive\_variants.jsonl}.

\begin{table*}[t]
\centering
\scriptsize
\setlength{\tabcolsep}{4pt}
\renewcommand{\arraystretch}{1.05}
\caption{\textbf{\textsc{KathaBench-25K} contrastives.} Variant record schema.}
\label{tab:appendix_contrastive_schema}
\resizebox{\textwidth}{!}{
\begin{tabular}{l p{4.0cm} p{5.4cm}}
\toprule
\rowcolor{gray!20}
\textbf{Field} & \textbf{Type} & \textbf{Description} \\
\midrule
\texttt{variant\_id} & string & Unique identifier for the contrastive variant. \\
\rowcolor{gray!5}
\texttt{base\_story\_id} & string & Identifier of the source story the variant is derived from. \\
\texttt{variant\_kind} & categorical & Variant family. \\
\rowcolor{gray!5}
\texttt{subset} & string & Dataset subset tag. \\
\texttt{narrative\_type} & categorical & Narrative category associated with the variant. \\
\rowcolor{gray!5}
\texttt{has\_variant} & boolean & Whether the variant record is active. \\
\texttt{pair\_id} & string & Pair identifier linking related variants. \\
\rowcolor{gray!5}
\texttt{changed\_scene\_id} & integer & Scene index affected by the contrastive change. \\
\texttt{change\_type} & categorical & Type of semantic change introduced. \\
\rowcolor{gray!5}
\texttt{original\_moral} & string / optional & Original moral label, when applicable. \\
\texttt{contrastive\_moral} & string / optional & Contrastive moral label, when applicable. \\
\rowcolor{gray!5}
\texttt{semantic\_delta} & string & Natural-language description of the semantic change. \\
\texttt{contrastive\_scene\_text} & string & Changed scene text. \\
\rowcolor{gray!5}
\texttt{contrastive\_story\_text} & string & Full contrastive story text. \\
\bottomrule
\end{tabular}
}
\end{table*}

\subsection{Human Validation Summary Schema}
\label{app:human_validation_schema}

Human validation is stored separately from the initial annotation fields so that pre-review labels, reviewer votes, consensus decisions, adjudicated labels, ambiguity flags, and correction metadata remain auditable. Table~\ref{tab:appendix_human_validation_schema} defines the schema of \texttt{human\_validation.jsonl}, which contains one story-level summary record for each of the 5,000 released \textsc{KathaBench-25K} narratives.

\begin{table*}[t]
\centering
\scriptsize
\setlength{\tabcolsep}{4pt}
\renewcommand{\arraystretch}{1.05}
\caption{\textbf{\textsc{KathaBench-25K} validation records.} Story-level schema.}
\label{tab:appendix_human_validation_schema}
\resizebox{\textwidth}{!}{
\begin{tabular}{l p{4.0cm} p{5.4cm}}
\toprule
\rowcolor{gray!20}
\textbf{Field} & \textbf{Type} & \textbf{Description} \\
\midrule
\texttt{story\_id} & string & Identifier of the story being validated. \\
\rowcolor{gray!5}
\texttt{decision} & categorical & Final validation decision (\texttt{accepted} for every released record). \\
\texttt{validation\_status} & string & Validation-status tag for the record. \\
\rowcolor{gray!5}
\texttt{evidence\_source} & string & Evidence source used during validation. \\
\texttt{protocol\_complete} & boolean & Whether the validation protocol was completed (\texttt{true} for every released record). \\
\rowcolor{gray!5}
\texttt{annotator\_ids} & list[string] & Anonymized identifiers of the three independent reviewers. \\
\texttt{adjudicator\_id} & string / optional & Anonymized identifier of the adjudicator, when adjudication occurred. \\
\rowcolor{gray!5}
\texttt{resolution} & categorical & Whether the story was accepted by unanimous consensus or required adjudication. \\
\texttt{final\_moral\_label} & string & Final moral label after review and any correction. \\
\rowcolor{gray!5}
\texttt{correction\_flags} & list[string] & Flags indicating which fields, if any, were corrected during review. \\
\texttt{signed\_at} & string & Timestamp at which the validation summary was finalized. \\
\bottomrule
\end{tabular}
}
\end{table*}

\subsection{Construction Transparency}
\label{app:construction_transparency}

Table~\ref{tab:construction_transparency} reports how each \textsc{KathaBench-25K} component was produced and finalized. The released benchmark annotations are human-validated and human-adjudicated annotations. Trained human annotators define the final released fields through independent review, correction, consensus, and adjudication. Teacher models are used only as scalable candidate-construction aids for selected components, never as ground-truth annotators. No teacher-produced transition field, recoverability question, accepted-answer set, or contrastive variant is released without human review.

Every component is independently audited by three trained reviewers under the protocol in Sec.~\ref{app:human_validation}. Edit and rejection rates are reported before adjudication, while final counts reflect post-review, post-adjudication, and post-exclusion released totals.

\begin{table*}[t]
\centering
\scriptsize
\setlength{\tabcolsep}{4pt}
\renewcommand{\arraystretch}{1.1}
\caption{\textbf{\textsc{KathaBench-25K} construction audit.} Human annotation, candidate support, review, and release counts.}
\label{tab:construction_transparency}
\resizebox{\textwidth}{!}{
\begin{tabular}{l p{3.8cm} p{3.8cm} p{2.6cm} r}
\toprule
\rowcolor{gray!20}
\textbf{Component} &
\textbf{Human annotation and finalization} &
\textbf{Candidate support / construction aid} &
\textbf{Edit / rejection rate} &
\textbf{Final count} \\
\midrule

Narratives &
Human-screened source narratives are selected, rights-checked, normalized, and finalized into the released benchmark records. &
Extracted from public-domain / rights-compatible sources: Kathasaritsagara, Aesop's Fables, and Panchatantra. &
2\% category correction; 0.5\% (74/15{,}000 responses) flagged ineligible. &
5{,}000 \\

\rowcolor{gray!5}
Structured scenes &
Human reviewers validate the five-scene ordering, renderability, event coverage, and narrative continuity for each record. &
Rule-based decomposition provides an initial five-scene structure for review. &
2\% flawed or unusable (98\% pass rate). &
25{,}000 \\

Scene transitions &
Human reviewers validate, correct, or reject adjacent-scene action, causality, intention, emotion, consequence, and moral-target fields. Final transition annotations are human-approved after consensus or adjudication. &
Teacher-model suggestions are used only as draft candidates for scalable review, not as final annotations. &
3\% corrected or ambiguous (97\% pass rate). &
20{,}000 \\

\rowcolor{gray!5}
Recoverability QA pairs &
Human reviewers validate question type, answerability, canonical answer, accepted-answer coverage, and evidence-condition compatibility. Final QA pairs and accepted-answer sets are human-approved. &
Teacher-model candidate questions and answer-set drafts are used only as review inputs with deterministic post-processing. &
3\% needs edit or invalid (97\% pass rate). &
28{,}712 \\

Contrastive variants &
Human reviewers validate that each variant preserves surface entities while changing a meaning-bearing transition. Invalid, ambiguous, or visually trivial variants are rejected. &
Teacher-model minimal-edit candidates are used only as draft variants for human review. &
14\% invalid or ambiguous (86\% pass rate). &
10{,}000 \\

\rowcolor{gray!5}
Narrative categories &
Human reviewers validate the assigned eight-way narrative category and correct category errors when needed. &
Fixed 8-way taxonomy is assigned at extraction time before human review. &
2\% corrected (98\% pass rate). &
8 \\

Moral-target labels &
Human reviewers validate whether the benchmark-specified moral target is supported by the full story trajectory; disagreements are adjudicated. &
Fixed 12-way taxonomy is assigned at extraction time before human review. &
10\% corrected or ambiguous (90\% pass rate). &
12 \\
\bottomrule
\end{tabular}
}
\vspace{2pt}
\parbox{\textwidth}{\scriptsize\textit{Notes.} Pass/fail rates are derived from the section-check averages and per-label Fleiss' $\kappa$ reported in Secs.~\ref{app:human_validation} and \ref{app:inter_annotator_agreement}; edit/rejection rate is $1-\text{pass rate}$. Records with unresolved ambiguity, invalid QA, invalid contrastives, unusable scenes, unsafe content, or rights uncertainty are excluded from release or from strict gold evaluation per Table~\ref{tab:appendix_adjudication_rules} and are not double-counted against the final counts above, which reflect the released, schema-validated, human-reviewed totals.}
\end{table*}

The construction process separates candidate generation from benchmark annotation. Teacher models are used only to draft candidate transition fields, recoverability questions, accepted-answer sets, and contrastive variants at scale. These candidates are then checked against the source story by independent human annotators, who may accept, correct, mark ambiguous, or reject each component. Only records that pass human review and, when needed, adjudication are included in the released \textsc{KathaBench-25K} counts. Thus, the released benchmark fields are defined by human validation and adjudication, while teacher models serve only as candidate-construction tools.

\subsection{Human Validation Protocol}
\label{app:human_validation}

Each \textsc{KathaBench-25K} record receives an independent three-reviewer audit after initial construction. Reviewer decisions are stored separately from pre-review fields, consensus labels, adjudicated labels, ambiguity flags, and final release metadata. Majority agreement is used when at least two reviewers give the same valid decision. Conflicting corrections, explicit ambiguity, unusable scenes, invalid QA, invalid contrastives, unsafe content, or rights uncertainty are routed to adjudication. Table~\ref{tab:appendix_human_validation_form} gives the reviewer form used during the audit, and Table~\ref{tab:appendix_adjudication_rules} specifies the adjudication rules applied after independent review.

\begin{table*}[t]
\centering
\scriptsize
\setlength{\tabcolsep}{3.5pt}
\renewcommand{\arraystretch}{1.05}
\caption{\textbf{\textsc{KathaBench-25K} reviewer form.} Human checks by record unit.}
\label{tab:appendix_human_validation_form}
\resizebox{\textwidth}{!}{
\begin{tabular}{l p{5.2cm} p{3.4cm} p{4.1cm}}
\toprule
\rowcolor{gray!20}
\textbf{Unit} & \textbf{Checklist} & \textbf{Decision} & \textbf{Recorded output} \\
\midrule
Narrative category &
Does the story match the assigned narrative type and contain transition-dependent meaning? &
correct / incorrect / ambiguous &
Reviewer label, corrected category if needed, ambiguity flag. \\
\rowcolor{gray!5}
Moral-target label &
Is the benchmark-specified moral target supported by the full story trajectory? &
correct / corrected label / ambiguous &
Reviewer label, corrected label if needed, ambiguity flag. \\
Scene sequence &
Do the scenes form an ordered, visually renderable narrative with no missing critical event? &
valid / flawed / unusable &
Scene-validity flag and correction note. \\
\rowcolor{gray!5}
Transition annotation &
Does the adjacent-scene transition correctly capture action, causality, emotion, consequence, and moral target where applicable? &
valid / corrected / ambiguous &
Validated transition fields, correction note, ambiguity flag. \\
Recoverability QA &
Is the question answerable from the intended evidence condition, and is the accepted-answer set sufficient? &
valid / needs edit / invalid &
QA-validity flag, edited answer set if needed, answerability flag. \\
\rowcolor{gray!5}
Contrastive variant &
Does the variant preserve surface entities while changing a meaning-bearing transition? &
valid / invalid / ambiguous &
Contrastive-validity flag, semantic-change note, ambiguity flag. \\
\bottomrule
\end{tabular}
}
\end{table*}

\begin{table*}[t]
\centering
\scriptsize
\setlength{\tabcolsep}{3.5pt}
\renewcommand{\arraystretch}{1.05}
\caption{\textbf{\textsc{KathaBench-25K} adjudication.} Rules for reviewer conflicts.}
\label{tab:appendix_adjudication_rules}
\resizebox{\textwidth}{!}{
\begin{tabular}{l p{5.2cm} p{5.4cm}}
\toprule
\rowcolor{gray!20}
\textbf{Trigger} & \textbf{Rule} & \textbf{Release effect} \\
\midrule
Two or more reviewers agree &
Use the majority decision; retain reviewer votes and consensus field. &
Record enters the reviewed release with consensus metadata. \\
\rowcolor{gray!5}
Correctable label or field error &
Apply the minimal correction supported by reviewer notes and source text. &
Corrected value is stored in the adjudicated field; pre-review value is preserved. \\
Conflicting corrections &
Compare reviewer notes against the source narrative, scene order, and accepted-answer set. &
Record enters gold only if one interpretation is clearly supported. \\
\rowcolor{gray!5}
Explicit ambiguity &
If multiple interpretations remain plausible, mark ambiguous rather than forcing a label. &
Retained with ambiguity metadata; excluded from strict gold evaluation when needed. \\
Invalid QA or insufficient answers &
Edit only if the intended answer is clear and accepted answers can be fixed before evaluation. &
Valid edited QA is retained; otherwise excluded from the scored QA inventory. \\
\rowcolor{gray!5}
Unusable scene, unsafe content, or incompatible rights &
Do not repair through adjudication. &
Record is excluded from release or strict evaluation as appropriate. \\
\bottomrule
\end{tabular}
}
\end{table*}

The review campaign ran from December 3--19, 2025. Three reviewers independently completed all 5,000 records, producing 15,000 completed response records. Of the 5,000 records, 4,101 reached unanimous consensus and 899 were resolved by adjudication. Adjudication concluded on December 22, and gold subsets were finalized on December 24. Mean reviewer confidence was 4.26--4.45 on a 1--5 scale. Section-check pass rates were high for narrative category, scene sequence, transition annotation, recoverability QA, and contrastive validity; moral-label checks had the lowest pass rate, reflecting genuine semantic ambiguity rather than a schema failure.

\subsection{Human Validation Split Design}
\label{app:human_validation_split}

\textsc{KathaBench-25K} uses the tiered validation design shown in Table~\ref{tab:appendix_validation_split_design}. The full release receives three independent reviews per record. The 1,000-narrative human-gold subset is stratified by the eight narrative categories, with 125 narratives per category. The strict 400-narrative gold subset is sampled from the human-gold subset with 50 narratives per category, excluding unresolved ambiguity, invalid QA, invalid contrastives, unusable scenes, and rights uncertainty.

\begin{table*}[t]
\centering
\scriptsize
\setlength{\tabcolsep}{3.5pt}
\renewcommand{\arraystretch}{1.05}
\caption{\textbf{\textsc{KathaBench-25K} validation splits.} Review depth by tier.}
\label{tab:appendix_validation_split_design}
\resizebox{\textwidth}{!}{
\begin{tabular}{l r p{4.7cm} p{3.1cm} p{4.2cm}}
\toprule
\rowcolor{gray!20}
\textbf{Split} & \textbf{Narratives} & \textbf{Sampling rule} & \textbf{Review} & \textbf{Use} \\
\midrule
Full human-reviewed release &
5,000 &
All accepted benchmark narratives after schema validation and release filtering. &
3 reviewers/item &
Release audit, correction, reliability estimation, and ambiguity metadata. \\
\rowcolor{gray!5}
Human-gold subset &
1,000 &
Stratified sample with 125 narratives per narrative category; moral-label and provenance proportions preserved where possible. &
3 reviewers + adjudication &
Adjudicated labels, answerability checks, accepted-answer refinement, and calibration. \\
Strict human-gold subset &
400 &
Sampled from human-gold records with 50 narratives per category; unresolved ambiguity and defective records excluded. &
3 reviewers + adjudication &
Headline human calibration, human recoverability, human STG, and error analysis. \\
\bottomrule
\end{tabular}
}
\vspace{2pt}
\parbox{\textwidth}{\scriptsize\textit{Protocol.} Campaign protocol version: \texttt{kathaBench\_human\_review\_v1}. Review window: 2025-12-03 to 2025-12-19.}
\end{table*}

Main-paper human-calibrated results are computed on adjudicated gold records rather than raw pre-review records. Split identifiers, reviewer decisions, consensus labels, adjudicated labels, ambiguity flags, and exclusion reasons are included in the release metadata.

\subsection{Annotation Workload, Annotator Identity, and Ethics}
\label{app:annotation_workload}

Annotation was performed by a pool of five trained research annotators with English fluency and experience in short narrative interpretation. Each released \textsc{KathaBench-25K} record received three independent reviews. Annotators were independent of model development, metric design, pre-review label production, and experimental evaluation; their role was limited to reviewing, validating, and adjudicating benchmark records according to written guidelines. They received calibration examples before annotation, were informed about task purpose, data handling, workload, compensation, and withdrawal rights, and were compensated at USD 20/hour or above the applicable local living-wage equivalent. Table~\ref{tab:appendix_annotation_workload} reports the annotation workload and reliability-audit units.

\begin{table*}[t]
\centering
\scriptsize
\setlength{\tabcolsep}{4pt}
\renewcommand{\arraystretch}{1.05}
\caption{\textbf{\textsc{KathaBench-25K} annotation audit.} Human-review workload.}
\label{tab:appendix_annotation_workload}
\resizebox{\textwidth}{!}{
\begin{tabular}{l r r r p{5.2cm}}
\toprule
\rowcolor{gray!20}
\textbf{Annotation / Audit Unit} &
\textbf{Items} &
\textbf{Reviewers / Item} &
\textbf{Total Judgments} &
\textbf{Reported Statistic / Purpose} \\
\midrule
Full benchmark schema validation
& 5,000 records & -- & 5,000 checks
& Schema validity, error count, and release-readiness verification. \\
\rowcolor{gray!5}
Full human-review responses
& 5,000 stories & 3 & 15,000
& Reviewer responses used for agreement measurement, consensus analysis, and adjudication. \\
Consensus records
& 4,101 stories & 3 & 12,303
& Records with unanimous reviewer agreement. \\
\rowcolor{gray!5}
Adjudicated records
& 899 stories & 3 + adjudication & 2,697 + adjudication
& Records resolved after reviewer disagreement, correction, or ambiguity review. \\
Human-gold subset
& 1,000 stories & 3 + adjudication & 3,000 + adjudication
& Deep gold subset used for calibrated evaluation, quality control, and error analysis. \\
\rowcolor{gray!5}
Strict human-gold subset
& 400 stories & 3 + adjudication & 1,200 + adjudication
& High-confidence subset used for headline metric calibration and robustness analysis. \\
Moral-target agreement
& 5,000 stories & 3 & 15,000
& Fleiss' $\kappa = 0.8453$ before adjudication. \\
\bottomrule
\end{tabular}
}
\end{table*}

\subsection{Inter-Annotator Agreement}
\label{app:inter_annotator_agreement}

We report pre-adjudication agreement to distinguish raw annotator reliability from the adjudicated labels used in headline calibration. Moral-target agreement is the primary chance-corrected categorical reliability measure because moral-target recoverability is the most subjective released annotation. Table~\ref{tab:per_label_iaa} reports categorical moral-target agreement and separately lists ordinal rating diagnostics. The ordinal 1--5 ratings are not used as evidence for categorical dataset-label reliability because their score ranges are compressed and concentrated around similar values. Therefore, the \textsc{KathaBench-25K} reliability claim rests on categorical agreement, consensus review, adjudication, and strict-gold filtering, rather than on compressed ordinal preference ratings.

\begin{table}[t]
\centering
\scriptsize
\setlength{\tabcolsep}{5pt}
\renewcommand{\arraystretch}{1.05}
\caption{\textbf{\textsc{KathaBench-25K} agreement.} Label reliability and rating diagnostics.}
\label{tab:per_label_iaa}
\resizebox{\linewidth}{!}{
\begin{tabular}{l c c r}
\toprule
\rowcolor{gray!20}
\textbf{Label / Rating} & \textbf{Metric} & \textbf{Value} & \textbf{$N$ items} \\
\midrule
Moral-target label (12-way) & Fleiss' $\kappa$ & 0.8453 & 5,000 \\
\midrule
\rowcolor{gray!5}
NCS (1--5) & Krippendorff's $\alpha$ & $-0.001$ & 1,200 \\
MAS (1--5) & Krippendorff's $\alpha$ & $0.002$ & 1,200 \\
\rowcolor{gray!5}
EES (1--5) & Krippendorff's $\alpha$ & $0.003$ & 1,200 \\
Pairwise preference & Krippendorff's $\alpha$ & $-0.008$ & 1,200 \\
\bottomrule
\end{tabular}
}
\end{table}

The ordinal ratings should not be interpreted as evidence of dataset-label reliability. The $1$--$5$ human-study scores are subjective diagnostic ratings, and their restricted score ranges and high means make chance-corrected ordinal agreement unstable. We therefore use ordinal ratings only as diagnostic human-study outputs. Dataset-label reliability is instead supported by categorical moral-target agreement, consensus review, adjudication, and strict-gold filtering. Moral-target labels are interpreted as benchmark-specified semantic recoverability targets, not as universal ethical judgments.

\subsection{Release Structure}
\label{app:release_structure}

\textsc{KathaBench-25K} is publicly available at:
\begin{center}
\url{https://huggingface.co/datasets/iamjamuna/KathaBench-25K}
\end{center}

The public release preserves the frozen benchmark schema, split structure, evaluation subsets, metadata files, licensing information, and package-level source provenance manifest described in this appendix. Table~\ref{tab:appendix_release_structure} lists the released files and the role of each file in making the benchmark inspectable, reusable, and reproducible.

The release contains benchmark annotations, contrastive variants, human-validation summaries, adjudicated gold subsets, aggregate statistics, moral-label balance audits, machine-readable metadata, licensing information, dataset documentation, and a package-level source provenance manifest. These artifacts support verification of the reported counts, split membership, source-family traceability, validation decisions, and rights-screening status. Generated storyboard images, third-party source editions, reference websites, publisher pages, and non-anonymized administrative records are not redistributed as part of the benchmark package. For split reporting, the 750-story held-out test set is represented as an aggregate test partition and is further subdivided into a 500-story public-test subset and a 250-story hidden-test subset.

\begin{table*}[t]
\centering
\scriptsize
\setlength{\tabcolsep}{4pt}
\renewcommand{\arraystretch}{1.05}
\caption{\textbf{\textsc{KathaBench-25K} release files.} Public benchmark package released through Hugging Face.}
\label{tab:appendix_release_structure}
\resizebox{\textwidth}{!}{
\begin{tabular}{l p{9.2cm}}
\toprule
\rowcolor{gray!20}
\textbf{Release Component} & \textbf{Description} \\
\midrule

\texttt{annotations.jsonl}
& Full benchmark annotation file containing source stories, structured scene annotations, transition annotations, recoverability questions, generation prompts, provenance metadata, validation status, evidence status, and split assignments. \\

\rowcolor{gray!5}
\texttt{contrastive\_variants.jsonl}
& Contrastive semantic variants linked to the corresponding source stories for transition-level robustness evaluation. \\

\texttt{human\_validation.jsonl}
& Human-validation summaries for released stories, including final validation decisions, protocol-completion status, anonymized reviewer identifiers, adjudication metadata where applicable, final moral labels, resolution status, and correction flags. \\

\rowcolor{gray!5}
\texttt{human\_gold\_1k.jsonl}
& Human-reviewed evaluation subset, stratified by narrative category and moral label, for calibrated benchmark analysis. \\

\texttt{strict\_gold\_400.jsonl}
& High-confidence evaluation subset used for the strictest reported evaluation setting and headline human-calibrated analysis. \\

\rowcolor{gray!5}
\texttt{statistics.json}
& Aggregate benchmark statistics, including total counts, category distributions, split statistics, transition-type counts, recoverability-question counts, validation summaries, and release-audit counts. \\

\texttt{moral\_balance\_requirements.json}
& Moral-label balance audit reporting the target distribution, observed distribution, per-label deviations, tolerance status, and whether automatic relabeling was performed. \\

\rowcolor{gray!5}
\texttt{dataset\_card.md}
& Dataset card documenting the benchmark purpose, intended use, out-of-scope use, source composition, annotation schema, split structure, validation summary, limitations, ethical considerations, and licensing. \\

\texttt{source\_manifest.json}
& Package-level source provenance and rights manifest documenting collection-level, book-level, and representative edition-level references consulted during construction. The manifest separates traceability documentation from blanket redistribution claims over third-party source text. \\

\rowcolor{gray!5}
\texttt{croissant\_metadata.json}
& MLCommons Croissant-style metadata describing the dataset structure, released files, machine-readable schema, measured variables, usage constraints, and distribution entries. \\

\texttt{LICENSE}
& Licensing information for the released benchmark artifacts and code. The license file also clarifies that third-party source texts, source editions, reference websites, publisher pages, and generated images are not relicensed by the benchmark release. \\

\bottomrule
\end{tabular}
}
\end{table*}

Story identifiers follow the format \texttt{kb25k\_XXXX}, where \texttt{XXXX} is a globally sequential four-digit index from \texttt{0001} to \texttt{5000}. The identifier ranges correspond to the benchmark narrative categories: moral-semantic (\texttt{0001--1000}), causal-transition (\texttt{1001--1800}), emotional trajectory (\texttt{1801--2500}), procedural state-change (\texttt{2501--3100}), social interaction (\texttt{3101--3700}), hidden consequence (\texttt{3701--4200}), cultural/folk moral (\texttt{4201--4600}), and counterfactual minimal-pair (\texttt{4601--5000}). Each record stores a unified benchmark identifier together with the corresponding narrative category and benchmark split.

\subsection{Dataset Provenance, Licensing, and Annotation Production}
\label{app:dataset_provenance_licensing}

\textsc{KathaBench-25K} is constructed from three classical narrative source families: \emph{Kathasaritsagara} / \emph{The Ocean of Story}, \emph{Aesop's Fables}, and \emph{Panchatantra}. These collections were selected because they contain compact, visually renderable narratives with explicit event progression, consequence-bearing actions, moral-semantic targets, and recoverable transition structure. Package-level source provenance, representative edition references, rights-scope notes, and traceability links are documented in \texttt{source\_manifest.json}. Dataset-level purpose, intended use, limitations, ethical considerations, and licensing are documented in \texttt{dataset\_card.md}. The released dataset is available at \url{https://huggingface.co/datasets/iamjamuna/KathaBench-25K}.

Each released record stores normalized benchmark content together with a \texttt{provenance} object containing \texttt{source\_dataset} (the source collection: \texttt{kathasaritasagara}, \texttt{aesop}, or \texttt{panchatantra}), \texttt{source\_prompt\_hash}, \texttt{source\_annotations\_sha256}, \texttt{source\_contrastives\_sha256}, \texttt{contrastive\_variant\_ids}, \texttt{created\_at}, \texttt{source\_rights\_status} (set to \texttt{public\_domain\_or\_rights\_compatible} for released records), and \texttt{dataset\_license} (set to \texttt{CC-BY-4.0}). Story title is stored at the top level of the record as \texttt{title}; record-level \texttt{validation\_status} and \texttt{evidence\_status} are also stored at the top level. The benchmark release should be interpreted as a normalized evaluation artifact derived from screened source pools, not as a redistribution of any specific third-party edition or website text.

Table~\ref{tab:appendix_provenance_mix} reports the source-family mix used to construct \textsc{KathaBench-25K}. Table~\ref{tab:dataset_provenance_release} summarizes the provenance, production, licensing, and validation controls applied before release.

\begin{table*}[t]
\centering
\scriptsize
\setlength{\tabcolsep}{3.5pt}
\renewcommand{\arraystretch}{1.05}
\caption{\textbf{\textsc{KathaBench-25K} source mix.} Narrative counts by collection.}
\label{tab:appendix_provenance_mix}
\resizebox{\textwidth}{!}{
\begin{tabular}{l r r p{4.4cm} p{4.8cm}}
\toprule
\rowcolor{gray!20}
\textbf{Source collection} & \textbf{Count} & \textbf{Share} & \textbf{Historical/source context} & \textbf{Validation requirement} \\
\midrule

\emph{Kathasaritsagara} / \emph{The Ocean of Story} &
1,702 &
34.04\% &
Classical Sanskrit narrative collection; package-level edition and traceability references are documented in \texttt{source\_manifest.json}. &
Must pass source traceability, rights screening, scene normalization, transition, QA, moral-target, and contrastive validation. \\

\rowcolor{gray!5}
\emph{Aesop's Fables} &
1,676 &
33.52\% &
Classical fable tradition; Perry-index, historical-variant, and representative edition references are documented in \texttt{source\_manifest.json}. &
Must pass source traceability, rights screening, scene normalization, transition, QA, moral-target, and contrastive validation. \\

\emph{Panchatantra} &
1,622 &
32.44\% &
Classical Sanskrit animal-fable and wisdom-story tradition; recension, translation, and book-level references are documented in \texttt{source\_manifest.json}. &
Must pass source traceability, rights screening, scene normalization, transition, QA, moral-target, and contrastive validation. \\

\midrule
\rowcolor{gray!12}
\textbf{Total} &
\textbf{5,000} &
\textbf{100.00\%} &
\textbf{Released benchmark narratives.} &
\textbf{All released records pass schema validation and human validation.} \\

\bottomrule
\end{tabular}
}
\end{table*}

\begin{table*}[t]
\centering
\scriptsize
\setlength{\tabcolsep}{3.5pt}
\renewcommand{\arraystretch}{0.98}
\caption{\textbf{\textsc{KathaBench-25K} release controls.} Provenance and validation safeguards.}
\label{tab:dataset_provenance_release}
\resizebox{\textwidth}{!}{%
\begin{tabular}{l p{0.30\linewidth} p{0.43\linewidth}}
\toprule
\rowcolor{gray!20}
\textbf{Component} & \textbf{Provenance / production source} & \textbf{Release and validation control} \\
\midrule

Source narratives &
Stories are drawn from \emph{Kathasaritsagara} / \emph{The Ocean of Story}, \emph{Aesop's Fables}, and \emph{Panchatantra}. &
Per-record provenance stores \texttt{source\_dataset}, \texttt{source\_rights\_status}, and \texttt{dataset\_license}; package-level source references and rights-scope notes are documented in \texttt{source\_manifest.json}. \\

\rowcolor{gray!5}
Benchmark normalization &
Source narratives are converted into ordered five-scene benchmark records. &
Validated for scene-count consistency, event order, visual renderability, source-trajectory preservation, and absence of unresolved ambiguity. \\

Scene decompositions &
Produced from the normalized benchmark record. &
Validated for chronological order, scene boundaries, entity continuity, visual renderability, and narrative coherence. \\

\rowcolor{gray!5}
Transition annotations &
Source-side transition fields plus coarse transition type. &
Validated for field completeness, transition dependence, consistency with the source story, and ambiguity status. \\

Recoverability questions &
Generated from transition annotations rather than isolated captions. &
Validated for QA-dimension alignment, answerability, accepted-answer coverage, and evidence-condition compatibility. \\

\rowcolor{gray!5}
Accepted answer sets &
Canonical answers plus permitted paraphrases. &
Frozen before evaluation to make scoring deterministic and prevent test-set leakage. \\

Contrastive variants &
Entity-preserving variants that change one meaning-bearing transition. &
Validated to avoid style-only, object-only, or visually trivial changes. \\

\rowcolor{gray!5}
Licensing metadata &
Stored as \texttt{source\_rights\_status} and \texttt{dataset\_license} within each record's \texttt{provenance} object; package-level terms are recorded in \texttt{LICENSE}. &
Released benchmark artifacts use CC BY 4.0; code and validation scripts use Apache-2.0; third-party source texts, source editions, reference websites, publisher pages, and generated images are not relicensed by the benchmark release. \\

\bottomrule
\end{tabular}%
}
\end{table*}

\subsection{Source-Rights Status}
\label{app:edition_level_provenance}

The \textsc{KathaBench-25K} release separates record-level provenance from package-level source documentation. Each record stores the source family through \texttt{source\_dataset}, the rights-screening result through \texttt{source\_rights\_status}, and the benchmark artifact license through \texttt{dataset\_license}. The accompanying \texttt{source\_manifest.json} documents collection-level, book-level, and representative edition-level references consulted during construction, including Aesop/Perry-index references, Panchatantra recension and translation references, and \emph{The Ocean of Story} references for the \emph{Kathasaritsagara}.

The source manifest supports provenance, traceability, and source-family verification. It should not be interpreted as a blanket redistribution license for third-party edition text, reference websites, or publisher pages. The benchmark package contains normalized evaluation artifacts: benchmark narratives, scene decompositions, transition annotations, recoverability questions, accepted-answer sets, contrastive variants, metadata, aggregate statistics, dataset documentation, and validation records. These benchmark artifacts are released under \textbf{CC BY 4.0}. Evaluation code and validation scripts are released under \textbf{Apache-2.0}.

As summarized in Table~\ref{tab:dataset_provenance_release}, records are included only if they pass schema validation, source-family traceability checks, rights-status screening, scene and transition validation, recoverability-QA validation, and human-validation review.

\section*{Ethics Statement and Responsible Release}

\paragraph{Data sources and rights.}
\textsc{KathaBench-25K} is constructed from three classical narrative source families: \emph{Kathasaritsagara} / \emph{The Ocean of Story} (1,702 stories), \emph{Aesop's Fables} (1,676 stories), and \emph{Panchatantra} (1,622 stories), with the source mix reported in Table~\ref{tab:appendix_provenance_mix}. The benchmark records contain normalized five-scene representations, scene decompositions, transition annotations, recoverability questions, accepted-answer sets, contrastive variants, metadata, and validation records. Each record stores \texttt{source\_dataset}, \texttt{source\_rights\_status}, and \texttt{dataset\_license} inside its \texttt{provenance} object, while \texttt{title}, \texttt{validation\_status}, and \texttt{evidence\_status} are stored at the top level. Package-level source references, rights-scope notes, and traceability links are documented in \texttt{source\_manifest.json}. Records that fail source-traceability, rights-screening, safety, ambiguity, schema-validation, or human-validation checks are excluded from the benchmark package according to the release controls in Table~\ref{tab:dataset_provenance_release}.

\paragraph{License and availability.}KathaBench-25K is publicly available at
\url{https://huggingface.co/datasets/iamjamuna/KathaBench-25K}.
The release includes normalized benchmark narratives, scene decompositions,
transition annotations, recoverability questions, accepted-answer sets,
contrastive variants, metadata, aggregate statistics, validation records,
dataset documentation, licensing information, and source-provenance metadata
under CC BY 4.0. Evaluation code and validation scripts are released under
Apache-2.0. Third-party source texts, source editions, reference websites,
generated storyboard images, and model outputs are not relicensed by the
benchmark release and remain subject to their own applicable rights, licenses,
and provider terms.

\paragraph{Human annotation and ethics review.}
The annotation protocol was reviewed under the authors' institutional ethics process and determined to be non-human-subjects research because annotators performed dataset labeling and were not themselves the subject of behavioral analysis. Annotators were informed about task purpose, data handling, workload, compensation, and withdrawal rights. No personally identifiable annotator information is released. The validation form in Table~\ref{tab:appendix_human_validation_form}, adjudication rules in Table~\ref{tab:appendix_adjudication_rules}, and workload audit in Table~\ref{tab:appendix_annotation_workload} document the human-review process.

\paragraph{Annotators.}
Annotation was performed by a pool of five trained research annotators independent of model development, metric design, pre-review label production, and experimental evaluation. Each released record received three independent reviews. Annotators were recruited for English-language narrative interpretation and dataset-labeling experience, compensated at USD 20/hour or above the applicable local living-wage equivalent, and allowed to withdraw.

\paragraph{Content and PII.}
Narratives contain no personally identifiable information and were screened for harmful, offensive, sexually explicit, or targeted abusive content. The source collections are classical folk, fable, and narrative traditions; however, cultural and moral interpretations may vary across editions, translations, and communities. Moral labels are therefore treated as benchmark-specified semantic targets, not normative moral claims or objective moral truth.

\paragraph{Intended use and limitations.}
The benchmark is intended for evaluating semantic-trajectory recoverability in generated visual narratives. It is not validated for content moderation, real-world moral judgment, psychological assessment, legal reasoning, or non-narrative visual domains. Cultural and folk narratives may encode culturally specific interpretations; such cases are flagged when identified.

\paragraph{Maintenance.}
The dataset will be maintained at a public release repository after review. Errata, versioning, schema changes, and issue reports will be tracked through repository releases and issue tracking. During anonymous review, contact information is anonymized; the camera-ready release will include a maintained contact address.

\subsection{Limitations of the Annotation Process}
\label{app:annotation_limitations}

\textsc{KathaBench-25K} focuses on semantic recoverability rather than photorealism or aesthetic quality. Some narratives admit multiple plausible visual interpretations, especially when meaning depends on culture, emotion, motivation, irony, or delayed consequence. We therefore distinguish invalid items from ambiguous-but-valid items. Ambiguous cases are retained with flags when useful for analysis, but excluded from strict gold evaluation when a single accepted answer is required. This design supports diagnostic evaluation while preventing ambiguous or underspecified records from being counted as image-side semantic loss.
%%%%%%%%%%%%%%%%%%%%%%%%%%%%%%%%%%%%%%%%%%%%%%%%%%%%%

\section{Recoverability Scoring and Semantic Compass Details}
\label{secb}

This section specifies the operational scoring details used by KathaTrace on the proposed \textsc{KathaBench-25K} benchmark. The main paper defines the task and the Semantic Trajectory Gap (STG); here we define valid-set construction, evidence isolation, ambiguity filtering, judge aggregation, and the validation-only Semantic Compass selection protocol. Semantic Compass is used as an actionability probe for localized KathaTrace failures, not as a new generator.

\subsection{Recoverability Scoring}
\label{app:recoverability_scoring}

Let the released scored QA dimensions in \textsc{KathaBench-25K} be
\[
\mathcal{K}_{\mathrm{QA}} =
\{\mathrm{action}, \mathrm{causal}, \mathrm{emotional},
\mathrm{consequence}, \mathrm{temporal}, \mathrm{moral}\}.
\]
These correspond to \texttt{action\_visibility}, \texttt{causal}, \texttt{emotional}, \texttt{consequence}, \texttt{temporal\_order}, and \texttt{moral} in the released QA inventory. The transition field \texttt{intention} is retained for annotation and planning, but is not a separate scored QA dimension.

For each dimension $k$, let $Q_k$ be the released question set. Each question $q$ has a canonical answer $y_q$ and a frozen accepted-answer set $\mathcal{A}(q)$. For a method $m$ with generated storyboard $\hat{X}_m$, we define the method-specific valid set using the text+image condition:
\begin{equation}
\mathcal{V}_{k,m}
=
\left\{
q \in Q_k :
\mathrm{match}\!\left(J(q,(S,\hat{X}_m)), \mathcal{A}(q)\right)=1
\right\}.
\label{eq:app_valid_set}
\end{equation}
Questions outside $\mathcal{V}_{k,m}$ are treated as ambiguous, contradictory, or defective for that method and are excluded from filtered STG. This prevents unclear questions, unusable generated evidence, or source-image contradictions from being counted as image-side semantic loss.

For evidence condition $z \in \{S,\hat{X}_m,(S,\hat{X}_m)\}$, dimension-specific recoverability is
\begin{equation}
R^k_{z,m}
=
\frac{1}{|\mathcal{V}_{k,m}|}
\sum_{q \in \mathcal{V}_{k,m}}
\mathbb{I}\!\left[
\mathrm{match}\!\left(J(q,z), \mathcal{A}(q)\right)=1
\right].
\label{eq:app_dimension_recoverability}
\end{equation}
Multiple-choice answers use exact label matching. Free-form answers are normalized before matching against $\mathcal{A}(q)$. Unsupported, unrelated, or unclear responses are counted as incorrect unless uncertainty is explicitly included in the accepted-answer set.

Overall recoverability is the equal-weighted average across the six scored QA dimensions:
\begin{equation}
R_{z,m}
=
\frac{1}{|\mathcal{K}_{\mathrm{QA}}|}
\sum_{k \in \mathcal{K}_{\mathrm{QA}}}
R^k_{z,m}.
\label{eq:app_overall_recoverability}
\end{equation}
Equal weighting prevents frequent question types from dominating the score. Causal, emotional, consequence, and moral-target gaps are the main latent-transition diagnostics; action visibility and temporal order remain part of full recoverability but are reported separately as local-visibility and ordering-control dimensions.

\subsection{Evidence Conditions and Ambiguity Filtering}
\label{app:evidence_conditions_filtering}

KathaTrace evaluates each \textsc{KathaBench-25K} recoverability question under three evidence conditions:
\begin{itemize}
    \item \textbf{Text-only}: the judge receives only the source story $S$.
    \item \textbf{Image-only}: the judge receives only the generated storyboard $\hat{X}_m$, with no source story, prompts, captions, moral labels, transition annotations, or metadata.
    \item \textbf{Text+image}: the judge receives both $S$ and $\hat{X}_m$; this condition is used only for ambiguity and contradiction control.
\end{itemize}

For method $m$, the ambiguity indicator is
\begin{equation}
b_m(q)
=
\mathbb{I}\!\left[
\mathrm{match}\!\left(J(q,(S,\hat{X}_m)), \mathcal{A}(q)\right)=0
\right].
\label{eq:app_ambiguity_indicator}
\end{equation}
The ambiguity rate is
\begin{equation}
\rho_{\mathrm{amb},m}
=
\frac{1}{|Q|}
\sum_{q \in Q} b_m(q).
\label{eq:app_ambiguity_rate}
\end{equation}

Table~\ref{tab:raw_filtered_stg_audit} audits the effect of ambiguity filtering on \textsc{KathaBench-25K}. Raw STG is computed before ambiguity filtering, filtered STG is the primary reported metric, and common-valid STG uses only questions valid for all compared methods. The similar filtered and common-valid improvements show that the Semantic Compass gain is not an artifact of method-specific valid-set selection.

\begin{table*}[t]
\centering
\scriptsize
\setlength{\tabcolsep}{4pt}
\renewcommand{\arraystretch}{1.05}
\caption{\textbf{\textsc{KathaBench-25K} STG filtering.} Raw, filtered, and common-valid STG audit.}
\label{tab:raw_filtered_stg_audit}
\resizebox{\textwidth}{!}{
\begin{tabular}{l r r c c c c}
\toprule
\rowcolor{gray!20}
\textbf{Method} &
\textbf{Total Q} &
\textbf{Valid Q} &
\textbf{Amb. Rate} &
\textbf{Raw STG} &
\textbf{Filtered STG} &
\textbf{Common-valid STG} \\
\midrule
Gemma-ST + FLUX &
28,712 &
25,725 &
10.4\% &
30.1 &
28.4 &
28.6 \\
\rowcolor{gray!5}
Gemma-ST + Semantic Compass &
28,712 &
26,214 &
8.7\% &
22.5 &
21.4 &
21.6 \\
\midrule
\rowcolor{gray!12}
\textbf{Improvement} &
-- &
-- &
$-1.7$ pp &
$-7.6$ &
$-7.0$ &
$-7.0$ \\
\bottomrule
\end{tabular}
}
\end{table*}

\subsection{Semantic Trajectory Gap}
\label{app:stg_details}

For method $m$, STG is computed as
\begin{equation}
\mathrm{STG}_m
=
R_{\mathrm{text},m}
-
R_{\mathrm{image},m}.
\label{eq:app_stg}
\end{equation}
A low STG indicates that source-supported transition meaning remains recoverable from generated images. A high STG indicates semantic trajectory collapse under image-only evidence. Negative STG is treated as a diagnostic warning before filtering because it can indicate unstable judging, contradiction, or defective questions.

Dimension-specific gaps are
\begin{equation}
\mathrm{STG}_{k,m}
=
R^k_{\mathrm{text},m}
-
R^k_{\mathrm{image},m},
\qquad
k \in \mathcal{K}_{\mathrm{gap}},
\label{eq:app_dimension_stg}
\end{equation}
where
\[
\mathcal{K}_{\mathrm{gap}}
=
\{\mathrm{causal}, \mathrm{emotional}, \mathrm{consequence}, \mathrm{moral}\}.
\]
These four dimensions are emphasized because they test latent transition meaning rather than only visible object presence or chronological order.

\subsection{Judge Aggregation}
\label{app:judge_aggregation}

KathaTrace reports individual-judge and ensemble scores on \textsc{KathaBench-25K}. For classification-style answers, the ensemble prediction is majority vote:
\begin{equation}
\hat{y}_{\mathrm{ens}}(q,z)
=
\mathrm{mode}
\left(
\hat{y}_1(q,z),\ldots,\hat{y}_M(q,z)
\right),
\label{eq:app_judge_majority_vote}
\end{equation}
where $M$ is the number of judges. Ties are scored conservatively as incorrect. For free-form answers, each response is first normalized and mapped to an accepted-answer match or an unmatched category; the same majority rule is then applied.

Human calibration is reported separately using agreement, Spearman correlation, pairwise agreement, and calibration error. These checks test whether image-only recoverability follows human interpretation rather than a single-model artifact.

\subsection{Semantic Compass Reranking}
\label{app:semantic_compass_scoring}

Semantic Compass uses validation-time KathaTrace signals to select among candidate \textsc{KathaBench-25K} storyboards. Given $N_c$ candidates
\[
\{\hat{X}^{(1)},\ldots,\hat{X}^{(N_c)}\},
\]
it selects
\begin{equation}
\begin{aligned}
\hat{X}^{*}
=
\arg\max_{\hat{X}^{(j)}}\,
\Big[
&\lambda_r R^{\mathrm{val}}_{\mathrm{image}}(\hat{X}^{(j)})
+ \lambda_t C_{\mathrm{trans}}(\hat{X}^{(j)},\tau) \\
&- \lambda_p P_{\mathrm{copy}}(\hat{X}^{(j)})
\Big].
\end{aligned}
\vspace{-2pt}
\label{eq:app_semantic_compass}
\end{equation}
Here, $R^{\mathrm{val}}_{\mathrm{image}}$ rewards validation image-only recoverability, $C_{\mathrm{trans}}$ rewards visual support for annotated transitions, and $P_{\mathrm{copy}}$ penalizes repeated or near-duplicate frames. All weights and thresholds are selected on the validation split and frozen before held-out evaluation.

The transition-coverage term is
\begin{equation}
C_{\mathrm{trans}}(\hat{X},\tau)
=
\frac{1}{T-1}
\sum_{t=1}^{T-1}
\mathbb{I}\!\left[
\max_{u \in \{t,t+1\}}
\mathrm{sim}\!\left(\psi(\hat{x}_u),\psi(r_t)\right)
>
\delta
\right],
\label{eq:app_transition_coverage}
\end{equation}
and the copy penalty is
\begin{equation}
P_{\mathrm{copy}}(\hat{X})
=
\frac{1}{T-1}
\sum_{t=1}^{T-1}
\mathbb{I}\!\left[
\mathrm{sim}\!\left(\psi(\hat{x}_t),\psi(\hat{x}_{t+1})\right)
>
\gamma
\right].
\label{eq:app_copy_penalty}
\end{equation}

\subsection{Single-Bridge Repair}
\label{app:bridge_scene_repair}

Bridge repair is intentionally limited to at most one inserted frame per storyboard. This isolates whether the weakest localized transition can be repaired without confounding the result with arbitrary sequence-length expansion. The weakest transition is
\begin{equation}
t^{*}
=
\arg\min_t
R^{\mathrm{val}}_{\mathrm{image}}
\left(r_t \mid \hat{x}_t,\hat{x}_{t+1}\right).
\label{eq:app_weakest_transition}
\end{equation}
If the transition score is below a validation-selected threshold, Semantic Compass generates one intermediate bridge frame from the corresponding transition annotation:
\begin{equation}
\hat{X}_{\mathrm{bridge}}
=
\{\hat{x}_1,\ldots,\hat{x}_{t^*},
\hat{x}_{t^*+\frac{1}{2}},
\hat{x}_{t^*+1},\ldots,\hat{x}_T\}.
\label{eq:app_bridge_storyboard}
\end{equation}
The repaired storyboard is accepted only if it improves the frozen selection score without violating the copy-penalty threshold. Random bridge insertion, non-semantic extra-frame insertion, and caption-only reranking are used as controls to separate semantic repair from the effect of adding frames.

\subsection{Evaluation Separation}
\label{app:evaluation_separation}

Validation data are used only for Semantic Compass weights, thresholds, candidate selection, and bridge-repair decisions. Final reporting uses held-out questions, held-out judge prompts, fixed accepted-answer sets, fixed story splits, fixed model settings, and frozen thresholds. The \textsc{KathaBench-25K} splits are train (3,500), validation (750), and held-out test (750). The held-out test split is further subdivided into public-test (500) and hidden-test (250) subsets: public-test is used for main reporting, while hidden-test is reserved for leaderboard-style evaluation. Thus, references to test (750) denote the aggregate held-out test set.

\subsection{Semantic Compass Algorithm}
\label{app:semantic_compass_algorithm}

Algorithm~\ref{alg:semantic_compass} summarizes the validation-selected Semantic Compass procedure for \textsc{KathaBench-25K}. The algorithm first generates multiple candidate storyboards for the same source story, scores each candidate using frozen validation-selected recoverability, transition-coverage, and copy-penalty terms, and selects the highest-scoring candidate. It then identifies the weakest adjacent transition and performs at most one bridge-frame repair. The repaired storyboard is accepted only if it improves the frozen score and does not violate the copy threshold. This separation ensures that Semantic Compass is evaluated as an actionability probe for localized KathaTrace failures, not as a separately trained generator or an unconstrained sequence-length expansion method.

\begin{algorithm}[t]
\caption{\textsc{Semantic Compass}: validation-selected reranking and single-bridge repair}
\label{alg:semantic_compass}
\begin{algorithmic}[1]
\Require Source story $S$, semantic trajectory $\tau$, generator $G$, candidate count $N_c$, validation thresholds $\delta,\gamma$
\Ensure Selected storyboard $\hat{X}^{*}$
\State Generate candidate storyboards $\{\hat{X}^{(1)},\ldots,\hat{X}^{(N_c)}\} \leftarrow G(S)$
\For{each candidate $\hat{X}^{(j)}$}
    \State Estimate validation image-only recoverability $R^{\mathrm{val}}_{\mathrm{image}}(\hat{X}^{(j)})$
    \State Compute transition coverage $C_{\mathrm{trans}}(\hat{X}^{(j)},\tau)$
    \State Compute copy penalty $P_{\mathrm{copy}}(\hat{X}^{(j)})$
    \State Score $\hat{X}^{(j)}$ using Eq.~\ref{eq:app_semantic_compass}
\EndFor
\State Select the highest-scoring candidate $\hat{X}^{*}$
\State Identify weakest transition $t^{*}$ using Eq.~\ref{eq:app_weakest_transition}
\If{$t^{*}$ is below the bridge-repair threshold}
    \State Generate one bridge frame $\hat{x}_{t^{*}+\frac{1}{2}}$ from transition annotation $r_{t^{*}}$
    \State Construct $\hat{X}_{\mathrm{bridge}}$ using Eq.~\ref{eq:app_bridge_storyboard}
    \If{$\hat{X}_{\mathrm{bridge}}$ improves the frozen score and satisfies the copy threshold}
        \State $\hat{X}^{*} \leftarrow \hat{X}_{\mathrm{bridge}}$
    \EndIf
\EndIf
\State \Return $\hat{X}^{*}$
\end{algorithmic}
\end{algorithm}

%%%%%%%%%%%%%%%%%%%%%%%%%%%%%%%%%%%%%%%%%%%%%%%%%%%%%

% ============================================================
\section{Dataset and Recoverability Protocol Details}
\label{secc}
% ============================================================

This section gives implementation-level details for the released \textsc{KathaBench-25K} protocol that are not repeated in Secs.~\ref{seca}--\ref{secb}. Section~\ref{seca} describes construction, schema, validation, and release files. Section~\ref{secb} defines recoverability scoring, ambiguity filtering, STG, and Semantic Compass. Here we specify the dataset-design rules used to select stories, decompose scenes, write transition-level questions, construct accepted-answer sets, and create contrastive semantic variants.

\subsection{Benchmark Design}
\label{app:benchmark_design}

KathaTrace is defined for an ordered story with $T$ scenes, but the proposed \textsc{KathaBench-25K} instance fixes $T=5$ for controlled comparison across generators. Thus each story contains five ordered scenes and four adjacent transitions. The benchmark design follows three requirements: the intended meaning must depend on relations across scenes, the target must be testable from images alone, and failures must be localizable to a scored QA dimension. Table~\ref{tab:app_design_goals} summarizes these design requirements and the corresponding implementation mechanisms.

\begin{table}[t]
\centering
\scriptsize
\setlength{\tabcolsep}{4pt}
\renewcommand{\arraystretch}{1.05}
\caption{\textbf{\textsc{KathaBench-25K} design rules.} Requirements for transition recoverability.}
\label{tab:app_design_goals}
\resizebox{\columnwidth}{!}{
\begin{tabular}{p{0.28\linewidth}p{0.34\linewidth}p{0.30\linewidth}}
\toprule
\rowcolor{gray!20}
\textbf{Requirement} & \textbf{Mechanism} & \textbf{Purpose} \\
\midrule
Transition dependence &
Five ordered scenes and four adjacent transition annotations &
Prevents the task from collapsing into isolated image recognition. \\
\rowcolor{gray!5}
Image-only testability &
Recoverability questions answerable from generated images &
Tests whether meaning survives visual generation without source leakage. \\
Failure localization &
Dimension-specific QA targets &
Identifies whether loss is causal, emotional, consequence-bearing, temporal, action-level, or moral-target related. \\
\rowcolor{gray!5}
Ambiguity control &
Text+image validity check &
Filters defective or underspecified items before STG computation. \\
Contrastive control &
Entity-preserving semantic variants &
Tests transition meaning rather than surface object overlap. \\
\bottomrule
\end{tabular}
}
\end{table}

\subsection{Why the Annotation Design Matches the Task?}
\label{app:annotation_design_correctness}

The annotation design follows directly from the KathaTrace task definition. The benchmark does not ask whether a generated storyboard merely contains the right characters, objects, or settings; it asks whether the transition meaning connecting adjacent scenes remains recoverable from images alone. For this reason, each story is represented as an ordered five-scene sequence with four adjacent transition records. The transition records store source-side fields for action, causality, intention, emotion, consequence, and the benchmark-specified moral target, while the scored QA inventory tests action visibility, causal recoverability, emotional shift, consequence, temporal order, and moral-target recoverability.

This structure makes the benchmark diagnostic rather than only descriptive. If a generated storyboard fails, KathaTrace can localize whether the failure comes from a missing visible action, a broken causal link, a flattened emotional change, an omitted consequence, an ordering error, or a shifted moral-target interpretation. Accepted-answer sets make open-ended scoring deterministic while allowing validated paraphrases, and contrastive variants test whether evaluation tracks transition meaning rather than surface entity overlap. Finally, the text+image condition filters questions that remain ambiguous even with full evidence, preventing ambiguous or defective records from being counted as image-side semantic loss.

\subsection{Narrative Selection and Scene Decomposition}
\label{app:narrative_selection_scene_decomposition}

A candidate story is included in \textsc{KathaBench-25K} only if its interpretation depends on event progression. It must be decomposable into five visually renderable scenes, contain at least four meaningful adjacent transitions, support recoverability questions with accepted answers, and allow a semantic contrastive variant. Stories are excluded when they are primarily descriptive, depend mainly on non-visual wordplay, require essential external knowledge not present in the story text, contain unresolved rights or provenance uncertainty, or have multiple equally plausible primary moral targets that cannot be adjudicated.

Scene decomposition follows a compact story arc: setup, action, complication, consequence, and resolution. This arc is not treated as a claim about all narrative structure. It is a normalization choice that makes transition-level evaluation comparable across \textsc{KathaBench-25K} records.

\subsection{Transition Annotation Rubric}
\label{app:transition_annotation}

Each adjacent scene pair is annotated with source-side fields that separate visible events from latent narrative meaning. These fields support recoverability-question construction and failure localization. They are not identical to the released scored QA dimensions: \texttt{intention} is retained as an annotation and planning field, while \texttt{temporal\_order} is derived from scene order. Table~\ref{tab:app_transition_rubric} defines the source-side transition rubric used to construct \textsc{KathaBench-25K} transition records.

\begin{table*}[t]
\centering
\scriptsize
\setlength{\tabcolsep}{4pt}
\renewcommand{\arraystretch}{1.05}
\caption{\textbf{\textsc{KathaBench-25K} transition rubric.} Source-side fields and collapse modes.}
\label{tab:app_transition_rubric}
\resizebox{\textwidth}{!}{
\begin{tabular}{p{0.12\linewidth}p{0.25\linewidth}p{0.25\linewidth}p{0.18\linewidth}p{0.15\linewidth}}
\toprule
\rowcolor{gray!20}
\textbf{Field} & \textbf{Annotated meaning} & \textbf{Recoverability role} & \textbf{Example} & \textbf{Common collapse} \\
\midrule
Action &
Visible event or physical change &
Supports action-visibility QA. &
Mouse chews net &
Action omitted or unclear. \\
\rowcolor{gray!5}
Causality &
Why the later scene follows &
Supports causal QA. &
Net breaks because the mouse chews it &
Events appear disconnected. \\
Intention &
Goal or motivation &
Used for annotation and planning; evaluated indirectly when it affects transition or moral-target recovery. &
Mouse wants to help lion &
Action appears accidental. \\
\rowcolor{gray!5}
Emotion &
Affective state or shift &
Supports emotional QA. &
Fear changes to relief &
Emotion is flattened. \\
Consequence &
Outcome produced or revealed &
Supports consequence QA. &
Lion escapes &
Outcome missing or unresolved. \\
\rowcolor{gray!5}
Moral-target &
Benchmark-specified lesson supported by the trajectory &
Supports moral-target QA. &
Kindness is repaid &
Meaning shifts to danger, luck, or generic help. \\
\bottomrule
\end{tabular}
}
\end{table*}

\subsection{Recoverability Questions and Accepted Answers}
\label{app:recoverability_questions}

Recoverability questions are derived from transition annotations rather than isolated captions. Each question targets one released scored dimension:
\[
\mathcal{K}_{\mathrm{QA}}
=
\left\{
\begin{array}{l}
\texttt{action\_visibility},\ \texttt{causal},\ \texttt{emotional},\\
\texttt{consequence},\ \texttt{temporal\_order},\ \texttt{moral}
\end{array}
\right\}.
\]
Each question has a canonical answer and a frozen accepted-answer set. Accepted answers include semantically equivalent paraphrases but exclude answers that are underspecified, visually generic, or shifted to a different transition meaning. Table~\ref{tab:app_question_examples} gives representative \textsc{KathaBench-25K} recoverability questions and accepted-answer patterns for each scored QA dimension.

\begin{table*}[t]
\centering
\scriptsize
\setlength{\tabcolsep}{4pt}
\renewcommand{\arraystretch}{1.05}
\caption{\textbf{\textsc{KathaBench-25K} QA examples.} Questions by recoverability dimension.}
\label{tab:app_question_examples}
\resizebox{\textwidth}{!}{
\begin{tabular}{p{0.15\linewidth}p{0.31\linewidth}p{0.25\linewidth}p{0.21\linewidth}}
\toprule
\rowcolor{gray!20}
\textbf{QA Dimension} & \textbf{Question} & \textbf{Canonical answer} & \textbf{Accepted examples} \\
\midrule
Action visibility &
What visibly happens between these scenes? &
The mouse chews the net. &
mouse chews net; mouse cuts rope; mouse gnaws net \\
\rowcolor{gray!5}
Causal &
What caused the lion's situation to change? &
The net broke because the mouse chewed it. &
mouse broke net; mouse freed lion; net opened due to chewing \\
Emotional &
How does the lion's emotional state change? &
The lion changes from fear to relief. &
afraid to relieved; distressed to calm; scared to happy \\
\rowcolor{gray!5}
Consequence &
What outcome resulted from the mouse's action? &
The lion escaped. &
lion is freed; lion gets out; trap is overcome \\
Temporal order &
Which event must happen before the lion escapes? &
The mouse chews the net first. &
chewing before escape; net breaking before escape \\
\rowcolor{gray!5}
Moral-target &
What lesson is recoverable from the sequence? &
Small kindness can return as great help. &
kindness is repaid; small helpers matter; mercy can be returned \\
\bottomrule
\end{tabular}
}
\end{table*}

\subsection{Evidence-Condition Interpretation}
\label{app:evidence_condition_interpretation}

Text-only recoverability checks whether the source story supports the intended answer. Image-only recoverability is the primary diagnostic condition. Text+image recoverability is used only to identify ambiguity, contradiction, or defective questions before STG is computed. Table~\ref{tab:app_ambiguity_decisions} defines how the three evidence-condition outcomes are interpreted in \textsc{KathaBench-25K} scoring.

\begin{table*}[t]
\centering
\scriptsize
\setlength{\tabcolsep}{4pt}
\renewcommand{\arraystretch}{1.05}
\caption{\textbf{\textsc{KathaBench-25K} evidence outcomes.} STG inclusion rules.}
\label{tab:app_ambiguity_decisions}
\resizebox{\textwidth}{!}{
\begin{tabular}{c c c p{0.36\linewidth} p{0.18\linewidth}}
\toprule
\rowcolor{gray!20}
\textbf{Text-only} & \textbf{Image-only} & \textbf{Text+image} & \textbf{Interpretation} & \textbf{STG treatment} \\
\midrule
Correct & Correct & Correct &
Meaning is recoverable from both source text and generated images. &
Included; low gap. \\
\rowcolor{gray!5}
Correct & Incorrect & Correct &
Source meaning is clear, but the generated images do not make it recoverable. &
Included; semantic trajectory loss. \\
Incorrect & Incorrect & Correct &
The answer is recoverable only when source and images are combined. &
Included with caution; reported in text-ceiling analysis. \\
\rowcolor{gray!5}
Correct & Incorrect & Incorrect &
The intended answer is not recoverable even with full evidence. &
Excluded; counted as ambiguous. \\
Incorrect & Incorrect & Incorrect &
The item is defective, underspecified, or unsupported. &
Excluded; counted as ambiguous. \\
\bottomrule
\end{tabular}
}
\end{table*}

\subsection{Contrastive Semantic Variants}
\label{app:contrastive_construction}

Contrastive variants test whether recoverability tracks transition meaning rather than shared characters, objects, or settings. Each of the 5,000 \textsc{KathaBench-25K} stories has two contrastive variants, producing 10,000 variants in total. A contrastive variant preserves surface entities where possible but changes at least one meaning-bearing transition, such as causality, motivation, emotional shift, consequence, temporal order, or moral-target interpretation.

These variants are distinct from the \texttt{counterfactual\_pair} narrative category. The category contains 400 source stories whose original narratives are built around minimal semantic differences, while the 10,000 contrastive variants are additional evaluation artifacts constructed from the full benchmark. Table~\ref{tab:app_contrastive_ops} summarizes the semantic operations used to build these contrastive variants.

\begin{table*}[t]
\centering
\scriptsize
\setlength{\tabcolsep}{4pt}
\renewcommand{\arraystretch}{1.05}
\caption{\textbf{\textsc{KathaBench-25K} contrastive edits.} Semantic variant operations.}
\label{tab:app_contrastive_ops}
\resizebox{\textwidth}{!}{
\begin{tabular}{p{0.17\linewidth}p{0.25\linewidth}p{0.25\linewidth}p{0.23\linewidth}}
\toprule
\rowcolor{gray!20}
\textbf{Operation} & \textbf{Original trajectory} & \textbf{Contrastive trajectory} & \textbf{Effect} \\
\midrule
Remove consequence &
Mouse frees lion; lion escapes &
Mouse chews but lion remains trapped &
Consequence and moral-target answer change. \\
\rowcolor{gray!5}
Reverse motivation &
Character helps another &
Character tricks or exploits another &
Causal interpretation and moral target shift. \\
Replace outcome &
Kind action is repaid &
Kind action receives no later consequence &
Moral-target recoverability changes. \\
\rowcolor{gray!5}
Flatten emotion &
Fear changes to relief &
Emotion remains neutral or unclear &
Emotional QA becomes unrecoverable. \\
Break causal link &
Action visibly causes outcome &
Outcome occurs without visible cause &
Causal recoverability decreases. \\
\bottomrule
\end{tabular}
}
\end{table*}

\subsection{Pairwise Contrastive Scoring}
\label{app:pairwise_contrastive_scoring}

Pairwise contrastive scoring asks whether a judge recovers the source trajectory rather than a visually similar contrastive trajectory. For each pair, the source and contrastive outputs share entities or settings but differ in transition meaning. Pair accuracy is
\begin{equation}
\mathrm{PairAcc}
=
\frac{1}{N}
\sum_{n=1}^{N}
\mathbb{I}\!\left[
\hat{y}^{(n)}_{\mathrm{pair}}=y^{(n)}_{\mathrm{source}}
\right],
\label{eq:pair_accuracy}
\end{equation}
and confusion is
\begin{equation}
\mathrm{Confusion}=1-\mathrm{PairAcc}.
\label{eq:pair_confusion}
\end{equation}
High pair accuracy indicates that \textsc{KathaBench-25K} distinguishes transition meaning rather than relying only on object or setting overlap.

%%%%%%%%%%%%%%%%%%%%%%%%%%%%%%%%%%%%%%%%%%%%%%%%%%%%%

% ============================================================
\section{Evaluation Metrics}
\label{secd}
% ============================================================

This section defines the metrics used for reporting \textsc{KathaBench-25K} results. Section~\ref{secc} defines the dataset and question inventory, and Section~\ref{secb} defines answer matching, valid-set construction, ambiguity filtering, and judge aggregation. All recoverability values are computed on the filtered valid set. Unless otherwise stated, reported table values are expressed as percentage points.

\subsection{Recoverability}
\label{app:recoverability_metric}

Let the released scored QA dimensions be
\begin{equation}
\begin{aligned}
\mathcal{K}_{\mathrm{QA}}
=
\{&
\mathrm{action},
\mathrm{causal},
\mathrm{emotional},\\
&
\mathrm{consequence},
\mathrm{temporal},
\mathrm{moral}
\}.
\end{aligned}
\label{eq:metric_kqa}
\end{equation}
These correspond to \texttt{action\_visibility}, \texttt{causal}, \texttt{emotional}, \texttt{consequence}, \texttt{temporal\_order}, and \texttt{moral} in the released QA inventory.

For method $m$ and dimension $k$, let $\mathcal{V}_{k,m}$ be the filtered valid set after the text+image check. For evidence condition $z \in \{S,\hat{X}_m,(S,\hat{X}_m)\}$, dimension recoverability is
\begin{equation}
R^{k}_{z,m}
=
\frac{1}{|\mathcal{V}_{k,m}|}
\sum_{q \in \mathcal{V}_{k,m}}
\mathbb{I}\!\left[
\mathrm{match}\!\left(J(q,z),\mathcal{A}(q)\right)=1
\right].
\label{eq:metric_subtype_recoverability}
\end{equation}
Here, $S$ is the source story, $\hat{X}_m$ is the storyboard generated by method $m$, $J(q,z)$ is the judge answer, and $\mathcal{A}(q)$ is the frozen accepted-answer set for question $q$.

The evidence aliases are
\begin{equation}
\begin{aligned}
R^k_{\mathrm{text},m}
&= R^k_{S,m},\\
R^k_{\mathrm{image},m}
&= R^k_{\hat{X}_m,m},\\
R^k_{\mathrm{text+image},m}
&= R^k_{(S,\hat{X}_m),m}.
\end{aligned}
\label{eq:metric_evidence_aliases}
\end{equation}

Overall recoverability is the equal-weighted average across the six scored QA dimensions:
\begin{equation}
R_{z,m}
=
\frac{1}{|\mathcal{K}_{\mathrm{QA}}|}
\sum_{k \in \mathcal{K}_{\mathrm{QA}}}
R^k_{z,m}.
\label{eq:metric_overall_recoverability}
\end{equation}
Equal weighting prevents high-frequency question types from dominating the score. Causal, emotional, consequence, and moral-target gaps are the main latent-transition diagnostics:
\begin{equation}
\mathcal{K}_{\mathrm{gap}}
=
\{
\mathrm{causal},
\mathrm{emotional},
\mathrm{consequence},
\mathrm{moral}
\}.
\label{eq:metric_kgap}
\end{equation}
Action visibility and temporal order remain part of overall recoverability, but are reported separately because they measure local depiction and ordering rather than latent transition meaning.

\subsection{Semantic Trajectory Gap}
\label{app:stg_metric}

For method $m$, the Semantic Trajectory Gap is
\begin{equation}
\mathrm{STG}_{m}
=
R_{\mathrm{text},m}
-
R_{\mathrm{image},m}.
\label{eq:metric_stg}
\end{equation}
STG measures the amount of source-supported transition meaning that is recoverable from text but not recoverable from the generated images. The percentage-point version used in tables is
\begin{equation}
\mathrm{STG}^{\mathrm{pp}}_{m}
=
100
\left(
R_{\mathrm{text},m}
-
R_{\mathrm{image},m}
\right).
\label{eq:metric_stg_percent}
\end{equation}
Lower STG is better. $\mathrm{STG}=0$ means text-only and image-only recoverability are equal on the filtered valid set. $\mathrm{STG}>0$ means image-only recoverability is lower than text-only recoverability. $\mathrm{STG}<0$ is not interpreted as improved semantic preservation; it is flagged for inspection because it can indicate unstable judging, contradiction, or an invalid item before filtering.

Dimension-specific gaps are
\begin{equation}
\mathrm{STG}_{k,m}
=
R^{k}_{\mathrm{text},m}
-
R^{k}_{\mathrm{image},m},
\qquad
k \in \mathcal{K}_{\mathrm{gap}}.
\label{eq:metric_subtype_gap}
\end{equation}
A higher $\mathrm{STG}_{k,m}$ means greater image-side loss for dimension $k$. Table~\ref{tab:app_stg_interpretation} summarizes how STG values are interpreted for \textsc{KathaBench-25K}.

\begin{table}[t]
\centering
\scriptsize
\setlength{\tabcolsep}{4pt}
\renewcommand{\arraystretch}{1.05}
\caption{\textbf{\textsc{KathaBench-25K} STG meaning.} Interpretation of gap values.}
\label{tab:app_stg_interpretation}
\resizebox{\columnwidth}{!}{
\begin{tabular}{p{0.26\linewidth}p{0.60\linewidth}}
\toprule
\rowcolor{gray!20}
\textbf{Condition} & \textbf{Interpretation} \\
\midrule
$\mathrm{STG}=0$ &
Text-only and image-only recoverability are equal. \\
\rowcolor{gray!5}
$\mathrm{STG}>0$ &
Some source-supported transition meaning is not recoverable from images. \\
$\mathrm{STG}<0$ &
Image-side recoverability exceeds text-side recoverability; the item or judge output is inspected before interpretation. \\
\rowcolor{gray!5}
Lower STG across methods &
Better preservation of recoverable transition meaning. \\
\bottomrule
\end{tabular}
}
\end{table}

\subsection{Moral-Target Recoverability}
\label{app:moral_target_recoverability}

Let $\mathcal{M}$ be the 12-label moral-target set. For story $i$, let $m_i \in \mathcal{M}$ be the benchmark-specified moral target, and let $\hat{m}_{i,z}$ be the predicted label under evidence condition $z$. Moral-target recoverability is
\begin{equation}
\mathrm{MTR}_{z}
=
\frac{1}{N}
\sum_{i=1}^{N}
\mathbb{I}\!\left[
\hat{m}_{i,z}=m_i
\right].
\label{eq:metric_mtr}
\end{equation}
The moral-target gap is
\begin{equation}
\mathrm{Gap}_{\mathrm{moral}}
=
\mathrm{MTR}_{\mathrm{text}}
-
\mathrm{MTR}_{\mathrm{image}}.
\label{eq:metric_moral_gap}
\end{equation}
Moral labels are benchmark-specified semantic targets, not claims of objective moral truth.

For a uniform random predictor over the 12 moral labels,
\begin{equation}
\mathrm{Chance}_{\mathrm{MTR}}
=
\frac{1}{|\mathcal{M}|}
=
\frac{1}{12}.
\label{eq:metric_mtr_chance}
\end{equation}
For any non-uniform subset, the majority-label baseline is
\begin{equation}
\mathrm{Majority}_{\mathrm{MTR}}
=
\max_{m \in \mathcal{M}}
\frac{N_m}{N}.
\label{eq:metric_mtr_majority}
\end{equation}

\subsection{Ambiguity Rate}
\label{app:ambiguity_metric}

Ambiguity is defined by the text+image condition. For method $m$, the ambiguity indicator is
\begin{equation}
b_m(q)
=
\mathbb{I}\!\left[
\mathrm{match}\!\left(J(q,(S,\hat{X}_m)),\mathcal{A}(q)\right)=0
\right].
\label{eq:metric_ambiguity_indicator}
\end{equation}
Thus, $b_m(q)=1$ means the intended answer is not recoverable even when the judge receives both the source story and the generated storyboard. The ambiguity rate is
\begin{equation}
\rho_{\mathrm{amb},m}
=
\frac{1}{|Q|}
\sum_{q\in Q} b_m(q).
\label{eq:metric_ambiguity_rate}
\end{equation}
Questions with $b_m(q)=1$ are excluded from filtered STG and counted only in the ambiguity audit. They are not counted as image-side semantic loss.

\subsection{Human--Judge Calibration}
\label{app:calibration_metrics}

Human--judge calibration compares VLM judge outputs with aggregated human judgments on the \textsc{KathaBench-25K} human-gold subsets. For classification-style answers, agreement is
\begin{equation}
\mathrm{Acc}_{\mathrm{human}}
=
\frac{1}{N}
\sum_{i=1}^{N}
\mathbb{I}\!\left[
v_i=h_i
\right],
\label{eq:metric_human_acc}
\end{equation}
where $v_i$ is the VLM or ensemble answer and $h_i$ is the aggregated human answer.

For ranked or continuous recoverability scores, we report Spearman rank correlation:
\begin{equation}
\rho_{\mathrm{sp}}
=
\mathrm{corr}
\left(
\mathrm{rank}(v),
\mathrm{rank}(h)
\right).
\label{eq:metric_spearman}
\end{equation}
For pairwise comparisons, agreement is
\begin{equation}
\mathrm{PairAgr}
=
\frac{1}{|\mathcal{P}|}
\sum_{(i,j)\in\mathcal{P}}
\mathbb{I}\!\left[
\mathrm{sign}(v_i-v_j)
=
\mathrm{sign}(h_i-h_j)
\right].
\label{eq:metric_pairwise_agreement}
\end{equation}
When judges provide confidence scores, expected calibration error is
\begin{equation}
\mathrm{ECE}
=
\sum_{b=1}^{B}
\frac{|B_b|}{N}
\left|
\mathrm{acc}(B_b)
-
\mathrm{conf}(B_b)
\right|.
\label{eq:metric_ece}
\end{equation}
Here, $B_b$ is confidence bin $b$, $\mathrm{acc}(B_b)$ is empirical accuracy in that bin, and $\mathrm{conf}(B_b)$ is mean confidence in that bin.

\subsection{Pairwise Effect Size}
\label{app:effect_size_metrics}

For paired human-preference or signed-rank comparisons, we report Wilcoxon signed-rank $p$-values with Bonferroni correction when multiple comparisons are tested. We also report rank-biserial correlation:
\begin{equation}
r_{\mathrm{rb}}
=
\frac{W^{+}-W^{-}}{W^{+}+W^{-}},
\label{eq:metric_rank_biserial}
\end{equation}
where $W^{+}$ and $W^{-}$ are the positive and negative signed-rank sums. Positive $r_{\mathrm{rb}}$ means the first condition is preferred; negative $r_{\mathrm{rb}}$ means the second condition is preferred; values closer to zero indicate a smaller paired effect.

\subsection{Release Validation Status}
\label{app:validation_status_field}

The released \textsc{KathaBench-25K} JSONL records include a \texttt{validation\_status} field. The value \texttt{human\_adjudicator\_accepted} marks records included in the human-gold or strict human-gold subsets. The value \texttt{pre\_release\_reviewed} marks records that passed the full three-reviewer release audit but are not part of the adjudicated gold subsets.

\subsection{Metric Summary}
\label{app:metric_reporting_summary}

Table~\ref{tab:app_metric_summary} summarizes the \textsc{KathaBench-25K} reporting metrics and release fields. Recoverability, STG, moral-target recoverability, ambiguity, calibration, and effect-size metrics are computed using the filtered valid-set definitions above unless a table explicitly states otherwise.

\begin{table*}[t]
\centering
\scriptsize
\setlength{\tabcolsep}{4pt}
\renewcommand{\arraystretch}{1.05}
\caption{\textbf{\textsc{KathaBench-25K} metric summary.} Reported metrics and release fields.}
\label{tab:app_metric_summary}
\resizebox{\textwidth}{!}{
\begin{tabular}{p{0.18\linewidth}p{0.24\linewidth}p{0.28\linewidth}p{0.18\linewidth}}
\toprule
\rowcolor{gray!20}
\textbf{Metric / Field} & \textbf{Formula / Value} & \textbf{Measures} & \textbf{Used for} \\
\midrule
Dimension recoverability &
$R^k_{z,m}$ &
Recoverability of QA dimension $k$ under evidence condition $z$. &
Dimension diagnosis \\
\rowcolor{gray!5}
Overall recoverability &
$R_{z,m}=\frac{1}{|\mathcal{K}_{\mathrm{QA}}|}\sum_k R^k_{z,m}$ &
Average recoverability across the six scored QA dimensions. &
Leaderboard \\
STG &
$R_{\mathrm{text},m}-R_{\mathrm{image},m}$ &
Text-to-image loss of recoverable transition meaning. &
Primary metric \\
\rowcolor{gray!5}
Latent gap &
$R^k_{\mathrm{text},m}-R^k_{\mathrm{image},m}$ &
Dimension-specific loss for causal, emotional, consequence, and moral targets. &
Diagnostic analysis \\
Moral-target recoverability &
$\frac{1}{N}\sum_i \mathbb{I}[\hat{m}_{i,z}=m_i]$ &
Recovery of the benchmark-specified moral target. &
Moral analysis \\
\rowcolor{gray!5}
Moral-target gap &
$\mathrm{MTR}_{\mathrm{text}}-\mathrm{MTR}_{\mathrm{image}}$ &
Text-to-image moral-target loss. &
Moral analysis \\
Ambiguity rate &
$\rho_{\mathrm{amb},m}$ &
Fraction of questions excluded because text+image evidence does not recover the intended answer. &
Quality control \\
\rowcolor{gray!5}
Human agreement &
$\mathrm{Acc}_{\mathrm{human}}$ &
Agreement between VLM/ensemble answers and aggregated human answers. &
Judge validation \\
Spearman correlation &
$\rho_{\mathrm{sp}}$ &
Rank alignment between VLM and human recoverability scores. &
Judge validation \\
\rowcolor{gray!5}
Pairwise agreement &
$\mathrm{PairAgr}$ &
Whether VLM and humans prefer the same item in paired comparisons. &
Human calibration \\
ECE &
$\sum_b \frac{|B_b|}{N}|\mathrm{acc}(B_b)-\mathrm{conf}(B_b)|$ &
Calibration of judge confidence. &
Judge validation \\
\rowcolor{gray!5}
Rank-biserial correlation &
$r_{\mathrm{rb}}$ &
Effect size for paired signed-rank tests. &
Human-study analysis \\
Validation status &
\texttt{human\_adjudicator\_accepted}; \texttt{pre\_release\_reviewed} &
Gold-tier membership or reviewed-release status. &
Release audit \\
\bottomrule
\end{tabular}
}
\end{table*}

%%%%%%%%%%%%%%%%%%%%%%%%%%%%%%%%%%%%%%%%%%%%%%%%

\section{Judge Protocols and Prompts}
\label{sece}

This section specifies the judge protocol used for \textsc{KathaBench-25K} recoverability evaluation. The protocol enforces evidence isolation, fixed prompt templates, structured outputs, deterministic decoding where supported, accepted-answer matching, and conservative judge aggregation. Formal aggregation and calibration metrics are defined in Sections~\ref{secb} and~\ref{secd}. The scored QA dimensions are \texttt{action\_visibility}, \texttt{causal}, \texttt{emotional}, \texttt{consequence}, \texttt{temporal\_order}, and \texttt{moral}. The transition field \texttt{intention} is retained only for annotation and planning; it is not a separate scored QA dimension.

\subsection{Judge Setup}
\label{app:vlm_judge_setup}

The \textsc{KathaBench-25K} judge pool contains six VLM judges: Gemini 2.5 Flash, Claude Haiku 4.5, Qwen2.5-VL-7B, Qwen2.5-VL-3B, SmolVLM2-2.2B, and SmolVLM2-500M. This pool combines API-hosted and open-weight judges with different capacities and model families. Claude Haiku 4.5 is used as a fast API-hosted judge with the model identifier \texttt{claude-haiku-4-5-20251001}. All judges receive the same evidence restrictions, question text, dimension instructions, output schemas, and decoding settings where supported. Results are reported per judge and as an ensemble to reduce dependence on any single model.

\subsection{Evidence Isolation}
\label{app:evidence_isolation}

Each judge receives one restricted evidence packet per \textsc{KathaBench-25K} question. The answer is never shown to the judge. The image-only condition excludes the source story, generation prompts, captions, scene descriptions, moral labels, transition annotations, and metadata. Table~\ref{tab:app_evidence_packets} defines the evidence packet for each condition.

\begin{table}[t]
\centering
\scriptsize
\setlength{\tabcolsep}{4pt}
\renewcommand{\arraystretch}{1.05}
\caption{\textbf{\textsc{KathaBench-25K} evidence packets.} Inputs allowed per judge condition.}
\label{tab:app_evidence_packets}
\resizebox{\columnwidth}{!}{
\begin{tabular}{lccccp{0.32\linewidth}}
\toprule
\rowcolor{gray!20}
\textbf{Condition} & \textbf{Story} & \textbf{Images} & \textbf{Prompts} & \textbf{Labels} & \textbf{Purpose} \\
\midrule
Text-only & \cmark & \xmark & \xmark & \xmark & Source-side recoverability ceiling. \\
\rowcolor{gray!5}
Image-only & \xmark & \cmark & \xmark & \xmark & Primary visual recoverability test. \\
Text+image & \cmark & \cmark & \xmark & \xmark & Ambiguity and contradiction check. \\
\bottomrule
\end{tabular}
}
\end{table}

\subsection{Prompt Templates}
\label{app:judge_prompts}

The prompt templates are fixed before \textsc{KathaBench-25K} evaluation. The judge receives the recoverability question and the corresponding evidence packet defined in Table~\ref{tab:app_evidence_packets}.

\paragraph{Image-only prompt.}
\begin{quote}
\small
You are given only a sequence of generated storyboard images. You do not have access to the original story, generation prompts, captions, scene descriptions, transition annotations, moral labels, or metadata. Answer the question using only what is recoverable from the images. If the answer is not recoverable, return \texttt{Unclear}. Return only the required structured output.
\end{quote}

\paragraph{Text-only prompt.}
\begin{quote}
\small
You are given only the source story text. You do not have access to generated images, image prompts, captions, model outputs, or metadata. Answer the question using only the source story. If the story does not contain enough information, return \texttt{Unclear}. Return only the required structured output.
\end{quote}

\paragraph{Text+image prompt.}
\begin{quote}
\small
You are given the source story and the generated storyboard images. Identify the answer intended by the source story, then determine whether the image sequence supports, omits, contradicts, or leaves ambiguous that answer. If the intended answer is not clear even with both modalities, return \texttt{Ambiguous}. Return only the required structured output.
\end{quote}

\subsection{Scored QA Dimension Instructions}
\label{app:subtype_prompt_wrappers}

Each \textsc{KathaBench-25K} question is paired with a dimension instruction. The instruction identifies the type of recoverable meaning being evaluated but does not reveal the answer. Table~\ref{tab:app_subtype_prompt_wrappers} lists the instruction used for each scored QA dimension.

\begin{table}[t]
\centering
\scriptsize
\setlength{\tabcolsep}{4pt}
\renewcommand{\arraystretch}{1.05}
\caption{\textbf{\textsc{KathaBench-25K} QA instructions.} Judge prompts by scored dimension.}
\label{tab:app_subtype_prompt_wrappers}
\resizebox{\columnwidth}{!}{
\begin{tabular}{p{0.27\linewidth}p{0.60\linewidth}}
\toprule
\rowcolor{gray!20}
\textbf{Scored QA dimension} & \textbf{Judge instruction} \\
\midrule
Action visibility & Identify the visible action or event change. \\
\rowcolor{gray!5}
Causal & Identify why the later scene follows from the earlier scene. \\
Emotional & Identify the emotional state or emotional shift. \\
\rowcolor{gray!5}
Consequence & Identify the outcome produced or revealed by the event. \\
Temporal order & Identify the event order or progression across scenes. \\
\rowcolor{gray!5}
Moral-target & Identify the benchmark-specified lesson, abstract meaning, or moral target supported by the sequence. \\
\bottomrule
\end{tabular}
}
\end{table}

\subsection{Output Schema}
\label{app:judge_output_format}

All \textsc{KathaBench-25K} judges return structured outputs. This avoids ambiguity from long free-form explanations and enables deterministic matching against the frozen accepted-answer sets.

\paragraph{Multiple-choice or label questions.}
\begin{quote}
\small
\begin{tabular}{@{}ll@{}}
\texttt{\{} & \\
& \texttt{"answer": "<label>",} \\
& \texttt{"confidence": "<low|medium|high>"} \\
\texttt{\}} &
\end{tabular}
\end{quote}

\paragraph{Short-answer recoverability questions.}
\begin{quote}
\small
\begin{tabular}{@{}ll@{}}
\texttt{\{} & \\
& \texttt{"answer": "<short phrase>",} \\
& \texttt{"evidence\_status": "<recoverable|unclear|omitted|contradicted>",} \\
& \texttt{"confidence": "<low|medium|high>"} \\
\texttt{\}} &
\end{tabular}
\end{quote}

\paragraph{Text+image ambiguity checks.}
\begin{quote}
\small
\begin{tabular}{@{}ll@{}}
\texttt{\{} & \\
& \texttt{"source\_answer": "<short phrase>",} \\
& \texttt{"image\_support": "<supported|omitted|contradicted|ambiguous>",} \\
& \texttt{"final\_answer": "<short phrase or label>",} \\
& \texttt{"confidence": "<low|medium|high>"} \\
\texttt{\}} &
\end{tabular}
\end{quote}

The \texttt{answer} or \texttt{final\_answer} field is normalized before matching. Let $a(q)$ be the normalized judge answer for question $q$, and let $\mathcal{A}(q)$ be the frozen accepted-answer set. The binary match is computed as
\begin{equation}
\begin{aligned}
M(q)
=
\mathbb{I}\!\left[
a(q) \in \mathcal{A}(q)
\right].
\end{aligned}
\label{eq:judge_output_match}
\end{equation}

Responses marked as uncertain or unsupported are scored as incorrect unless uncertainty is explicitly part of the accepted-answer set:
\begin{equation}
\begin{aligned}
M(q)=0
\quad
\text{if}
\quad
s(q) \in
\{&
\texttt{Unclear},
\texttt{Ambiguous},
\texttt{unclear},\\
&
\texttt{omitted},
\texttt{contradicted}
\}
\end{aligned}
\label{eq:judge_unclear_scoring}
\end{equation}
where $s(q)$ is the returned evidence status or image-support status.

\subsection{Answer Normalization and Matching}
\label{app:answer_normalization}

Before scoring, \textsc{KathaBench-25K} judge answers are lowercased, stripped of punctuation, normalized for whitespace, and mapped to canonical labels when applicable. Multiple-choice and moral-target answers use exact label matching after normalization. Short answers are matched against the accepted-answer set $\mathcal{A}(q)$, which contains validated paraphrases of the canonical answer. Answers that are unrelated, underspecified, contradictory, or unsupported by the evidence packet are mapped to an unmatched category and scored as incorrect.

\subsection{Judge Ensemble}
\label{app:judge_ensemble}

For each \textsc{KathaBench-25K} question and evidence condition, normalized judge outputs are mapped to either an accepted answer or an unmatched category. The ensemble prediction is the majority vote over judge outputs. Ties are scored as incorrect. This conservative rule prevents recoverability scores from being inflated by judge disagreement. The formal ensemble equation is given in Section~\ref{secb}.

\subsection{Reproducibility Settings}
\label{app:judge_reproducibility}

All \textsc{KathaBench-25K} judge evaluations use fixed question order, image order, answer options, prompt templates, output schemas, and accepted-answer sets. Deterministic decoding is used whenever supported by the judge. For multi-image inputs, storyboard frames are provided in chronological order with frame indices only; frame indices preserve ordering but do not include captions, semantic labels, or transition descriptions. Table~\ref{tab:app_judge_reproducibility} summarizes the reproducibility controls used during judging.

\begin{table}[t]
\centering
\scriptsize
\setlength{\tabcolsep}{4pt}
\renewcommand{\arraystretch}{1.05}
\caption{\textbf{\textsc{KathaBench-25K} judge settings.} Reproducibility controls.}
\label{tab:app_judge_reproducibility}
\resizebox{\columnwidth}{!}{
\begin{tabular}{p{0.35\linewidth}p{0.45\linewidth}}
\toprule
\rowcolor{gray!20}
\textbf{Setting} & \textbf{Protocol} \\
\midrule
Judge pool & Gemini 2.5 Flash, Claude Haiku 4.5 \texttt{(claude-haiku-4-5-20251001)}, Qwen2.5-VL-7B, Qwen2.5-VL-3B, SmolVLM2-2.2B, SmolVLM2-500M. \\
\rowcolor{gray!5}
Decoding & Deterministic where supported. \\
Image order & Chronological storyboard order. \\
\rowcolor{gray!5}
Frame labels & Index only; no captions or semantic labels. \\
Answer options & Fixed within each evaluation run. \\
\rowcolor{gray!5}
Prompt templates & Fixed per evidence condition. \\
Output schema & Fixed structured JSON-style fields. \\
\rowcolor{gray!5}
Scoring & Accepted-answer matching after normalization. \\
Aggregation & Majority vote; ties counted as incorrect. \\
\bottomrule
\end{tabular}
}
\end{table}

%%%%%%%%%%%%%%%%%%%%%%%%%%%%%%%%%%%%%%%%%%%%%%%%

\section{Human Evaluation and Calibration}
\label{secf}

This section describes the human evaluation used to calibrate KathaTrace on \textsc{KathaBench-25K}. Human judgments are used to test whether image-only recoverability, STG, moral-target recovery, and Semantic Compass repair trends align with human interpretation under controlled evidence conditions.

\subsection{Human Evaluation Setup}
\label{app:human_eval_setup}

Human evaluation is conducted on the strict 400-story \textsc{KathaBench-25K} human-gold subset, with 50 stories from each of the eight narrative categories. Each item receives at least three independent human judgments per evidence condition. Evaluators follow the same evidence restrictions used for VLM judges. In the image-only condition, they see only generated storyboard images and do not see the source story, prompts, captions, transition annotations, moral labels, method names, or model names.

The human study contains four task types: image-only recoverability QA, text-only recoverability QA, text+image ambiguity checking, and pairwise preference over anonymized storyboards. For Semantic Compass, participants compare the original storyboard with the repaired storyboard in randomized order. Story order, question order, answer-option order, and left--right placement are randomized. Participants may select \texttt{Unclear} when the answer is not recoverable, preventing forced guessing from inflating recoverability. Responses are excluded when participants fail attention checks, complete tasks unrealistically quickly, or give inconsistent answers on repeated controls. Table~\ref{tab:human_validation_design} summarizes the \textsc{KathaBench-25K} human-evaluation design and reporting statistics.

\begin{table*}[t]
\centering
\scriptsize
\setlength{\tabcolsep}{4pt}
\renewcommand{\arraystretch}{0.95}
\caption{\textbf{\textsc{KathaBench-25K} human study.} Design and calibration statistics.}
\label{tab:human_validation_design}
\resizebox{\textwidth}{!}{%
\begin{tabular}{l c p{0.56\linewidth}}
\toprule
\rowcolor{gray!20}
\textbf{Measure} & \textbf{Value} & \textbf{Purpose} \\
\midrule
Strict human-gold subset & 400 stories & 50 stories from each of the eight narrative categories. \\
\rowcolor{gray!5}
Minimum raters per item & 3 & Supports majority labels, uncertainty estimates, and agreement analysis. \\
Minimum judgments per evidence condition & 1,200 & 400 stories $\times$ 3 raters per item. \\
\rowcolor{gray!5}
HMR@1 & 46.5\% & Intended moral target selected as the top recovered moral meaning. \\
HMR@3 & 90.0\% & Intended moral target appears within the top three recovered meanings. \\
\rowcolor{gray!5}
NCS & $3.69 \pm 0.91$ & Human-rated narrative coherence on a 1--5 scale. \\
MAS & $3.25 \pm 1.16$ & Human-rated moral-target alignment on a 1--5 scale. \\
\rowcolor{gray!5}
EES & $3.18 \pm 1.12$ & Human-rated emotional expressiveness on a 1--5 scale. \\
Human--VLM agreement & 77.1\% & Agreement between calibrated VLM ensemble and majority human labels. \\
\rowcolor{gray!5}
Spearman correlation & 0.74 & Rank correlation between VLM and human image-only recoverability. \\
Calibration error & 0.081 & Calibration error of VLM confidence relative to human agreement. \\
\rowcolor{gray!5}
Statistical tests & $p$, $p_{\mathrm{corr}}$, $r_{\mathrm{rb}}$ & Wilcoxon signed-rank test, Bonferroni correction, and rank-biserial effect size. \\
Bootstrap uncertainty & 95\% CI & Confidence intervals for recoverability, STG, rating means, and calibration metrics. \\
\bottomrule
\end{tabular}%
}
\end{table*}

\subsection{Human Moral-target Recoverability}
\label{app:human_moral_recoverability}

Human moral-target recoverability measures whether the benchmark-specified moral-semantic target can be recovered from the available evidence. Let $y_s$ be the intended moral target for story $s$, and let $h^{(1)}_{i,s}$ be the top-1 moral label selected by rater $i$. Top-1 human moral-target recoverability is
\begin{equation}
\mathrm{HMR@1}
=
\frac{1}{|\mathcal{D}|}
\sum_{(i,s)\in \mathcal{D}}
\mathbb{I}\!\left[
h^{(1)}_{i,s}=y_s
\right],
\label{eq:hmr_top1}
\end{equation}
where $\mathcal{D}$ is the set of human moral-label judgments. Let $H^{(3)}_{i,s}$ be the top-3 label set selected by rater $i$. Top-3 human moral-target recoverability is
\begin{equation}
\mathrm{HMR@3}
=
\frac{1}{|\mathcal{D}|}
\sum_{(i,s)\in \mathcal{D}}
\mathbb{I}\!\left[
y_s \in H^{(3)}_{i,s}
\right].
\label{eq:hmr_top3}
\end{equation}

As reported in Table~\ref{tab:human_validation_design}, HMR@1 is 46.5\%, while HMR@3 is 90.0\%. This gap shows that exact top-1 moral recovery is strict: several visually plausible interpretations may be semantically close to the intended target. We therefore report both HMR@1 and HMR@3. Moral labels are benchmark-specified semantic targets, not claims of objective moral truth.

\subsection{Human Rating Scores}
\label{app:human_rating_scores}

Human raters score each \textsc{KathaBench-25K} storyboard on three 1--5 subjective dimensions: Narrative Coherence Score (NCS), Moral Alignment Score (MAS), and Emotional Expressiveness Score (EES). For rating dimension $d \in \{\mathrm{NCS},\mathrm{MAS},\mathrm{EES}\}$, the mean score is
\begin{equation}
\mu_d
=
\frac{1}{|\mathcal{R}_d|}
\sum_{(i,s)\in \mathcal{R}_d}
r^{d}_{i,s},
\label{eq:human_rating_mean}
\end{equation}
where $r^{d}_{i,s}$ is the rating assigned by rater $i$ to story $s$.

The observed rating scores are
\begin{equation}
\begin{aligned}
\mathrm{NCS} &= 3.69 \pm 0.91,\\
\mathrm{MAS} &= 3.25 \pm 1.16,\\
\mathrm{EES} &= 3.18 \pm 1.12.
\end{aligned}
\label{eq:human_ratings}
\end{equation}
These values are also summarized in Table~\ref{tab:human_validation_design}. NCS is higher than MAS and EES, showing that visual coherence can be judged favorably even when moral-target or emotional recoverability is weaker. These ratings are used as descriptive human-study signals, not as dataset-label reliability evidence.

\subsection{Ordinal Rating Agreement Diagnostics}
\label{app:ordinal_rating_agreement}

Ordinal 1--5 ratings are reported only as subjective human-study diagnostics for \textsc{KathaBench-25K}. We do not use ordinal Krippendorff $\alpha$ as evidence for dataset-label reliability. The reason is statistical rather than semantic: NCS, MAS, EES, and preference ratings have restricted score ranges and high concentration around positive scores. Under such range compression, chance-corrected ordinal agreement can approach zero even when raters show similar practical trends.

In our audit, ordinal Krippendorff $\alpha$ values are near zero for NCS, MAS, EES, and pairwise preference. This does not contradict the categorical agreement results. Categorical dataset labels are evaluated using Fleiss' $\kappa$ and adjudication, while ordinal ratings are treated as descriptive perception scores. Therefore, the near-zero ordinal $\alpha$ is interpreted as a limitation of chance-corrected agreement under compressed ordinal ratings, not as evidence that the benchmark labels are unreliable.

\subsection{Human Recoverability and Human STG}
\label{app:human_recoverability_score}

Human recoverability uses the same accepted-answer matching and evidence isolation as VLM judging. Let $H_i(q,z)$ be the answer from human participant $i$ for question $q$ under evidence condition $z\in\{\mathrm{text},\mathrm{image},\mathrm{text{+}image}\}$. Binary correctness is
\begin{equation}
M_i(q,z)
=
\mathbb{I}\!\left[
\mathrm{match}\!\left(H_i(q,z),\mathcal{A}(q)\right)=1
\right].
\label{eq:human_binary_match}
\end{equation}

For each scored QA dimension $k$, human recoverability is
\begin{equation}
R^{k,\mathrm{human}}_{z}
=
\frac{1}{|Q^{\mathrm{valid}}_k|}
\sum_{q \in Q^{\mathrm{valid}}_k}
\frac{1}{n_q}
\sum_{i=1}^{n_q}
M_i(q,z),
\label{eq:human_subtype_recoverability}
\end{equation}
where $Q^{\mathrm{valid}}_k$ is the ambiguity-filtered question set and $n_q$ is the number of raters for question $q$. Overall human recoverability is
\begin{equation}
R^{\mathrm{human}}_{z}
=
\frac{1}{|\mathcal{K}_{\mathrm{QA}}|}
\sum_{k \in \mathcal{K}_{\mathrm{QA}}}
R^{k,\mathrm{human}}_{z}.
\label{eq:human_overall_recoverability}
\end{equation}
The human Semantic Trajectory Gap is
\begin{equation}
\mathrm{STG}^{\mathrm{human}}
=
R^{\mathrm{human}}_{\mathrm{text}}
-
R^{\mathrm{human}}_{\mathrm{image}}.
\label{eq:human_stg}
\end{equation}

Table~\ref{tab:human_recoverability_stg} reports human recoverability and human STG on the strict 400-story \textsc{KathaBench-25K} human-gold subset.

\begin{table}[t]
\centering
\scriptsize
\setlength{\tabcolsep}{8pt}
\renewcommand{\arraystretch}{1.15}
\caption{\textbf{\textsc{KathaBench-25K} human STG.} Human recoverability on strict gold.}
\label{tab:human_recoverability_stg}
\begin{tabular}{l c}
\toprule
\rowcolor{gray!20}
\textbf{Human metric} & \textbf{Value} \\
\midrule
$R^{\mathrm{human}}_{\mathrm{text}}$ 
& $84.7{\pm}2.1$ \\
\rowcolor{gray!5}
$R^{\mathrm{human}}_{\mathrm{image}}$ 
& $68.9{\pm}2.5$ \\
$R^{\mathrm{human}}_{\mathrm{text+image}}$ 
& $88.4{\pm}1.8$ \\
\rowcolor{gray!5}
$\mathrm{STG}^{\mathrm{human}}$ 
& $15.8{\pm}2.7$ \\
\bottomrule
\end{tabular}
\end{table}

Human STG is positive, confirming that some source-supported transition meaning remains easier to recover from text than from generated images. This supports the central recoverability diagnosis without treating either human or VLM judgments as objective truth.

\subsection{Human--VLM Calibration}
\label{app:human_vlm_calibration}

To test whether automated recoverability scores align with human judgments on \textsc{KathaBench-25K}, we compare VLM outputs with majority human answers. Let $h_q$ be the majority human answer for question $q$, and let $v_q$ be the VLM or VLM-ensemble answer. Human--VLM agreement is
\begin{equation}
\mathrm{Acc}_{\mathrm{human}}
=
\frac{1}{|Q|}
\sum_{q \in Q}
\mathbb{I}[v_q = h_q].
\label{eq:human_vlm_acc}
\end{equation}
For continuous recoverability scores, Spearman rank correlation is
\begin{equation}
\rho_{\mathrm{sp}}
=
\mathrm{corr}
\left(
\mathrm{rank}(R^{\mathrm{VLM}}_{\mathrm{image}}),
\mathrm{rank}(R^{\mathrm{human}}_{\mathrm{image}})
\right).
\label{eq:human_vlm_spearman}
\end{equation}
When judges provide confidence scores, expected calibration error is
\begin{equation}
\mathrm{ECE}
=
\sum_{b=1}^{B}
\frac{|B_b|}{N}
\left|
\mathrm{acc}(B_b)
-
\mathrm{conf}(B_b)
\right|,
\label{eq:human_ece}
\end{equation}
where $B_b$ is confidence bin $b$, $\mathrm{acc}(B_b)$ is empirical accuracy in that bin, and $\mathrm{conf}(B_b)$ is mean confidence in that bin.

Table~\ref{tab:human_vlm_calibration_summary} reports the calibration summary for the VLM ensemble against aggregated human judgments.

\begin{table*}[t]
\centering
\scriptsize
\setlength{\tabcolsep}{4pt}
\renewcommand{\arraystretch}{0.95}
\caption{\textbf{\textsc{KathaBench-25K} human--VLM calibration.} Ensemble alignment with humans.}
\label{tab:human_vlm_calibration_summary}
\resizebox{0.82\textwidth}{!}{%
\begin{tabular}{lccc}
\toprule
\rowcolor{gray!20}
\textbf{Setting} &
\textbf{Human--VLM Acc.} $\uparrow$ &
\textbf{Spearman} $\uparrow$ &
\textbf{ECE} $\downarrow$ \\
\midrule
Calibrated VLM ensemble & 77.1\% & 0.74 & 0.081 \\
\bottomrule
\end{tabular}%
}
\end{table*}

The calibrated VLM ensemble achieves 77.1\% agreement with majority human labels, Spearman correlation $\rho_{\mathrm{sp}}=0.74$, and $\mathrm{ECE}=0.081$. These results show substantial but imperfect alignment. KathaTrace therefore uses VLMs as scalable calibrated proxies, not as final arbiters of narrative meaning.

\subsection{Preference Scale and Statistical Testing}
\label{app:human_stats}

For pairwise semantic preference and bridge-repair preference, participants use a five-point ordinal scale:
\begin{equation}
p_{ij} \in \{-2,-1,0,+1,+2\}.
\label{eq:human_preference_scale}
\end{equation}
Positive scores indicate preference for the lower-STG, KathaTrace-guided, or Semantic Compass repaired storyboard. Negative scores indicate preference for the baseline or unrepaired storyboard. A score of zero indicates no preference or comparable recoverability.

Because the preference scale is ordinal, pairwise comparisons use a one-sided Wilcoxon signed-rank test:
\begin{equation}
H_0: \mathrm{median}(p_{ij}) = 0,
\qquad
H_1: \mathrm{median}(p_{ij}) > 0.
\label{eq:wilcoxon_hypothesis}
\end{equation}
When multiple criteria are tested, Bonferroni correction is applied:
\begin{equation}
\alpha_{\mathrm{corr}}
=
\frac{\alpha}{C},
\label{eq:bonferroni}
\end{equation}
where $C$ is the number of tested criteria.

Rank-biserial correlation is reported as the effect size:
\begin{equation}
r_{\mathrm{rb}}
=
\frac{W^{+}-W^{-}}{W^{+}+W^{-}},
\label{eq:rank_biserial}
\end{equation}
where $W^{+}$ and $W^{-}$ are the positive and negative signed-rank sums.

%%%%%%%%%%%%%%%%%%%%%%%%%%%%%%%%%%%%%%%%%%%%

\section{Implementation and Reproducibility}
\label{secg}

This section summarizes the implementation, evaluation stack, planner--generator configurations, fairness controls, seed policy, compute setup, reproducibility artifacts, and practical evaluation cost used for the proposed \textsc{KathaBench-25K} benchmark. KathaTrace is an evaluation and diagnostic framework rather than a trainable storyboard generator. Reproducibility therefore depends on fixed data splits, fixed transition annotations, fixed recoverability questions, fixed accepted-answer sets, fixed evidence conditions, deterministic judge prompts, archived judge outputs, valid-subset accounting, and held-out evaluation questions.

\subsection{Evaluation Stack}
\label{app:kathatrace_stack}

KathaTrace is implemented as a modular evaluation pipeline for \textsc{KathaBench-25K}. Given a source narrative, generated storyboard, transition annotations, recoverability questions, and accepted-answer sets, the pipeline constructs text-only, image-only, and text+image evidence packets. VLM judges answer fixed questions under each evidence condition, and the scoring module computes text-side recoverability, image-side recoverability, Semantic Trajectory Gap (STG), latent dimension gaps, moral-target recoverability, ambiguity flags, valid-subset statistics, and human-calibration summaries. Table~\ref{tab:kathatrace_stack} summarizes the components of this evaluation stack and clarifies that none of them trains a new generator.

\begin{table*}[t]
\centering
\scriptsize
\setlength{\tabcolsep}{4pt}
\renewcommand{\arraystretch}{0.95}
\caption{\textbf{\textsc{KathaBench-25K} evaluation stack.} Components used for recoverability scoring.}
\label{tab:kathatrace_stack}
\resizebox{\textwidth}{!}{%
\begin{tabular}{lp{0.28\linewidth}p{0.48\linewidth}c}
\toprule
\rowcolor{gray!20}
\textbf{Component} & \textbf{Input} & \textbf{Role} & \textbf{Trainable} \\
\midrule
Story decomposition &
Source narrative $S$ &
Converts each story into ordered structured scenes. &
No \\
Transition annotation &
Adjacent scenes $(s_t,s_{t+1})$ &
Stores source-side annotation fields: action, causality, intention, emotion, consequence, and benchmark-specified moral target. &
No \\
Question builder &
Transition annotations &
Creates scored QA-dimension recoverability questions and accepted answer sets. &
No \\
Evidence-packet builder &
$S$, $\hat{X}$, questions &
Constructs text-only, image-only, and text+image judge inputs. &
No \\
Judge interface &
Evidence packets &
Queries VLM judges with fixed prompts and structured output schemas. &
No \\
Scoring module &
Judge answers and accepted answers &
Computes recoverability, STG, latent dimension gaps, moral-target recoverability, ambiguity rate, and valid-subset statistics. &
No \\
Calibration module &
Human and VLM outputs &
Computes human agreement, Spearman correlation, pairwise agreement, and calibration error. &
No \\
Semantic Compass &
Candidate storyboards and trajectory annotations &
Performs validation-selected reranking and optional bridge-scene repair. &
No \\
\bottomrule
\end{tabular}%
}
\end{table*}

The stack is generator-agnostic: any method can be evaluated if it produces an ordered image sequence for the same source narrative. KathaTrace does not require access to generator internals.

\subsection{Planner--Generator Configurations}
\label{app:planner_generator_configurations}

We evaluate each storyboard system as a planner--generator pipeline. The planner converts the source story into a generator-facing representation, and the generator renders the ordered storyboard images. All pipelines are evaluated on the same \textsc{KathaBench-25K} splits, transition annotations, recoverability questions, accepted-answer sets, evidence conditions, judge prompts, ambiguity-filtering protocol, and seed schedule where supported. Table~\ref{tab:planner_generator_architecture} lists the evaluated planner--generator interfaces and the fairness treatment applied to each configuration.

\begin{table*}[t]
\centering
\scriptsize
\setlength{\tabcolsep}{3.5pt}
\renewcommand{\arraystretch}{0.95}
\caption{\textbf{\textsc{KathaBench-25K} planner--generator settings.} Evaluated storyboard pipelines.}
\label{tab:planner_generator_architecture}
\resizebox{\textwidth}{!}{%
\begin{tabular}{lp{0.20\linewidth}p{0.18\linewidth}p{0.19\linewidth}p{0.22\linewidth}}
\toprule
\rowcolor{gray!20}
\textbf{Pipeline} & \textbf{Planner representation} & \textbf{Generator} & \textbf{Controlled inputs} & \textbf{Fairness treatment} \\
\midrule
Direct + SDXL &
Raw source narrative converted to a direct generation prompt. &
SDXL &
Same source story and chronological scene order. &
Weakest compatible prompt baseline. \\
Rule + SDXL &
Rule-structured prompts with explicit scene ordering. &
SDXL &
Same source story, scene count, and ordering. &
Tests simple structural prompting. \\
Caption + SDXL &
Caption-level description for each ordered scene. &
SDXL &
Same scene decomposition and chronological ordering. &
Tests scene-caption prompting without transition fields. \\
Scene + SDXL &
Structured scene-level prompt including entities, action, and local state. &
SDXL &
Same scene boundaries, entities, and scene order. &
Tests whether richer scene prompts improve recoverability. \\
Gemma-ST + SDXL &
Transition-aware plan with source-side transition fields: action, causality, intention, emotion, consequence, benchmark-specified moral target, generator prompts, recoverability questions, and accepted answers. &
SDXL &
Same source stories and fixed transition annotation schema. &
Tests transition-aware planning with SDXL rendering. \\
Gemma-ST + FLUX &
Same transition-aware Gemma-ST plan. &
FLUX &
Same planner outputs and scene order. &
Primary open-generator configuration. \\
Gemma-ST + StoryDiffusion &
Same transition-aware Gemma-ST plan, adapted to StoryDiffusion-compatible input format. &
StoryDiffusion &
Same source stories and planner fields where supported. &
Tests story-specialized generation under compatible inputs. \\
Gemma-ST + ConSistory &
Same transition-aware Gemma-ST plan, adapted to ConSistory-compatible input format. &
ConSistory &
Same source stories and planner fields where supported. &
Tests identity-consistency generation under compatible inputs. \\
Closed + GPT-4o + GPT-image-1 &
System-native prompting from the source story and scene plan. &
GPT-4o + GPT-image-1 &
Source story and scene order only; internal rewriting not exposed. &
Reported as a closed-system reference. \\
Closed + Nano Banana &
System-native prompting from the source story and scene plan. &
Nano Banana &
Source story and scene order only; internal rewriting not exposed. &
Reported as a closed-system reference. \\
Gemma-ST + Compass + FLUX &
Gemma-ST plan followed by validation-selected Semantic Compass reranking and optional bridge repair. &
FLUX &
Three seed-generated candidates, validation-selected Compass configuration, and held-out final questions. &
Primary actionability configuration, not a new generator. \\
Oracle planner + FLUX &
Human-gold transition plan on calibrated subset. &
FLUX &
Gold transition fields on calibrated subset only. &
Upper bound; not a deployable baseline. \\
\bottomrule
\end{tabular}%
}
\end{table*}

Closed systems are reported separately because their internal prompt rewriting, safety filtering, planning, hardware, and model-version details are not fully controllable. The oracle planner is evaluated only on a calibrated subset and is used as an upper bound, not as a directly comparable deployed method.

\subsection{Fair Comparison Controls}
\label{app:kathatrace_baseline_protocol}

Storyboard systems differ in their supported input formats. To avoid penalizing a method for interface mismatch, each planner--generator pipeline is evaluated under its strongest compatible setting while preserving the same \textsc{KathaBench-25K} source narratives, scene order, transition annotations, recoverability questions, accepted-answer sets, evidence conditions, judge prompts, and ambiguity-filtering protocol.

If a method fails to produce a complete storyboard, the example is counted as invalid for that method. Text-side recoverability is computed on the same valid subset used for the method's image-side evaluation, so STG reflects the evaluated examples. Common-valid-subset results are computed separately for stricter cross-method comparison. Table~\ref{tab:baseline_fairness_protocol} summarizes the fairness controls used for all reported planner--generator comparisons.

\begin{table*}[t]
\centering
\scriptsize
\setlength{\tabcolsep}{4pt}
\renewcommand{\arraystretch}{0.95}
\caption{\textbf{\textsc{KathaBench-25K} baseline controls.} Fairness rules for method comparison.}
\label{tab:baseline_fairness_protocol}
\resizebox{\textwidth}{!}{%
\begin{tabular}{lp{0.30\linewidth}p{0.46\linewidth}}
\toprule
\rowcolor{gray!20}
\textbf{Control} & \textbf{Protocol} & \textbf{Purpose} \\
\midrule
Story split &
Fixed train / validation / test splits. &
Prevents cherry-picking examples across methods. \\
Scene order &
All storyboards are evaluated in chronological order. &
Ensures transition questions refer to the same adjacent scene structure. \\
Seed schedule &
Open generators use generation seeds $\{42,123,2026\}$. &
Controls stochastic generation across compatible open methods. \\
Accepted answers &
Accepted answer sets are frozen before test evaluation. &
Prevents test-set leakage through answer-set revision. \\
Evidence packets &
Text-only, image-only, and text+image packets are constructed identically for all methods. &
Prevents source-text leakage into image-only evaluation. \\
Judge prompts &
Prompt templates and structured output schemas are fixed. &
Makes VLM judging auditable and repeatable. \\
Invalid generations &
Empty outputs, corrupted images, missing frames, or incomplete storyboards are logged per method. &
Ensures failures are not silently removed. \\
Method-valid subset &
Each method's Text Rec. and Image Rec. are computed on the same valid examples. &
Ensures STG is computed on paired evidence for the same stories. \\
Common-valid subset &
A stricter subset containing only examples valid for all compared methods is also computed. &
Checks whether ranking depends on method-specific failures. \\
Closed systems &
Closed/internal systems are reported as API reference baselines with model name, access date, and available version metadata. &
Avoids overclaiming fairness when internals are not controllable. \\
\bottomrule
\end{tabular}%
}
\end{table*}

\subsection{Semantic Compass Selection Protocol}
\label{app:semantic_compass_implementation}

This subsection details how Semantic Compass selects candidates, freezes validation-time settings, separates validation from held-out evaluation, and controls for sequence-length effects. Semantic Compass is a post-generation reranking and bridge-repair layer; it does not train a new generator.

\textsc{KathaBench-25K} uses three primary splits: train (3,500 stories), validation (750 stories), and held-out test (750 stories). The held-out test split is further subdivided into public-test (500) and hidden-test (250). Candidate count, score weights $\lambda_r,\lambda_t,\lambda_p$, transition-coverage threshold $\delta$, copy threshold $\gamma$, and repair threshold are selected only on the validation split and frozen before test evaluation. In the reported setup, Semantic Compass uses the three open-generator outputs produced by generation seeds $\{42,123,2026\}$ as the fixed candidate pool, so $N_c=3$.

Final Semantic Compass reporting uses held-out recoverability questions and held-out judge prompts, disjoint from validation-time selection. This separation prevents test STG from being optimized by the same questions or prompts used for reranking or repair. Because bridge repair changes sequence length, we compare against length-controlled alternatives: random bridge insertion, non-semantic extra-frame insertion, caption-only bridge insertion, candidate reranking alone, and bridge repair alone. Table~\ref{tab:semantic_compass_selection_protocol} summarizes the selection quantities, where they are chosen, and how they are used in held-out reporting.

\begin{table*}[t]
\centering
\scriptsize
\setlength{\tabcolsep}{4pt}
\renewcommand{\arraystretch}{0.95}
\caption{\textbf{\textsc{KathaBench-25K} Compass protocol.} Validation-selected settings and controls.}
\label{tab:semantic_compass_selection_protocol}
\resizebox{\textwidth}{!}{%
\begin{tabular}{lp{0.28\linewidth}p{0.46\linewidth}}
\toprule
\rowcolor{gray!20}
\textbf{Quantity} & \textbf{Value / selection source} & \textbf{Purpose} \\
\midrule
Candidate count $N_c$ & 3 candidates per story & One candidate from each generation seed: 42, 123, and 2026. \\
Recoverability weight $\lambda_r$ & Validation-selected; frozen before test evaluation & Rewards image-only recoverability. \\
Transition weight $\lambda_t$ & Validation-selected; frozen before test evaluation & Rewards visual support for annotated transitions. \\
Copy penalty $\lambda_p$ & Validation-selected; frozen before test evaluation & Penalizes repeated or near-duplicate frames. \\
Transition threshold $\delta$ & Validation-selected; frozen before test evaluation & Determines whether an annotated transition is visually supported. \\
Copy threshold $\gamma$ & Validation-selected; frozen before test evaluation & Detects near-duplicate adjacent frames. \\
Repair threshold & Validation-selected; frozen before test evaluation & Decides whether the weakest transition receives a bridge frame. \\
Selection questions & Validation questions only & Used for candidate selection and repair decisions, not final reporting. \\
Final reporting questions & Held-out test questions & Used only for final recoverability and STG reporting. \\
Judge prompts & Held-out judge-prompt variants for final reporting & Reduces prompt-specific overfitting. \\
Length controls & Random bridge, non-semantic extra frame, caption-only bridge, rerank only, bridge only & Tests whether gains come from semantic repair rather than sequence length. \\
\bottomrule
\end{tabular}%
}
\end{table*}

\subsection{Seeds, Runs, and Determinism}
\label{app:runs_seeds_determinism}

All deterministic preprocessing, split construction, question ordering, bootstrap sampling, and candidate ordering use master seed 42. For stochastic open-generator outputs, we use three generation seeds, $\{42,123,2026\}$, and report story-level means with 95\% bootstrap confidence intervals computed using 10,000 bootstrap resamples. VLM judging uses deterministic decoding and fixed output schemas; therefore each judge is run once per evidence packet, and raw judge outputs are archived for reproducibility. Open-generator and open-VLM experiments are run on NVIDIA A100 80GB GPUs. Closed systems are reported as API-based reference baselines with model name, access date, and available version metadata, since their internal hardware, prompt rewriting, and model versions are not fully controllable. Table~\ref{tab:runs_seeds} reports the run and seed configuration used in \textsc{KathaBench-25K}.

\begin{table*}[t]
\centering
\scriptsize
\setlength{\tabcolsep}{4pt}
\renewcommand{\arraystretch}{0.95}
\caption{\textbf{\textsc{KathaBench-25K} run seeds.} Determinism and generation settings.}
\label{tab:runs_seeds}
\resizebox{\textwidth}{!}{%
\begin{tabular}{lp{0.18\linewidth}p{0.18\linewidth}p{0.24\linewidth}p{0.20\linewidth}}
\toprule
\rowcolor{gray!20}
\textbf{Component / pipeline} & \textbf{Runs / story} & \textbf{Seed values} & \textbf{Selection / reporting rule} & \textbf{Notes} \\
\midrule
Split creation & 1 deterministic run & Master seed 42 & Fixed train / validation / test split files. & Shared by all methods. \\
Question ordering & 1 deterministic run & Master seed 42 & Fixed question order for judge packets. & Prevents order drift. \\
Bootstrap CI & 10,000 resamples & Master seed 42 & 95\% confidence intervals over stories. & Used for reported uncertainty. \\
Direct + SDXL & 3 generations & 42, 123, 2026 & Story-level mean across seeds. & Open generator; seed-controlled. \\
Rule + SDXL & 3 generations & 42, 123, 2026 & Story-level mean across seeds. & Open generator; seed-controlled. \\
Caption + SDXL & 3 generations & 42, 123, 2026 & Story-level mean across seeds. & Open generator; seed-controlled. \\
Scene + SDXL & 3 generations & 42, 123, 2026 & Story-level mean across seeds. & Open generator; seed-controlled. \\
Gemma-ST + SDXL & 3 generations & 42, 123, 2026 & Story-level mean across seeds. & Open generator; seed-controlled. \\
Gemma-ST + FLUX & 3 generations & 42, 123, 2026 & Story-level mean across seeds. & Primary open-generator setting. \\
Gemma-ST + StoryDiffusion & 3 generations & 42, 123, 2026 & Story-level mean across seeds when seed control is supported. & Story-specialized open method. \\
Gemma-ST + ConSistory & 3 generations & 42, 123, 2026 & Story-level mean across seeds when seed control is supported. & Identity-consistency open method. \\
Bridge-frame generation & 3 generations & 42, 123, 2026 & Same seed schedule as open-generator outputs. & Used for Semantic Compass repair candidates. \\
Semantic Compass candidate generation & 3 candidates & 42, 123, 2026 & Validation-selected reranking and bridge-repair rule. & $N_c=3$ fixed candidates/story. \\
VLM judging & 1 deterministic pass per judge & Fixed decoding; no sampling seed required & One pass for each judge and evidence packet. & Raw outputs archived. \\
Human evaluation & 1 annotation round & Not applicable & Includes overlap and adjudication subsets. & Human outputs archived. \\
Closed API systems & 1 API run & Not exposed & System-native output retained; model/version/date logged. & Internal randomness not controllable. \\
\bottomrule
\end{tabular}%
}
\end{table*}

\subsection{Compute Environment}
\label{app:kathatrace_compute}

KathaTrace does not require training a large model. Its main compute cost comes from storyboard generation and VLM judging. Cached judge outputs are reused for recoverability scoring, ambiguity analysis, latent dimension-gap computation, valid-subset accounting, and calibration. Table~\ref{tab:hardware_software_environment} summarizes the compute environment, seed settings, and execution modes used for \textsc{KathaBench-25K} experiments.

\begin{table*}[t]
\centering
\scriptsize
\setlength{\tabcolsep}{4pt}
\renewcommand{\arraystretch}{0.95}
\caption{\textbf{\textsc{KathaBench-25K} compute setup.} Hardware, seeds, and execution modes.}
\label{tab:hardware_software_environment}
\resizebox{\textwidth}{!}{%
\begin{tabular}{lp{0.30\linewidth}p{0.44\linewidth}}
\toprule
\rowcolor{gray!20}
\textbf{Item} & \textbf{Reported value} & \textbf{Use} \\
\midrule
Primary GPU & NVIDIA A100 80GB & Open-generator inference and open-weight VLM evaluation. \\
Open-generator hardware & NVIDIA A100 80GB & SDXL, FLUX, StoryDiffusion, ConSistory, and bridge-frame generation. \\
Open-VLM hardware & NVIDIA A100 80GB & Qwen2.5-VL-7B, Qwen2.5-VL-3B, SmolVLM2-2.2B, and SmolVLM2-500M evaluation where locally executed. \\
Closed API systems & API execution; hardware not exposed & GPT-4o + GPT-image-1 and Nano Banana are reported with model name, access date, and available version metadata. \\
Master seed & 42 & Split creation, question ordering, bootstrap sampling, candidate ordering, and deterministic sampling. \\
Generation seeds & 42, 123, 2026 & Stochastic open-generator outputs and bridge-frame generation. \\
VLM judging runs & 1 deterministic pass per judge & Fixed prompts, fixed evidence packets, fixed output schemas, and archived raw outputs. \\
Bootstrap resampling & 10,000 resamples, seed 42 & 95\% bootstrap confidence intervals over stories. \\
Human evaluation & 1 collected annotation round with overlap and adjudication & Human QA, calibration, preference, and bridge-repair evaluation. \\
Closed-system runs & 1 API run per prompt & Output retained with model/version/date metadata where available. \\
\bottomrule
\end{tabular}%
}
\end{table*}

The number of judge calls scales as:
\begin{equation}
C_{\mathrm{judge}}
=
N_{\mathrm{stories}}
\cdot
N_{\mathrm{methods}}
\cdot
N_{\mathrm{questions}}
\cdot
N_{\mathrm{conditions}}
\cdot
N_{\mathrm{judges}},
\label{eq:impl_judge_cost}
\end{equation}
where $N_{\mathrm{conditions}}=3$ for text-only, image-only, and text+image evaluation. For the reported benchmark configuration, \textsc{KathaBench-25K} contains 5,000 narratives, 25,000 structured scenes, 20,000 transition annotations, 28,712 recoverability questions, 10,000 contrastive variants, and six scored QA dimensions. The split is 3,500 train, 750 validation, and 750 held-out test examples. Main planner--generator tables are reported on the held-out evaluation split with method-specific valid-subset accounting.

\subsection{Reproducibility Checklist}
\label{app:kathatrace_reproducibility_controls}

The held-out test split contains 750 stories and is subdivided into public-test (500) and hidden-test (250). References to test (750) denote the aggregate held-out test set. Table~\ref{tab:kathatrace_reproducibility_checklist} lists the artifacts required to audit the \textsc{KathaBench-25K} benchmark, evaluation protocol, Semantic Compass selection procedure, and reported metrics.

\begin{table*}[t]
\centering
\scriptsize
\setlength{\tabcolsep}{4pt}
\renewcommand{\arraystretch}{0.95}
\caption{\textbf{\textsc{KathaBench-25K} reproducibility checklist.} Artifacts for auditing results.}
\label{tab:kathatrace_reproducibility_checklist}
\resizebox{\textwidth}{!}{%
\begin{tabular}{lp{0.22\linewidth}p{0.50\linewidth}}
\toprule
\rowcolor{gray!20}
\textbf{Artifact} & \textbf{Reported count / setting} & \textbf{Purpose} \\
\midrule
Split files &
3,500 / 750 / 750 &
Fix train, validation and test partitions. \\
Source narratives &
5,000 narratives &
Define stories used for generation and text-side evaluation. \\
Scene decompositions &
25,000 structured scenes; avg. 5 scenes/story &
Fix scene boundaries and chronological story structure. \\
Transition annotations &
20,000 transitions; avg. 4 transitions/story &
Define source-side transition fields: action, causality, intention, emotion, consequence, and benchmark-specified moral target. \\
Transition annotation fields &
6 fields &
Action, causality, intention, emotion, consequence, and benchmark-specified moral target. \\
Scored QA dimensions &
6 dimensions &
Action visibility, causal recoverability, emotional shift, consequence, temporal order, and moral-target recoverability. \\
Main latent-gap dimensions &
4 dimensions &
Causal, emotional, consequence, and moral-target gaps. Action visibility and temporal order are reported separately as local/control dimensions. \\
Recoverability questions &
28,712 questions &
Fix dimension-specific evaluation queries. \\
Contrastive variants &
10,000 variants &
Enable object-overlap and semantic-drift controls. \\
Evidence conditions &
3 conditions &
Text-only, image-only, and text+image evaluation. \\
Judge models &
6 VLM judges &
Gemini 2.5 Flash, Claude Haiku 4.5, Qwen2.5-VL-7B, Qwen2.5-VL-3B, SmolVLM2-2.2B, and SmolVLM2-500M. \\
Master seed &
42 &
Controls split creation, question ordering, bootstrap sampling, candidate ordering, and deterministic sampling. \\
Generation seeds &
42, 123, 2026 &
Controls stochastic open-generator outputs and bridge-frame generation. \\
Bootstrap configuration &
10,000 resamples, seed 42 &
Produces 95\% confidence intervals over stories. \\
VLM judging &
1 deterministic pass per judge &
Uses fixed decoding, fixed prompts, fixed output schemas, and archived raw outputs. \\
Held-out evaluation split &
750 validation examples; 750 test examples &
Validation is used for selection and thresholds; test is used for final reporting. \\
Planner settings &
5 planners + oracle subset &
Direct, rule, caption, scene, Gemma-ST, and oracle upper-bound planning. \\
Generator settings &
6 generator settings &
SDXL, FLUX, StoryDiffusion, ConSistory, GPT-4o + GPT-image-1, and Nano Banana. \\
Evaluator baselines &
6 evaluator settings &
Object-overlap QA, generic story QA, multi-frame TIFA-style QA, multi-frame DSG-style QA, KathaTrace without contrastives, and KathaTrace without dimension-specific questions. \\
Length-control baselines &
6 repair/control settings &
Base pipeline, random extra frame, non-semantic extra frame, caption-only bridge, candidate rerank only, and bridge-scene repair only. \\
Open-generation hardware &
NVIDIA A100 80GB &
Runs SDXL, FLUX, StoryDiffusion, ConSistory, and bridge-frame generation. \\
Closed-system metadata &
Model name, access date, and available version metadata &
Documents GPT-4o + GPT-image-1 and Nano Banana API-based reference baselines. \\
Invalid-generation logs &
Per-method valid/invalid counts &
Track empty outputs, corrupted images, missing frames, and incomplete storyboards. \\
Accepted answer sets &
Frozen before test evaluation &
Make answer matching deterministic and prevent test-set leakage. \\
Judge prompts and output schemas &
Fixed per evidence condition &
Prevent source-text leakage and make outputs scoreable. \\
Raw judge outputs and normalized answers &
Stored for all scored questions &
Allow recomputation of recoverability, STG, latent gaps, moral-target recoverability, and ambiguity rates. \\
Human-evaluation forms and scripts &
Stored with preference and QA outputs &
Reproduce human QA, pairwise preference, bridge-repair trials, confidence intervals, and agreement metrics. \\
Semantic Compass config &
Validation-selected and frozen &
Stores $N_c=3$, $\lambda_r,\lambda_t,\lambda_p$, $\delta$, $\gamma$, repair threshold, held-out question set, and held-out judge-prompt set. \\
\bottomrule
\end{tabular}%
}
\end{table*}

For deterministic reporting, we store generated storyboard paths, split identifiers, scene decompositions, transition annotations, valid-subset masks, judge responses, normalized answers, match decisions, ambiguity flags, repair decisions, seed values, candidate order, and final metric tables. Each reported result can therefore be traced to the corresponding story, planner, generator, seed, question, evidence condition, judge, accepted answer set, and repair configuration.

\subsection{Practical Evaluation Cost}
\label{app:practical_evaluation_cost}

KathaTrace deliberately trades lightweight caption-style scoring for diagnostic recoverability evaluation. Its evaluation cost comes from testing each recoverability question under fixed evidence conditions, normalizing structured judge outputs, matching responses against frozen accepted-answer sets, and aggregating transition-level metrics. This design is more expensive than caption or object-overlap metrics, but it is necessary for measuring whether source-supported transition meaning remains recoverable from generated storyboards.

The released \textsc{KathaBench-25K} benchmark contains 5,000 stories, 25,000 annotated scenes, 20,000 adjacent transitions, and 28,712 recoverability questions. Evaluating each question under the three evidence conditions---text-only, image-only, and text+image---requires 86,136 judge calls for a single judge. The cost scales linearly with the number of judges; for example, a three-judge ensemble requires 258,408 judge calls. Storyboard generation is evaluated separately from judge calls and requires 25,000 generated panels for the full benchmark, 5,000 panels for \texttt{human\_gold\_1k.jsonl}, and 2,000 panels for \texttt{strict\_gold\_400.jsonl}.

To support different compute budgets, we release two evaluation subsets in addition to the full benchmark. Table~\ref{tab:evaluation_budget_levels} reports the number of stories, panels, QA pairs, and judge calls for the strict, human-gold, and full \textsc{KathaBench-25K} evaluation settings. The \texttt{human\_gold\_1k.jsonl} subset provides a medium-scale evaluation setting, while \texttt{strict\_gold\_400.jsonl} provides a rapid high-confidence setting for implementation checks, method debugging, and low-budget reproducibility. For this low-budget setting, we recommend reporting the same evidence conditions, judge model, valid-subset accounting, and bootstrap confidence intervals used in the full benchmark. The strict subset is not a replacement for full-benchmark reporting, but it provides a fixed smoke test for verifying implementation correctness before running the complete evaluation.

\begin{table}[t]
\centering
\scriptsize
\setlength{\tabcolsep}{3.5pt}
\renewcommand{\arraystretch}{1.05}
\caption{\textbf{\textsc{KathaBench-25K} evaluation budgets.} Judge calls for one judge.}
\label{tab:evaluation_budget_levels}
\resizebox{\linewidth}{!}{
\begin{tabular}{lrrrr}
\toprule
\rowcolor{gray!20}
\textbf{Setting} & \textbf{Stories} & \textbf{Panels} & \textbf{QA pairs} & \textbf{Judge calls} \\
\midrule
\texttt{strict\_gold\_400} & 400 & 2,000 & $\sim$2,297 & $\sim$6,891 \\
\rowcolor{gray!5}
\texttt{human\_gold\_1k} & 1,000 & 5,000 & $\sim$5,742 & $\sim$17,226 \\
Full benchmark & 5,000 & 25,000 & 28,712 & 86,136 \\
\bottomrule
\end{tabular}
}
\vspace{-4pt}
\end{table}

Since transition annotations, recoverability questions, accepted-answer sets, and evaluation metadata are precomputed, evaluating a new method only requires generating storyboards and running the fixed protocol. We recommend reporting the evaluation subset, number of generated panels, judge calls, judges, evidence conditions, and confidence intervals together with STG.

\paragraph{Leaderboard evaluation cost.}
For leaderboard-style use, the held-out test split is divided into a public-test subset
and a hidden-test subset. The public-test subset supports open comparison and debugging,
while the hidden-test subset supports leaderboard-style evaluation without exposing final
test examples. We report leaderboard cost in generated panels and judge calls rather than
wall-clock time, because wall-clock runtime depends on the submitted generator, batching
strategy, API latency, and hardware. Once a method has produced the required storyboard
panels, the benchmark-side evaluation cost is fixed by the number of stories, QA pairs,
evidence conditions, and judges.

\begin{table}[t]
\centering
\scriptsize
\setlength{\tabcolsep}{3.5pt}
\renewcommand{\arraystretch}{1.05}
\caption{\textbf{\textsc{KathaBench-25K} leaderboard cost.} Approximate panel and one-judge evaluation cost.}
\label{tab:leaderboard_cost}
\resizebox{\linewidth}{!}{
\begin{tabular}{lrrrr}
\toprule
\rowcolor{gray!20}
\textbf{Setting} & \textbf{Stories} & \textbf{Panels} & \textbf{QA pairs} & \textbf{Judge calls} \\
\midrule
Public-test & 500 & 2{,}500 & $\sim$2{,}871 & $\sim$8{,}613 \\
\rowcolor{gray!5}
Hidden-test & 250 & 1{,}250 & $\sim$1{,}436 & $\sim$4{,}308 \\
Full held-out test & 750 & 3{,}750 & $\sim$4{,}307 & $\sim$12{,}921 \\
\bottomrule
\end{tabular}
}
\vspace{-4pt}
\end{table}

Table~\ref{tab:leaderboard_cost} separates generator-side cost from benchmark-side
evaluation cost. Generator-side cost is the number of storyboard panels a submitted
method must produce. Benchmark-side cost is the number of judge calls required after
the panels are generated. For one judge, the judge-call count is approximately three
times the number of QA pairs because KathaTrace evaluates text-only, image-only, and
text+image evidence conditions. For multi-judge evaluation, the values in
Table~\ref{tab:leaderboard_cost} scale linearly with the number of judges.

\subsection{Community Adoption and Reuse}
\label{app:community_adoption}

KathaBench-25K is designed so that a new storyboard generator can be evaluated without changing the benchmark. For example, a researcher proposing a new StoryDiffusion-style, FLUX-based, or consistency-aware storyboard method can take the same source stories from \texttt{annotations.jsonl}, generate five ordered panels for each story, and run the fixed KathaTrace judge packets. The evaluator does not need access to the generator's model weights, hidden prompts, training data, or intermediate plans. It only needs the generated storyboard images, the released story identifier, the fixed recoverability questions, and the frozen accepted-answer sets.

This makes comparison practical and diagnostic. If a new method improves character consistency but still omits the key rescue action in a reciprocity story, KathaTrace will show a high consequence or moral-target gap rather than only reporting that the images look coherent. If a method preserves the main characters in \emph{The Monkey and the Crocodile} but fails to communicate the crocodile's hidden betrayal, the image-only causal or moral-target recoverability will remain low. If a model adds visually pleasing extra frames but does not repair the missing transition, the length-control protocol separates that from true semantic improvement. Thus, future users can diagnose whether a method improves visual quality, entity consistency, causal recovery, emotional recovery, consequence recovery, or moral-target recovery.

The release also supports different evaluation budgets. A full evaluation uses 5,000 stories, 25,000 panels, 28,712 recoverability questions, and 86,136 judge calls for one judge across the three evidence conditions. A medium-cost evaluation can use \texttt{human\_gold\_1k.jsonl}, with 1,000 stories and approximately 17,226 one-judge calls. A rapid smoke test can use \texttt{strict\_gold\_400.jsonl}, with 400 stories and approximately 6,891 one-judge calls. These subsets let researchers debug their pipeline before running the full benchmark, while still using the same evidence isolation, accepted-answer matching, ambiguity filtering, and STG computation.

The reusable artifacts are intended to make results auditable rather than one-off. The release includes fixed train/validation/test splits, source narratives, five-scene decompositions, adjacent-scene transition annotations, recoverability questions, accepted-answer sets, contrastive variants, human-validation metadata, ambiguity flags, and evaluation subsets. A future paper can therefore report not only an aggregate STG score, but also causal, emotional, consequence, and moral-target gaps, ambiguity rate, valid-subset size, and common-valid STG. This allows the community to compare new visual narrative generators under the same protocol as the field improves, making KathaBench-25K a reusable diagnostic benchmark rather than only a static story-prompt collection.

%%%%%%%%%%%%%%%%%%%%%%%%%%%

\section{Human Perception Study}
\label{sec:human_perception_study}

This section reports the human perception study used to validate whether KathaTrace-guided outputs on \textsc{KathaBench-25K} are perceived as preserving narrative meaning better than matched baselines. The study is separate from the dataset-construction review in Sec.~\ref{seca} and the human--VLM calibration study in Sec.~\ref{secf}. Here, participants compare anonymized storyboards and rate whether causal, emotional, consequence-bearing, moral-target, and overall transition meaning is visually recoverable. Human judgments are used as perception evidence, not as objective narrative-truth labels.

\subsection{Study Protocol}
\label{app:human_study_design}

Participants completed image-only preference and recoverability tasks on \textsc{KathaBench-25K} storyboards. In the primary condition, they saw generated storyboard images only. They did not see the source story, generation prompts, captions, transition annotations, moral labels, method names, or model names. This prevents source-text leakage and tests whether the intended transition meaning is visually recoverable from the storyboard itself. Table~\ref{tab:human_perception_protocol} summarizes the study tasks, evidence shown to participants, and the purpose of each task.

For pairwise trials, participants compared two anonymized storyboards generated for the same \textsc{KathaBench-25K} source story. For bridge-repair trials, participants compared an original storyboard with a Semantic Compass repaired storyboard in randomized left--right order. Story order, question order, answer-option order, and left--right placement were randomized. Participants could select \texttt{Unclear} or no preference when the available evidence did not support a decision.

\begin{table*}[t]
\centering
\scriptsize
\setlength{\tabcolsep}{4pt}
\renewcommand{\arraystretch}{1.05}
\caption{\textbf{\textsc{KathaBench-25K} perception protocol.} Image-only tasks for human preference.}
\label{tab:human_perception_protocol}
\resizebox{\textwidth}{!}{
\begin{tabular}{p{0.22\linewidth} p{0.36\linewidth} p{0.34\linewidth}}
\toprule
\rowcolor{gray!20}
\textbf{Task} & \textbf{Evidence shown} & \textbf{Purpose} \\
\midrule
Image-only recoverability &
Generated storyboard images only. &
Tests whether transition meaning is recoverable without source-side evidence. \\
\rowcolor{gray!5}
Pairwise semantic preference &
Two anonymized storyboards for the same source story. &
Tests whether humans prefer the lower-STG or KathaTrace-guided output. \\
Bridge-repair preference &
Original storyboard and repaired storyboard in randomized order. &
Tests whether localized bridge repair improves perceived transition recoverability. \\
\rowcolor{gray!5}
Quality-control checks &
Repeated or attention-check trials. &
Filters inattentive, inconsistent, or unrealistically fast responses. \\
\bottomrule
\end{tabular}
}
\end{table*}

\subsection{Preference Scale}
\label{app:human_preference_scale}

Pairwise and bridge-repair trials use the five-point ordinal preference scale in Eq.~\ref{eq:human_preference_scale_final}. Positive scores favor the lower-STG, KathaTrace-guided, or Semantic Compass repaired storyboard. Negative scores favor the baseline or unrepaired storyboard. A score of zero indicates no preference or comparable recoverability. Table~\ref{tab:human_preference_scale_final} gives the interpretation of each score used in the \textsc{KathaBench-25K} perception study.

\begin{equation}
p_{ij} \in \{-2,-1,0,+1,+2\}.
\label{eq:human_preference_scale_final}
\end{equation}

\begin{table}[t]
\centering
\scriptsize
\setlength{\tabcolsep}{5pt}
\renewcommand{\arraystretch}{1.05}
\caption{\textbf{\textsc{KathaBench-25K} preference scale.} Ordinal scores for pairwise trials.}
\label{tab:human_preference_scale_final}
\begin{tabular}{c p{0.70\linewidth}}
\toprule
\rowcolor{gray!20}
\textbf{Score} & \textbf{Interpretation} \\
\midrule
$+2$ & Strongly prefer the KathaTrace-guided or repaired storyboard. \\
\rowcolor{gray!5}
$+1$ & Slightly prefer the KathaTrace-guided or repaired storyboard. \\
$0$ & No preference, or both storyboards are comparably recoverable. \\
\rowcolor{gray!5}
$-1$ & Slightly prefer the baseline or unrepaired storyboard. \\
$-2$ & Strongly prefer the baseline or unrepaired storyboard. \\
\bottomrule
\end{tabular}
\end{table}

\subsection{Perception Criteria}
\label{app:human_preference_questions}

The \textsc{KathaBench-25K} perception study uses 15 criteria. These criteria are broader than the released scored QA inventory and are used only for human preference analysis. Intention-related criteria are included because participants may naturally infer character goals from images; however, intention remains an annotation and planning field, not a released scored QA dimension. The scored QA inventory remains \texttt{action\_visibility}, \texttt{causal}, \texttt{emotional}, \texttt{consequence}, \texttt{temporal\_order}, and \texttt{moral}. Table~\ref{tab:human_preference_questions} lists the 15 perception-study questions shown to participants.

\begin{table*}[t]
\centering
\scriptsize
\setlength{\tabcolsep}{4pt}
\renewcommand{\arraystretch}{1.05}
\caption{\textbf{\textsc{KathaBench-25K} perception criteria.} Human preference questions by dimension.}
\label{tab:human_preference_questions}
\resizebox{\textwidth}{!}{
\begin{tabular}{c p{0.20\linewidth} p{0.58\linewidth}}
\toprule
\rowcolor{gray!20}
\textbf{ID} & \textbf{Dimension} & \textbf{Question shown to participants} \\
\midrule
Q1 & Causality & Can you tell why the later scene follows from the earlier scene? \\
\rowcolor{gray!5}
Q2 & Causality & Does the image sequence make the cause--effect relation clear? \\
Q3 & Intention-related perception & Can you infer what the main character is trying or planning to do? \\
\rowcolor{gray!5}
Q4 & Intention-related perception & Does the storyboard make the character's goal or motivation visually understandable? \\
Q5 & Emotion & Can you infer how the character's emotional state changes across the sequence? \\
\rowcolor{gray!5}
Q6 & Emotion & Does the storyboard visually support the emotional shift rather than leaving it unclear? \\
Q7 & Consequence & Can you tell what outcome results from the character's action or decision? \\
\rowcolor{gray!5}
Q8 & Consequence & Does the sequence make the consequence of the earlier event visually recoverable? \\
Q9 & Moral-target & Can you infer the intended lesson or moral meaning from the image sequence alone? \\
\rowcolor{gray!5}
Q10 & Moral-target & Does the storyboard visually support the intended moral meaning rather than a different interpretation? \\
Q11 & Trajectory & Does the sequence preserve a clear progression from one scene to the next? \\
\rowcolor{gray!5}
Q12 & Overall recoverability & Without seeing the source story, how well can you recover the main transition meaning from the images? \\
Q13 & Visual-coherence control & Are the images coherent in characters, setting, style, and scene continuity? \\
\rowcolor{gray!5}
Q14 & Bridge-repair preference & If an extra bridge scene is shown, does it make the missing transition meaning easier to recover? \\
Q15 & Overall preference & Overall, which storyboard better preserves the story's meaning while remaining visually coherent? \\
\bottomrule
\end{tabular}
}
\end{table*}

\subsection{Preference Aggregation and Statistical Testing}
\label{app:human_preference_aggregation_final}

For criterion $c$, the mean preference score is
\begin{equation}
\mu_c
=
\frac{1}{N_c}
\sum_{(i,j)\in D_c}
p_{ij},
\label{eq:human_preference_mean_final}
\end{equation}
where $D_c$ is the set of participant judgments for criterion $c$. We report mean preference with 95\% confidence intervals using participant-level bootstrap resampling.

Because preference scores are ordinal, significance is tested using a one-sided Wilcoxon signed-rank test:
\begin{equation}
H_0:\mathrm{median}(p_{ij})=0,
\qquad
H_1:\mathrm{median}(p_{ij})>0.
\label{eq:human_wilcoxon_final}
\end{equation}
A positive median indicates preference for the lower-STG, KathaTrace-guided, or repaired storyboard. For multiple criteria, we apply Bonferroni correction:
\begin{equation}
\alpha_{\mathrm{corr}}
=
\frac{\alpha}{C},
\label{eq:human_bonferroni_final}
\end{equation}
where $C$ is the number of tested criteria. Rank-biserial correlation is reported as the ordinal effect size:
\begin{equation}
r_{\mathrm{rb}}
=
\frac{W^+ - W^-}{W^+ + W^-},
\label{eq:human_rank_biserial_final}
\end{equation}
where $W^+$ and $W^-$ are the positive and negative signed-rank sums.

\subsection{Human Preference Results}
\label{app:human_preference_results}

\begin{figure*}[!t]
\centering
\includegraphics[width=\textwidth]{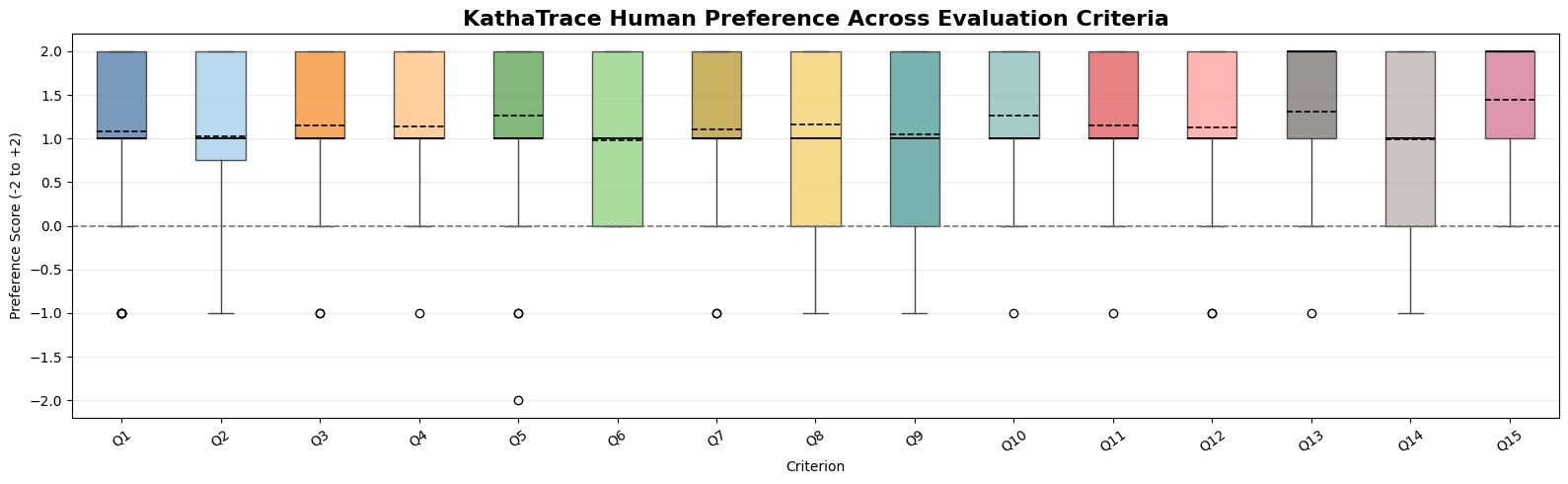}
\caption{\textbf{\textsc{KathaBench-25K} preference distributions.} Human scores across perception criteria.}
\label{fig:human_preference_boxplot}
\end{figure*}

\begin{figure*}[!t]
\centering
\includegraphics[width=\textwidth]{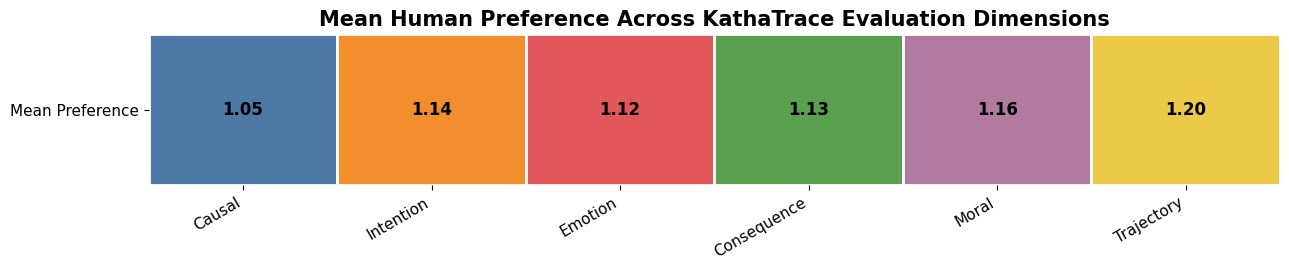}
\caption{\textbf{\textsc{KathaBench-25K} preference density.} Violin view of human ratings.}
\label{fig:human_preference_violin}
\end{figure*}

\begin{figure*}[!t]
\centering
\includegraphics[width=0.85\textwidth]{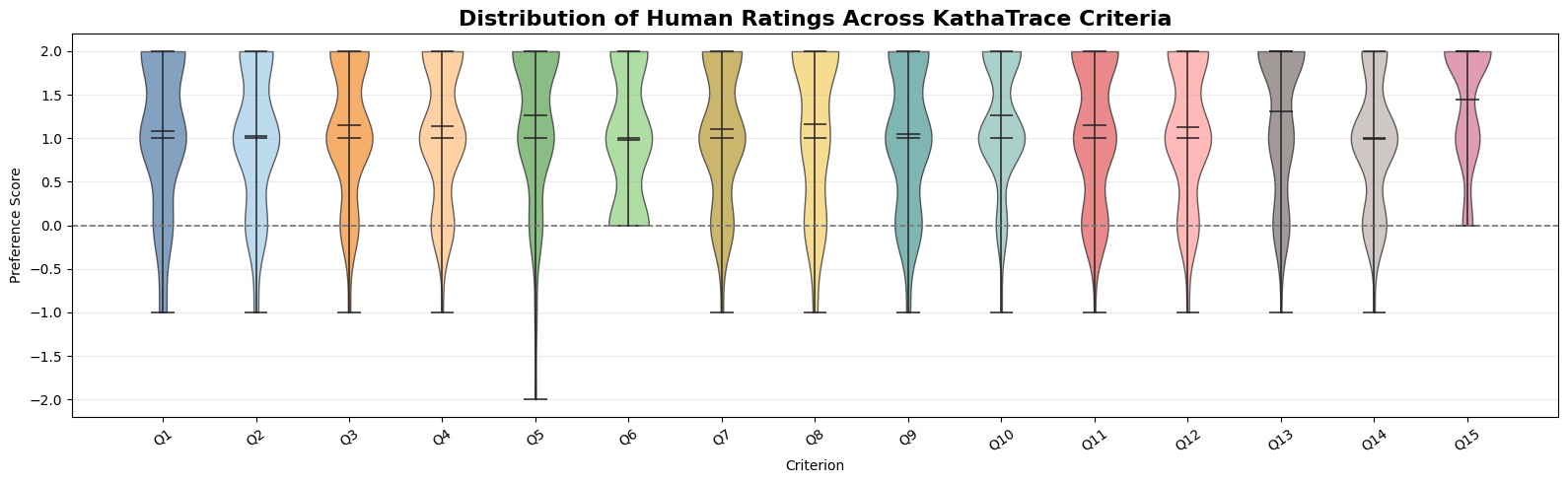}
\caption{\textbf{\textsc{KathaBench-25K} dimension preferences.} Mean preference by perception group.}
\label{fig:human_dimension_heatmap}
\end{figure*}

\begin{figure*}[!t]
\centering
\includegraphics[width=0.86\textwidth]{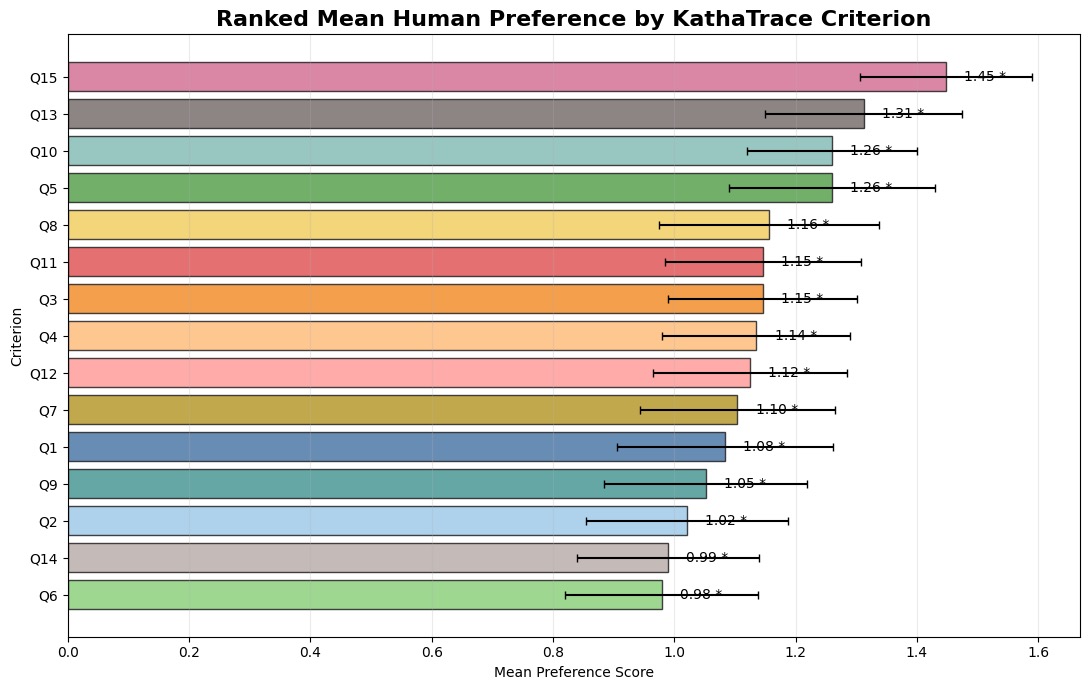}
\caption{\textbf{\textsc{KathaBench-25K} ranked preferences.} Mean score by criterion.}
\label{fig:human_preference_ranked}
\end{figure*}

\begin{figure*}[!t]
\centering
\includegraphics[width=0.86\textwidth]{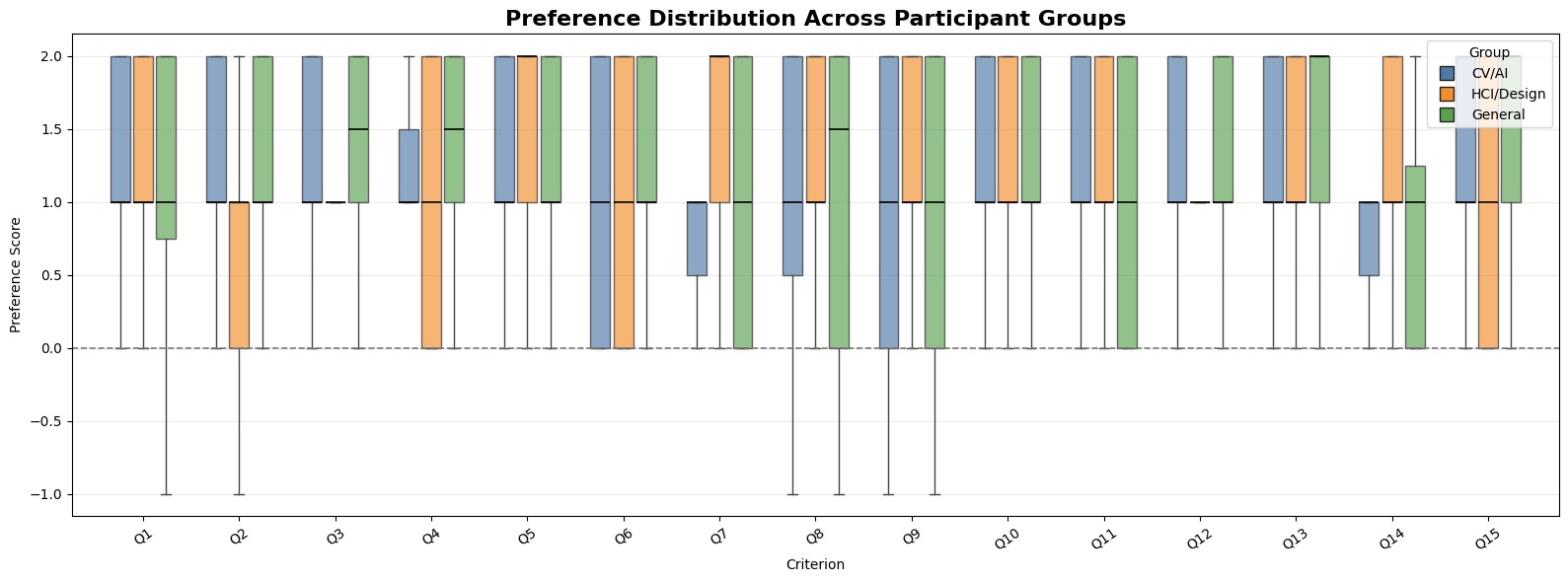}
\caption{\textbf{\textsc{KathaBench-25K} preference significance.} Corrected tests with confidence intervals.}
\label{fig:human_preference_significance}
\end{figure*}

\begin{figure*}[!t]
\centering
\includegraphics[width=\textwidth]{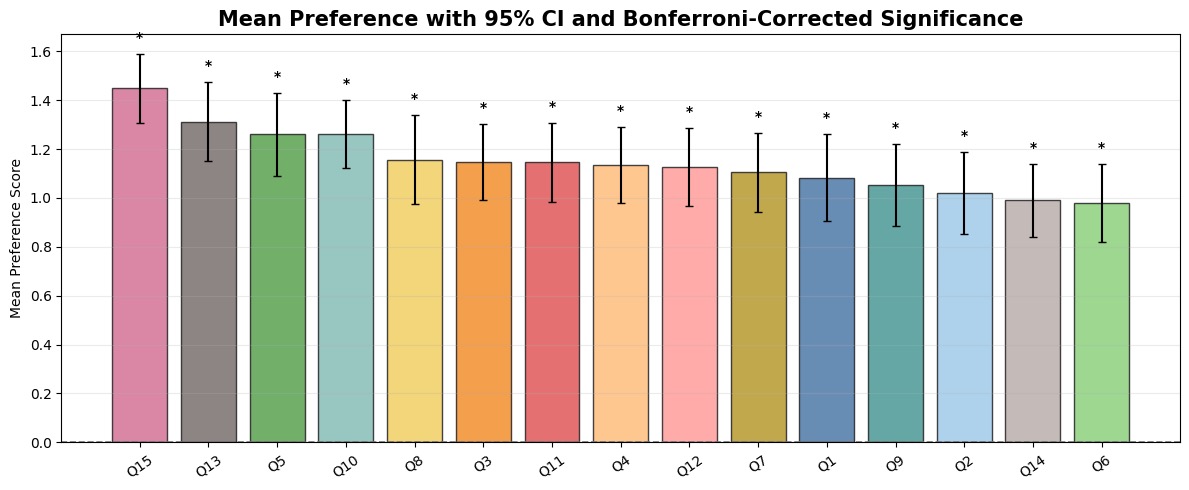}
\caption{\textbf{\textsc{KathaBench-25K} group preferences.} Robustness view across participant groups.}
\label{fig:human_group_boxplot}
\end{figure*}

Figures~\ref{fig:human_preference_boxplot}--\ref{fig:human_group_boxplot} summarize the \textsc{KathaBench-25K} perception-study results. Scores above zero indicate preference for the lower-STG, KathaTrace-guided, or repaired storyboard. The distributional plots in Figs.~\ref{fig:human_preference_boxplot} and~\ref{fig:human_preference_violin} show that preferences generally shift above the no-preference baseline. The grouped and ranked summaries in Figs.~\ref{fig:human_dimension_heatmap}--\ref{fig:human_preference_significance} show that the strongest preferences occur for overall meaning preservation, visual coherence, moral-target support, and transition recoverability. Figure~\ref{fig:human_group_boxplot} provides an additional robustness view across participant groups.

\subsection{Human--VLM Preference Calibration}
\label{app:human_vlm_calibration_final}

\begin{figure}[t]
\centering
\includegraphics[width=0.78\linewidth]{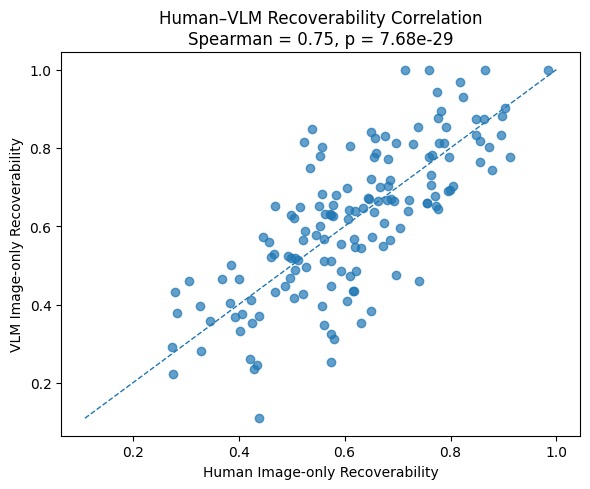}
\caption{\textbf{\textsc{KathaBench-25K} human--VLM calibration.} Image-only recoverability correlation.}
\label{fig:human_vlm_recoverability_correlation}
\end{figure}

Human--VLM preference calibration tests whether automated image-only recoverability follows human preference trends on \textsc{KathaBench-25K}. Let $h_q$ be the majority human answer or preference for item $q$, and let $v_q$ be the VLM or VLM-ensemble answer. Human--VLM agreement is
\begin{equation}
\mathrm{Acc}_{\mathrm{human}}
=
\frac{1}{|Q|}
\sum_{q\in Q}
\mathbb{I}[v_q=h_q].
\label{eq:human_vlm_accuracy_final}
\end{equation}
For continuous recoverability scores, Spearman rank correlation is
\begin{equation}
\rho_{\mathrm{sp}}
=
\mathrm{corr}
\left(
\mathrm{rank}(R^{\mathrm{VLM}}_{\mathrm{image}}),
\mathrm{rank}(R^{\mathrm{human}}_{\mathrm{image}})
\right).
\label{eq:human_vlm_spearman_final}
\end{equation}
For pairwise comparisons, agreement is
\begin{equation}
\mathrm{PairAgr}
=
\frac{1}{|\mathcal{P}|}
\sum_{(a,b)\in\mathcal{P}}
\mathbb{I}\!\left[
\mathrm{sign}(s^{\mathrm{VLM}}_a-s^{\mathrm{VLM}}_b)
=
\mathrm{sign}(s^{\mathrm{human}}_a-s^{\mathrm{human}}_b)
\right].
\label{eq:human_vlm_pairagr_final}
\end{equation}
Figure~\ref{fig:human_vlm_recoverability_correlation} visualizes the relationship between human image-only recoverability and VLM image-only recoverability.

\subsection{Inter-Rater Agreement for Perception Ratings}
\label{app:human_inter_rater_final}

The perception study includes subjective ordinal ratings and pairwise preferences. We report their agreement only as diagnostics because these ratings are not dataset-construction labels. For categorical QA labels, we use Fleiss' $\kappa$:
\begin{equation}
\kappa
=
\frac{\bar{P}-\bar{P}_e}{1-\bar{P}_e},
\label{eq:fleiss_kappa_final}
\end{equation}
where $\bar{P}$ is observed agreement and $\bar{P}_e$ is chance agreement. For ordinal preference scores, we report Krippendorff's $\alpha$:
\begin{equation}
\alpha
=
1-\frac{D_o}{D_e},
\label{eq:krippendorff_alpha_final}
\end{equation}
where $D_o$ is observed disagreement and $D_e$ is disagreement expected by chance.

Near-zero ordinal $\alpha$ values are interpreted cautiously. The perception ratings are compressed around positive scores and contain many nearby ratings rather than wide disagreement. In such restricted ordinal ranges, chance-corrected agreement can be near zero even when aggregate preference trends are stable. Therefore, ordinal $\alpha$ is not used as evidence for \textsc{KathaBench-25K} dataset-label reliability. Dataset-label reliability is reported separately using trained-annotator categorical agreement and adjudication.

%%%%%%%%%%%%%%%%%%%%%%%%%%%

\section{Additional Baseline and Control Results}
\label{app:additional_baseline_controls}

This section reports supplementary \textsc{KathaBench-25K} baseline and control results. The goal is to test whether KathaTrace measures transition-level recoverability rather than only visual quality, object overlap, generic QA faithfulness, or sequence length. All reported STG values use the same evidence-isolated judge protocol, frozen accepted-answer sets, ambiguity filtering, and valid-subset accounting defined in Secs.~\ref{secb}--\ref{sece}.

\subsection{Extended Planner--Generator Leaderboard}
\label{app:extended_leaderboard}

Table~\ref{tab:appendix_full_leaderboard} reports additional \textsc{KathaBench-25K} planner--generator settings beyond the main paper. Closed systems are included only as reference baselines because their internal prompt rewriting, safety filtering, hardware, and version behavior are not fully controllable. Semantic Compass is reported as a post-generation reranking and single-bridge repair setting applied to Gemma-ST + FLUX candidates; it is not a new standalone generator.

\begin{table*}[t]
\centering
\scriptsize
\setlength{\tabcolsep}{2.6pt}
\renewcommand{\arraystretch}{0.92}
\caption{\textbf{\textsc{KathaBench-25K} extended leaderboard.} Held-out planner--generator results.}
\label{tab:appendix_full_leaderboard}
\resizebox{\textwidth}{!}{%
\begin{tabular}{llccccc}
\toprule
\rowcolor{gray!20}
\textbf{Planner / Setting} &
\textbf{Generator / Repair Layer} &
\textbf{Visual} $\uparrow$ &
\textbf{Text Rec.} $\uparrow$ &
\textbf{Image Rec.} $\uparrow$ &
\textbf{STG} $\downarrow$ &
\textbf{Moral-target Rec.} $\uparrow$ \\
\midrule
Direct & SDXL
& $3.76{\pm}0.06$ & $74.8{\pm}1.0$ & $36.9{\pm}1.6$ & $37.9{\pm}1.8$ & $25.8{\pm}1.7$ \\
\rowcolor{gray!5}
Rule & SDXL
& $3.81{\pm}0.06$ & $75.1{\pm}1.0$ & $39.4{\pm}1.5$ & $35.7{\pm}1.7$ & $28.2{\pm}1.6$ \\
Caption & SDXL
& $3.91{\pm}0.05$ & $75.5{\pm}1.0$ & $42.7{\pm}1.5$ & $32.8{\pm}1.7$ & $31.6{\pm}1.6$ \\
\rowcolor{gray!5}
Scene & SDXL
& $4.02{\pm}0.05$ & $76.2{\pm}0.9$ & $45.3{\pm}1.4$ & $30.9{\pm}1.6$ & $34.1{\pm}1.5$ \\
Gemma-ST & SDXL
& $4.05{\pm}0.05$ & $77.0{\pm}0.9$ & $47.2{\pm}1.4$ & $29.8{\pm}1.5$ & $37.5{\pm}1.5$ \\
\rowcolor{gray!5}
Gemma-ST & StoryDiffusion~\cite{zhou2024storydiffusion}
& $4.07{\pm}0.05$ & $77.2{\pm}0.9$ & $48.0{\pm}1.4$ & $29.2{\pm}1.5$ & $39.1{\pm}1.5$ \\
Gemma-ST & ConSistory
& $4.11{\pm}0.05$ & $77.5{\pm}0.8$ & $48.6{\pm}1.3$ & $28.9{\pm}1.5$ & $40.4{\pm}1.4$ \\
\rowcolor{gray!5}
Gemma-ST & FLUX
& $4.24{\pm}0.05$ & $77.8{\pm}0.8$ & $49.4{\pm}1.3$ & $28.4{\pm}1.4$ & $42.9{\pm}1.4$ \\
\midrule
Closed reference & GPT-4o + GPT-image-1
& $4.35{\pm}0.04$ & $77.6{\pm}0.8$ & $51.4{\pm}1.3$ & $26.2{\pm}1.4$ & $41.7{\pm}1.4$ \\
\rowcolor{gray!5}
Closed reference & Gemini image API reference
& $4.29{\pm}0.04$ & $77.4{\pm}0.8$ & $50.8{\pm}1.3$ & $26.6{\pm}1.4$ & $41.1{\pm}1.4$ \\
\midrule
\rowcolor{blue!12}
\textbf{Gemma-ST} & \textbf{FLUX + Semantic Compass}
& $4.18{\pm}0.05$ & $78.1{\pm}0.8$ & $56.7{\pm}1.2$ & $21.4{\pm}1.3$ & $47.8{\pm}1.3$ \\
\bottomrule
\end{tabular}%
}
\vspace{2pt}

\raggedright
\footnotesize
\textit{Visual}: human visual-quality rating.
\textit{Text/Image Rec.}: recoverability under text-only/image-only evidence.
\textit{Moral-target Rec.}: recovery of the benchmark-specified moral-semantic target.
\vspace{-4pt}
\end{table*}

\subsection{Planner and Generator Ablations}
\label{app:planner_generator_ablation}

Table~\ref{tab:appendix_planner_generator_ablation} separates planning effects from rendering effects on \textsc{KathaBench-25K}. With FLUX fixed, transition-aware planning improves image-only recoverability and reduces STG. With Gemma-ST fixed, stronger renderers improve visual quality, but do not eliminate semantic trajectory loss. This supports the claim that STG is not reducible to image quality alone.

\begin{table*}[t]
\centering
\scriptsize
\setlength{\tabcolsep}{3.0pt}
\renewcommand{\arraystretch}{0.92}
\caption{\textbf{\textsc{KathaBench-25K} planner--generator ablations.} Planning and rendering effects.}
\label{tab:appendix_planner_generator_ablation}
\resizebox{\textwidth}{!}{%
\begin{tabular}{lccccc|lccccc}
\toprule
\rowcolor{gray!20}
\multicolumn{6}{c|}{\textbf{Planner ablation, FLUX fixed}} &
\multicolumn{6}{c}{\textbf{Generator ablation, Gemma-ST fixed}} \\
\rowcolor{gray!20}
\textbf{Planner} &
\textbf{Img.} $\uparrow$ &
\textbf{STG} $\downarrow$ &
\textbf{Moral} $\uparrow$ &
\textbf{Loc.} $\uparrow$ &
\textbf{Human Align.} $\uparrow$ &
\textbf{Generator} &
\textbf{Vis.} $\uparrow$ &
\textbf{Img.} $\uparrow$ &
\textbf{STG} $\downarrow$ &
\textbf{Moral} $\uparrow$ &
\textbf{Fail} $\downarrow$ \\
\midrule
Direct
& $40.8{\pm}1.5$ & $37.0{\pm}1.7$ & $29.7{\pm}1.6$ & $0.36{\pm}0.03$ & $0.46{\pm}0.04$
& SDXL
& $4.05{\pm}0.05$ & $47.2{\pm}1.4$ & $29.8{\pm}1.5$ & $37.5{\pm}1.5$ & $3.8{\pm}0.4$ \\

\rowcolor{gray!5}
Rule
& $43.1{\pm}1.5$ & $35.0{\pm}1.6$ & $31.4{\pm}1.6$ & $0.42{\pm}0.03$ & $0.51{\pm}0.04$
& StoryDiff.~\cite{zhou2024storydiffusion}
& $4.07{\pm}0.05$ & $48.0{\pm}1.4$ & $29.2{\pm}1.5$ & $39.1{\pm}1.5$ & $3.2{\pm}0.4$ \\

Caption
& $46.7{\pm}1.4$ & $32.1{\pm}1.6$ & $34.8{\pm}1.5$ & $0.49{\pm}0.03$ & $0.58{\pm}0.04$
& ConSistory
& $4.11{\pm}0.05$ & $48.6{\pm}1.3$ & $28.9{\pm}1.5$ & $40.4{\pm}1.4$ & $2.5{\pm}0.3$ \\

\rowcolor{gray!5}
Scene
& $49.5{\pm}1.4$ & $30.2{\pm}1.5$ & $38.6{\pm}1.5$ & $0.57{\pm}0.03$ & $0.64{\pm}0.04$
& FLUX
& $4.24{\pm}0.05$ & $49.4{\pm}1.3$ & $28.4{\pm}1.4$ & $42.9{\pm}1.4$ & $1.7{\pm}0.3$ \\

Gemma-ST
& $49.4{\pm}1.3$ & $28.4{\pm}1.4$ & $42.9{\pm}1.4$ & $0.66{\pm}0.03$ & $0.72{\pm}0.03$
& GPT-4o + GPT-image-1
& $4.35{\pm}0.04$ & $51.4{\pm}1.3$ & $26.2{\pm}1.4$ & $41.7{\pm}1.4$ & $1.2{\pm}0.2$ \\

\rowcolor{gray!5}
Oracle
& $61.4{\pm}1.2$ & $16.8{\pm}1.2$ & $52.6{\pm}1.3$ & $0.78{\pm}0.02$ & $0.81{\pm}0.03$
& Gemini image API reference
& $4.29{\pm}0.04$ & $50.8{\pm}1.3$ & $26.6{\pm}1.4$ & $41.1{\pm}1.4$ & $1.5{\pm}0.3$ \\
\bottomrule
\end{tabular}%
}
\vspace{2pt}

\raggedright
\footnotesize
\textit{Img.}: image-only recoverability.
\textit{Loc.}: dimension-level localization.
\textit{Human Align.}: agreement or rank correlation with human judgments.
\textit{Fail}: invalid generation rate.
\vspace{-4pt}
\end{table*}

\subsection{Evaluator-Baseline Analysis}
\label{app:evaluator_baseline_analysis}

Table~\ref{tab:app_evaluator_baselines} tests whether KathaTrace on \textsc{KathaBench-25K} reduces to generic VQA, object-overlap scoring, or scene-level faithfulness evaluation. Object-centric and generic QA baselines capture visible entities and broad plausibility, but provide weaker transition localization and lower human alignment. Full KathaTrace performs best because it combines transition annotations, dimension-specific questions, frozen accepted answers, contrastive variants, and ambiguity filtering.

\begin{table*}[t]
\centering
\scriptsize
\setlength{\tabcolsep}{3.2pt}
\renewcommand{\arraystretch}{0.92}
\caption{\textbf{\textsc{KathaBench-25K} evaluator controls.} Diagnostic value of scoring protocols.}
\label{tab:app_evaluator_baselines}
\resizebox{\textwidth}{!}{%
\begin{tabular}{l p{0.44\linewidth} c c c}
\toprule
\rowcolor{gray!20}
\textbf{Evaluator baseline} &
\textbf{Question construction} &
\textbf{Transition-specific} &
\textbf{Dim. Loc.} $\uparrow$ &
\textbf{Human Corr.} $\uparrow$ \\
\midrule
Object-overlap QA &
Visible entities, objects, and coarse actions without transition targets. &
No & 0.27 & 0.39 \\
\rowcolor{gray!5}
Generic story QA &
Story-level questions generated without transition labels. &
Partial & 0.31 & 0.44 \\
Multi-frame TIFA-style QA &
Scene-level faithfulness questions adapted to multiple frames. &
Partial & 0.44 & 0.52 \\
\rowcolor{gray!5}
Multi-frame DSG-style QA &
Atomic graph-like statements evaluated for frame-level support. &
Partial & 0.50 & 0.58 \\
KathaTrace w/o contrastives &
Transition questions without semantic contrastive variants. &
Yes & 0.51 & 0.56 \\
\rowcolor{gray!5}
KathaTrace w/o dimension-specific questions &
Story-level recoverability questions without subtype localization. &
Partial & 0.42 & 0.61 \\
\rowcolor{blue!12}
Full KathaTrace &
Transition annotations, dimension-specific questions, accepted answers, contrastives, and ambiguity filtering. &
Yes & 0.68 & 0.71 \\
\bottomrule
\end{tabular}%
}
\vspace{-4pt}
\end{table*}

\subsection{Length-Controlled Semantic Compass Controls}
\label{app:length_controlled_bridge_baselines}

Bridge-scene repair changes sequence length. Table~\ref{tab:app_length_controlled_bridge} therefore compares Semantic Compass with length-controlled alternatives on \textsc{KathaBench-25K} using Gemma-ST + FLUX. Random and non-semantic extra frames give only small improvements. Candidate reranking improves recoverability without changing sequence length. Full Semantic Compass gives the strongest improvement by combining validation-selected reranking with transition-localized bridge repair.

\begin{table*}[t]
\centering
\scriptsize
\setlength{\tabcolsep}{3.0pt}
\renewcommand{\arraystretch}{0.90}
\caption{\textbf{\textsc{KathaBench-25K} bridge controls.} Sequence-length ablations on Gemma-ST + FLUX.}
\label{tab:app_length_controlled_bridge}
\resizebox{\textwidth}{!}{%
\begin{tabular}{l p{0.43\linewidth} c c c c}
\toprule
\rowcolor{gray!20}
\textbf{Repair setting} &
\textbf{Repair rule} &
\textbf{Image Rec.} $\uparrow$ &
\textbf{STG} $\downarrow$ &
\textbf{Visual} $\uparrow$ &
\textbf{$\Delta$STG} $\downarrow$ \\
\midrule
Gemma-ST + FLUX base &
No reranking or repair. &
49.4 & 28.4 & 4.24 & -- \\
\rowcolor{gray!5}
Random extra frame &
Insert one additional frame at a random transition. &
50.0 & 27.8 & 4.15 & -0.6 \\
Non-semantic extra frame &
Insert a visually coherent frame not conditioned on the weak transition. &
50.5 & 27.3 & 4.19 & -1.1 \\
\rowcolor{gray!5}
Caption-only bridge &
Insert a frame generated from scene captions rather than transition annotations. &
51.5 & 26.3 & 4.20 & -2.1 \\
Candidate rerank only &
Select among seed candidates using the validation-selected trajectory score. &
52.0 & 25.8 & 4.27 & -2.6 \\
\rowcolor{gray!5}
Bridge repair only &
Insert one bridge frame at the weakest transition. &
54.0 & 24.0 & 4.13 & -4.4 \\
\rowcolor{blue!12}
Full Semantic Compass &
Combine candidate reranking with transition-localized bridge repair. &
56.7 & 21.4 & 4.18 & -7.0 \\
\bottomrule
\end{tabular}%
}
\vspace{-4pt}
\end{table*}

\subsection{Latent-Dimension Gap Analysis}
\label{app:subtype_baseline_ranking}

Table~\ref{tab:app_subtype_baseline_ranking} reports text-to-image gaps for the main latent dimensions in \textsc{KathaBench-25K}. Lower values indicate less loss of recoverable meaning from text to generated images. Action visibility and temporal order are excluded from this table because they are local-visibility and ordering-control dimensions. Intention is excluded because it is an annotation and planning field, not a released scored QA dimension.

\begin{table*}[t]
\centering
\scriptsize
\setlength{\tabcolsep}{4.0pt}
\renewcommand{\arraystretch}{0.90}
\caption{\textbf{\textsc{KathaBench-25K} latent gaps.} Causal, emotional, consequence, and moral-target losses.}
\label{tab:app_subtype_baseline_ranking}
\resizebox{\textwidth}{!}{%
\begin{tabular}{llccccc}
\toprule
\rowcolor{gray!20}
\textbf{Planner} &
\textbf{Generator} &
\textbf{Causal} $\downarrow$ &
\textbf{Emotion} $\downarrow$ &
\textbf{Consequence} $\downarrow$ &
\textbf{Moral-target} $\downarrow$ &
\textbf{Avg. Gap} $\downarrow$ \\
\midrule
Direct prompting & SDXL & 33.5 & 39.6 & 45.2 & 49.1 & 41.9 \\
\rowcolor{gray!5}
Caption planner & SDXL & 28.9 & 35.1 & 41.0 & 44.7 & 37.4 \\
Scene planner & SDXL & 25.8 & 31.9 & 38.2 & 41.6 & 34.4 \\
\rowcolor{gray!5}
Gemma-ST planner & SDXL & 23.6 & 29.4 & 35.5 & 38.7 & 31.8 \\
Gemma-ST planner & StoryDiffusion~\cite{zhou2024storydiffusion} & 22.7 & 28.8 & 34.9 & 38.2 & 31.2 \\
\rowcolor{gray!5}
Gemma-ST planner & ConSistory & 21.5 & 27.4 & 33.5 & 36.7 & 29.8 \\
Gemma-ST planner & FLUX & 19.7 & 25.9 & 31.2 & 34.8 & 27.9 \\
\rowcolor{blue!12}
Gemma-ST + Semantic Compass & FLUX & 15.8 & 22.1 & 26.4 & 29.6 & 23.5 \\
\bottomrule
\end{tabular}%
}
\vspace{-4pt}
\end{table*}

Across methods, causal gaps are smaller than consequence and moral-target gaps. This indicates that adjacent causal events are easier to depict than delayed outcomes or abstract story meanings. Semantic Compass reduces all four latent gaps, with the largest absolute gains on consequence and moral-target recoverability.

\subsection{Valid-Subset and Failure Handling}
\label{app:valid_subset_analysis}

Some generators fail on different \textsc{KathaBench-25K} examples. We therefore report explicit validity accounting. Table~\ref{tab:app_valid_subset} uses the test split. \emph{Total} is the number of requested stories, \emph{Valid} is the number with complete usable storyboards, \emph{Invalid rate} counts empty outputs, corrupted images, missing frames, or incomplete storyboards, \emph{Raw STG} is computed before validity filtering, \emph{Filtered STG} is computed on each method-valid subset, and \emph{Common-valid STG} is computed on the intersection of examples valid for all compared methods.

\begin{table*}[t]
\centering
\scriptsize
\setlength{\tabcolsep}{3.2pt}
\renewcommand{\arraystretch}{0.90}
\caption{\textbf{\textsc{KathaBench-25K} valid-subset audit.} Failure handling for test STG.}
\label{tab:app_valid_subset}
\resizebox{\textwidth}{!}{%
\begin{tabular}{llcccccc}
\toprule
\rowcolor{gray!20}
\textbf{Planner} &
\textbf{Generator} &
\textbf{Total} &
\textbf{Valid} &
\textbf{Invalid Rate} $\downarrow$ &
\textbf{Raw STG} $\downarrow$ &
\textbf{Filtered STG} $\downarrow$ &
\textbf{Common-valid STG} $\downarrow$ \\
\midrule
Direct prompting & SDXL & 500 & 480 & 4.0\% & 39.6 & 37.9 & 38.1 \\
\rowcolor{gray!5}
Rule planner & SDXL & 500 & 482 & 3.6\% & 37.2 & 35.7 & 35.9 \\
Caption planner & SDXL & 500 & 485 & 3.0\% & 34.1 & 32.8 & 33.0 \\
\rowcolor{gray!5}
Scene planner & SDXL & 500 & 487 & 2.6\% & 32.3 & 30.9 & 31.1 \\
Gemma-ST planner & SDXL & 500 & 481 & 3.8\% & 31.2 & 29.8 & 30.0 \\
\rowcolor{gray!5}
Gemma-ST planner & StoryDiffusion~\cite{zhou2024storydiffusion} & 500 & 484 & 3.2\% & 30.4 & 29.2 & 29.4 \\
Gemma-ST planner & ConSistory & 500 & 487 & 2.6\% & 29.8 & 28.9 & 29.1 \\
\rowcolor{gray!5}
Gemma-ST planner & FLUX & 500 & 492 & 1.6\% & 30.1 & 28.4 & 28.6 \\
Closed reference & GPT-4o + GPT-image-1 & 500 & 494 & 1.2\% & 27.7 & 26.2 & 26.4 \\
\rowcolor{gray!5}
Closed reference & Gemini image API reference & 500 & 493 & 1.4\% & 28.1 & 26.6 & 26.8 \\
\rowcolor{blue!12}
Gemma-ST + Semantic Compass & FLUX & 500 & 492 & 1.6\% & 22.5 & 21.4 & 21.6 \\
\bottomrule
\end{tabular}%
}
\vspace{-4pt}
\end{table*}

The common-valid column preserves the same qualitative ranking as the method-valid filtered STG column. This indicates that the observed improvements are not explained by invalid-generation filtering. Representative failure cases include missing causal bridge actions, under-specified moral decision points, unrecoverable emotional transitions, and delayed consequences that are visually plausible but semantically incomplete.

\subsection{Scientific Insights Revealed by KathaTrace}
\label{app:scientific_insights}

KathaTrace reveals three findings that are not captured by visual-quality or generic story-QA evaluation alone. First, visual coherence and semantic-transition recoverability are separable: generated storyboards may preserve characters, setting, and local plausibility while losing the causal, emotional, consequence-bearing, or moral-target link that explains why one scene follows another. Second, this gap is not explained only by renderer quality or generic QA. Transition-specific questions, contrastive variants, ambiguity filtering, and dimension-level scoring improve failure localization and human alignment over object-level, scene-level, and generic multi-frame QA baselines. Third, the gap is actionable: localized low-recoverability transitions can guide candidate reranking and bridge-scene repair, reducing STG without treating Semantic Compass as a new generator.

Together, these results identify semantic trajectory collapse as a distinct failure mode in visual narrative generation. The benchmark therefore contributes not only a dataset and metric, but also an empirical diagnostic: current story generators can produce visually plausible sequences while still failing to preserve recoverable transition meaning.

%%%%%%%%%%%%%%%%%%%%%%%%%%%%%%%%%%%%
\section{Additional Ablation Analysis}
\label{seci}

This section reports ablations that isolate the main design choices in KathaTrace on the proposed \textsc{KathaBench-25K} benchmark. Unless stated otherwise, all ablations use the same held-out split, generated storyboards, evidence conditions, judge prompts, ambiguity filtering, valid-subset accounting, and frozen accepted-answer sets. The goal is not to introduce new baselines, but to show which protocol components are necessary for reliable transition-level diagnosis.

\subsection{Component Ablation}
\label{app:component_ablation}

Table~\ref{tab:app_component_ablation} removes one KathaTrace component at a time on \textsc{KathaBench-25K}. \emph{Dimension localization} measures whether the evaluator identifies the failing semantic dimension. \emph{Human correlation} measures alignment with human recoverability judgments. \emph{Ambiguity leakage} is the percentage of defective or underspecified items that would remain in scoring without the corresponding control.

\begin{table*}[t]
\centering
\scriptsize
\setlength{\tabcolsep}{4pt}
\renewcommand{\arraystretch}{0.92}
\caption{\textbf{\textsc{KathaBench-25K} component ablation.} Effect of removing protocol controls.}
\label{tab:app_component_ablation}
\resizebox{\textwidth}{!}{%
\begin{tabular}{l p{0.40\linewidth} c c c}
\toprule
\rowcolor{gray!20}
\textbf{Setting} &
\textbf{Removed or weakened component} &
\textbf{Dim. Loc.} $\uparrow$ &
\textbf{Human Corr.} $\uparrow$ &
\textbf{Ambig. Leakage} $\downarrow$ \\
\midrule
Generic story QA only &
No transition labels, no dimension-specific questions, no contrastives. &
0.31 & 0.44 & 12.8 \\
\rowcolor{gray!5}
w/o transition labels &
Questions are not anchored to adjacent-scene transition annotations. &
0.38 & 0.49 & 10.9 \\
w/o accepted-answer sets &
Free-form answers are scored without curated paraphrase sets. &
0.44 & 0.51 & 9.6 \\
\rowcolor{gray!5}
w/o contrastive variants &
No entity-preserving semantic contrastive controls. &
0.51 & 0.56 & 8.7 \\
w/o dimension-specific questions &
Questions are story-level rather than action, causal, emotional, consequence, temporal, or moral-specific. &
0.42 & 0.61 & 7.9 \\
\rowcolor{gray!5}
w/o text+image filtering &
Defective or underspecified items are not removed before STG computation. &
0.57 & 0.64 & 15.4 \\
\rowcolor{blue!12}
\textbf{Full KathaTrace} &
\textbf{Transition labels, dimension-specific questions, accepted answers, contrastives, and ambiguity filtering.} &
\textbf{0.68} & \textbf{0.71} & \textbf{4.1} \\
\bottomrule
\end{tabular}%
}
\vspace{-4pt}
\end{table*}

The largest failure mode in Table~\ref{tab:app_component_ablation} is removing text+image filtering: ambiguous or defective items enter the score and make STG less interpretable. Full KathaTrace gives the strongest localization and the lowest ambiguity leakage.

\subsection{Evidence-Condition Ablation}
\label{app:evidence_condition_ablation}

KathaTrace uses three evidence conditions on \textsc{KathaBench-25K}. Text-only estimates source-side recoverability, image-only measures visual recoverability, and text+image identifies ambiguous or contradictory items. Table~\ref{tab:app_evidence_condition_ablation} shows that all three conditions are needed for a clean STG interpretation.

\begin{table*}[t]
\centering
\scriptsize
\setlength{\tabcolsep}{4pt}
\renewcommand{\arraystretch}{0.92}
\caption{\textbf{\textsc{KathaBench-25K} evidence ablation.} Contribution of each evidence condition.}
\label{tab:app_evidence_condition_ablation}
\resizebox{\textwidth}{!}{%
\begin{tabular}{l c c c c c c}
\toprule
\rowcolor{gray!20}
\textbf{Setting} &
\textbf{Text} &
\textbf{Image} &
\textbf{Text+Image} &
\textbf{STG Rel.} $\uparrow$ &
\textbf{Ambig. Detect.} $\uparrow$ &
\textbf{Human Corr.} $\uparrow$ \\
\midrule
Image-only only & No & Yes & No & 0.42 & 0.00 & 0.58 \\
\rowcolor{gray!5}
Text + image-only & Yes & Yes & No & 0.61 & 0.00 & 0.63 \\
Image + text+image only & No & Yes & Yes & 0.49 & 0.71 & 0.60 \\
\rowcolor{gray!5}
Text-only + text+image only & Yes & No & Yes & 0.37 & 0.78 & 0.52 \\
\rowcolor{blue!12}
\textbf{Full evidence protocol} & \textbf{Yes} & \textbf{Yes} & \textbf{Yes} & \textbf{0.73} & \textbf{0.84} & \textbf{0.71} \\
\bottomrule
\end{tabular}%
}
\vspace{-4pt}
\end{table*}

Image-only evaluation can rank \textsc{KathaBench-25K} storyboards, but it cannot distinguish generator failure from question ambiguity. Table~\ref{tab:app_evidence_condition_ablation} shows that text+image filtering supplies that control, while text-only scoring is needed to interpret STG as text-to-image semantic loss.

\subsection{Question-Construction Ablation}
\label{app:question_construction_ablation}

Table~\ref{tab:app_question_construction_ablation} compares question-construction strategies on \textsc{KathaBench-25K}. Object-level and scene-level questions are easier but less diagnostic. Transition-derived dimension-specific questions better identify semantic trajectory collapse because they target action visibility, causality, emotional shift, consequence, temporal order, and moral-target recovery.

\begin{table*}[t]
\centering
\scriptsize
\setlength{\tabcolsep}{3.5pt}
\renewcommand{\arraystretch}{0.92}
\caption{\textbf{\textsc{KathaBench-25K} question ablation.} Diagnostic value of QA construction.}
\label{tab:app_question_construction_ablation}
\resizebox{\textwidth}{!}{%
\begin{tabular}{l p{0.47\linewidth} c c}
\toprule
\rowcolor{gray!20}
\textbf{Question type} &
\textbf{Construction rule} &
\textbf{Dim. Loc.} $\uparrow$ &
\textbf{Human Corr.} $\uparrow$ \\
\midrule
Object-level QA &
Ask about visible characters, objects, and settings only. &
0.27 & 0.39 \\
\rowcolor{gray!5}
Scene-level QA &
Ask whether individual frames depict scene content. &
0.35 & 0.46 \\
Generic story QA &
Ask broad story questions without transition fields. &
0.31 & 0.44 \\
\rowcolor{gray!5}
Action-only transition QA &
Ask about visible actions but not latent transition meaning. &
0.47 & 0.54 \\
Causal/consequence QA only &
Ask why events happen and what follows. &
0.58 & 0.63 \\
\rowcolor{gray!5}
Moral-target QA only &
Ask about the benchmark-specified moral or abstract meaning. &
0.49 & 0.59 \\
\rowcolor{blue!12}
\textbf{Full dimension-specific QA} &
\textbf{Ask action-visibility, causal, emotional, consequence, temporal-order, and moral-target questions.} &
\textbf{0.68} & \textbf{0.71} \\
\bottomrule
\end{tabular}%
}
\vspace{-4pt}
\end{table*}

The full dimension-specific construction in Table~\ref{tab:app_question_construction_ablation} gives the highest localization and human correlation. Intention is not included as a scored QA dimension. It remains an annotation and planning field because it is harder to validate against a single canonical answer.

\subsection{Answer-Matching Ablation}
\label{app:answer_matching_ablation}

KathaTrace uses frozen accepted-answer sets for open-ended \textsc{KathaBench-25K} questions. Table~\ref{tab:app_answer_matching_ablation} shows why: exact matching rejects valid paraphrases, while keyword and embedding matching accept vague or object-only answers. Accepted-answer sets provide deterministic scoring while allowing validated paraphrases.

\begin{table*}[t]
\centering
\scriptsize
\setlength{\tabcolsep}{3.5pt}
\renewcommand{\arraystretch}{0.92}
\caption{\textbf{\textsc{KathaBench-25K} answer matching.} Scoring rules for open answers.}
\label{tab:app_answer_matching_ablation}
\resizebox{\textwidth}{!}{%
\begin{tabular}{l p{0.42\linewidth} c c c}
\toprule
\rowcolor{gray!20}
\textbf{Matching rule} &
\textbf{Description} &
\textbf{False Accept} $\downarrow$ &
\textbf{False Reject} $\downarrow$ &
\textbf{Human Corr.} $\uparrow$ \\
\midrule
Exact string match &
Judge answer must exactly equal the canonical answer. &
2.1 & 18.4 & 0.46 \\
\rowcolor{gray!5}
Keyword-only match &
Answer is correct if it contains selected entity or action words. &
12.7 & 8.9 & 0.55 \\
Embedding-only match &
Answer is correct if sentence similarity exceeds a threshold. &
15.8 & 7.4 & 0.58 \\
\rowcolor{gray!5}
LLM semantic match &
A separate language model decides answer equivalence. &
9.6 & 6.8 & 0.66 \\
\rowcolor{blue!12}
\textbf{Accepted-answer set} &
\textbf{Canonical labels plus frozen curated paraphrases.} &
\textbf{5.3} & 7.9 & \textbf{0.71} \\
\bottomrule
\end{tabular}%
}
\vspace{-4pt}
\end{table*}

The accepted-answer-set rule in Table~\ref{tab:app_answer_matching_ablation} is used because it avoids nondeterministic semantic matching while still handling validated paraphrases. This is important for \textsc{KathaBench-25K}, where many correct answers are short paraphrases of the same transition meaning.

\subsection{Judge-Aggregation Ablation}
\label{app:judge_aggregation_ablation}

Table~\ref{tab:app_judge_aggregation_ablation} compares individual judges with ensemble aggregation on \textsc{KathaBench-25K}. The ensemble improves agreement and calibration, so the main tables use ensemble recoverability.

\begin{table*}[t]
\centering
\scriptsize
\setlength{\tabcolsep}{4pt}
\renewcommand{\arraystretch}{0.92}
\caption{\textbf{\textsc{KathaBench-25K} judge aggregation.} Individual judges versus ensembles.}
\label{tab:app_judge_aggregation_ablation}
\resizebox{\textwidth}{!}{%
\begin{tabular}{l c c c c}
\toprule
\rowcolor{gray!20}
\textbf{Judge setting} &
\textbf{Human Acc.} $\uparrow$ &
\textbf{Spearman} $\uparrow$ &
\textbf{Pairwise Agr.} $\uparrow$ &
\textbf{ECE} $\downarrow$ \\
\midrule
Gemini 2.5 Flash only & 71.8 & 0.64 & 0.68 & 0.118 \\
\rowcolor{gray!5}
Claude Haiku 4.5 only & 72.6 & 0.67 & 0.69 & 0.111 \\
Qwen2.5-VL-7B only & 69.2 & 0.59 & 0.64 & 0.137 \\
\rowcolor{gray!5}
Qwen2.5-VL-3B only & 66.7 & 0.53 & 0.61 & 0.154 \\
SmolVLM2-2.2B only & 64.9 & 0.50 & 0.58 & 0.166 \\
\rowcolor{gray!5}
SmolVLM2-500M only & 61.8 & 0.45 & 0.54 & 0.184 \\
Majority vote ensemble & 75.4 & 0.71 & 0.73 & 0.092 \\
\rowcolor{blue!12}
\textbf{Human-calibrated ensemble} & \textbf{77.1} & \textbf{0.74} & \textbf{0.76} & \textbf{0.081} \\
\bottomrule
\end{tabular}%
}
\vspace{-4pt}
\end{table*}

The ensemble in Table~\ref{tab:app_judge_aggregation_ablation} is not treated as ground truth. It is used as a calibrated proxy whose behavior is checked against human judgments.

\subsection{Semantic Compass Scoring-Term Ablation}
\label{app:semantic_compass_scoring_ablation}

Semantic Compass uses three scoring terms: image-only recoverability, transition coverage, and a copy penalty. Table~\ref{tab:app_compass_scoring_ablation} ablates these terms on \textsc{KathaBench-25K} using Gemma-ST + FLUX candidates.

\begin{table*}[t]
\centering
\scriptsize
\setlength{\tabcolsep}{3.2pt}
\renewcommand{\arraystretch}{0.92}
\caption{\textbf{\textsc{KathaBench-25K} Compass scoring.} Reranking-term ablation on Gemma-ST + FLUX.}
\label{tab:app_compass_scoring_ablation}
\resizebox{\textwidth}{!}{%
\begin{tabular}{l c c c c c c}
\toprule
\rowcolor{gray!20}
\textbf{Score} &
\textbf{$R_{\mathrm{image}}$} &
\textbf{$C_{\mathrm{trans}}$} &
\textbf{$P_{\mathrm{copy}}$} &
\textbf{Image Rec.} $\uparrow$ &
\textbf{STG} $\downarrow$ &
\textbf{Copy Rate} $\downarrow$ \\
\midrule
Recoverability only & Yes & No & No & 52.0 & 25.8 & 8.4 \\
\rowcolor{gray!5}
Transition coverage only & No & Yes & No & 51.2 & 26.6 & 7.8 \\
Copy penalty only & No & No & Yes & 49.8 & 28.0 & 3.9 \\
\rowcolor{gray!5}
Recoverability + transition coverage & Yes & Yes & No & 53.4 & 24.4 & 6.7 \\
Recoverability + copy penalty & Yes & No & Yes & 52.8 & 25.0 & 4.6 \\
\rowcolor{gray!5}
Transition coverage + copy penalty & No & Yes & Yes & 52.4 & 25.4 & 4.8 \\
\rowcolor{blue!12}
\textbf{Full score} & \textbf{Yes} & \textbf{Yes} & \textbf{Yes} & \textbf{56.7} & \textbf{21.4} & \textbf{3.5} \\
\bottomrule
\end{tabular}%
}
\vspace{-4pt}
\end{table*}

Table~\ref{tab:app_compass_scoring_ablation} shows that recoverability and transition coverage improve semantic selection, while the copy penalty prevents repeated-frame shortcuts. The full score gives the best overall trade-off.

\subsection{Held-Out Judge and Held-Out Question Ablation}
\label{app:heldout_judge_question_ablation}

Table~\ref{tab:app_heldout_ablation} checks whether Semantic Compass gains persist when final \textsc{KathaBench-25K} evaluation uses held-out judge prompts and held-out questions. Scores are slightly lower under the strictest setting, but the improvement remains.

\begin{table*}[t]
\centering
\scriptsize
\setlength{\tabcolsep}{5pt}
\renewcommand{\arraystretch}{0.92}
\caption{\textbf{\textsc{KathaBench-25K} held-out validation.} Judge-prompt and question ablation.}
\label{tab:app_heldout_ablation}
\resizebox{0.80\textwidth}{!}{%
\begin{tabular}{l c c c}
\toprule
\rowcolor{gray!20}
\textbf{Evaluation protocol} &
\textbf{Image Rec.} $\uparrow$ &
\textbf{STG} $\downarrow$ &
\textbf{Human Corr.} $\uparrow$ \\
\midrule
Same judge prompts, same questions & 57.1 & 20.2 & 0.68 \\
\rowcolor{gray!5}
Same judge prompts, held-out questions & 56.4 & 20.9 & 0.70 \\
Held-out judge prompts, same questions & 56.0 & 21.2 & 0.71 \\
\rowcolor{blue!12}
\textbf{Held-out judge prompts, held-out questions} & \textbf{55.8} & \textbf{21.4} & \textbf{0.73} \\
\bottomrule
\end{tabular}
}
\vspace{-4pt}
\end{table*}

The strictest setting in Table~\ref{tab:app_heldout_ablation} reduces the risk that improvements are caused by prompt-specific or question-specific overfitting.

\subsection{Moral-Target Granularity Ablation}
\label{app:moral_label_ablation}

Table~\ref{tab:app_moral_label_ablation} compares moral-target label granularities for \textsc{KathaBench-25K}. Binary and coarse labels are easier but less diagnostic. Open-ended moral answers are expressive but less reliable. The final 12-label taxonomy provides a practical balance between specificity, chance level, and agreement.

\begin{table*}[t]
\centering
\scriptsize
\setlength{\tabcolsep}{4pt}
\renewcommand{\arraystretch}{0.92}
\caption{\textbf{\textsc{KathaBench-25K} moral granularity.} Label-taxonomy ablation.}
\label{tab:app_moral_label_ablation}
\resizebox{\textwidth}{!}{%
\begin{tabular}{l c c c c c c}
\toprule
\rowcolor{gray!20}
\textbf{Moral-target setting} &
\textbf{\# Labels} &
\textbf{Chance} &
\textbf{Text MTR} $\uparrow$ &
\textbf{Image MTR} $\uparrow$ &
\textbf{Moral Gap} $\downarrow$ &
\textbf{Human Agr.} $\uparrow$ \\
\midrule
Binary moral / non-moral & 2 & 50.0 & 78.4 & 64.6 & 13.8 & 0.69 \\
\rowcolor{gray!5}
Coarse moral categories & 4 & 25.0 & 67.5 & 53.3 & 14.2 & 0.72 \\
Alternative 10-label taxonomy & 10 & 10.0 & 48.2 & 36.1 & 12.1 & 0.55 \\
\rowcolor{blue!12}
\textbf{Final taxonomy} & \textbf{12} & \textbf{8.3} & \textbf{59.1} & \textbf{47.8} & \textbf{11.3} & \textbf{0.68} \\
Open-ended moral answer & open & -- & 44.7 & 33.4 & 11.3 & 0.49 \\
\bottomrule
\end{tabular}%
}
\vspace{-4pt}
\end{table*}

The 12-label setting in Table~\ref{tab:app_moral_label_ablation} is used because it preserves diagnostic specificity while keeping agreement and chance level interpretable. Moral labels are benchmark-specified semantic targets, not objective moral truths.

\subsection{Action-Dimension Ablation}
\label{app:action_dimension_ablation}

Action visibility is included in full \textsc{KathaBench-25K} recoverability, but the main latent-gap analysis focuses on causal, emotional, consequence, and moral-target gaps. Action is more visually local, while the latent dimensions test harder narrative meaning. Table~\ref{tab:app_action_dimension_ablation} reports how the choice of dimension set affects image recoverability, average STG, and localization.

\begin{table*}[t]
\centering
\scriptsize
\setlength{\tabcolsep}{4pt}
\renewcommand{\arraystretch}{0.92}
\caption{\textbf{\textsc{KathaBench-25K} action ablation.} Local action versus latent dimensions.}
\label{tab:app_action_dimension_ablation}
\resizebox{\textwidth}{!}{%
\begin{tabular}{l c c c p{0.35\linewidth}}
\toprule
\rowcolor{gray!20}
\textbf{Dimension set} &
\textbf{Image Rec.} $\uparrow$ &
\textbf{Avg. STG} $\downarrow$ &
\textbf{Dim. Loc.} $\uparrow$ &
\textbf{Interpretation} \\
\midrule
Action visibility only &
68.2 & 12.7 & 0.41 &
Mostly tests visible event recognition. \\
\rowcolor{gray!5}
Latent only: causal, emotional, consequence, moral-target &
54.6 & 23.5 & 0.68 &
Tests harder transition meaning. \\
Action + latent dimensions &
56.9 & 21.7 & 0.63 &
Useful for full recoverability, but less focused than latent-only analysis. \\
\rowcolor{gray!5}
No action, no moral-target &
56.1 & 22.4 & 0.59 &
Removes the easiest local action cue and the hardest abstract target. \\
Moral-target only &
45.1 & 32.1 & 0.52 &
Tests the hardest abstract semantic target. \\
\bottomrule
\end{tabular}%
}
\vspace{-4pt}
\end{table*}

Table~\ref{tab:app_action_dimension_ablation} explains the reporting choice: action visibility remains part of full recoverability, while the main latent-gap table emphasizes causal, emotional, consequence, and moral-target loss.

%%%%%%%%%%%%%%%%%%%%%%%%%%%%%%%%%%%%
\section{Additional Qualitative Analysis}
\label{sec:app_qualitative_analysis}

This section provides qualitative support for the main KathaTrace results on \textsc{KathaBench-25K}. The examples show how visually plausible storyboards can still lose recoverable transition meaning. We use three kinds of appendix evidence: dataset examples, same-story multi-model comparisons, and KathaTrace diagnostic/repair cases. These figures are illustrative and are not used as the sole basis for model ranking.

\subsection{Dataset Examples}
\label{app:dataset_examples}

\begin{figure*}[!t]
\centering
\includegraphics[width=\textwidth]{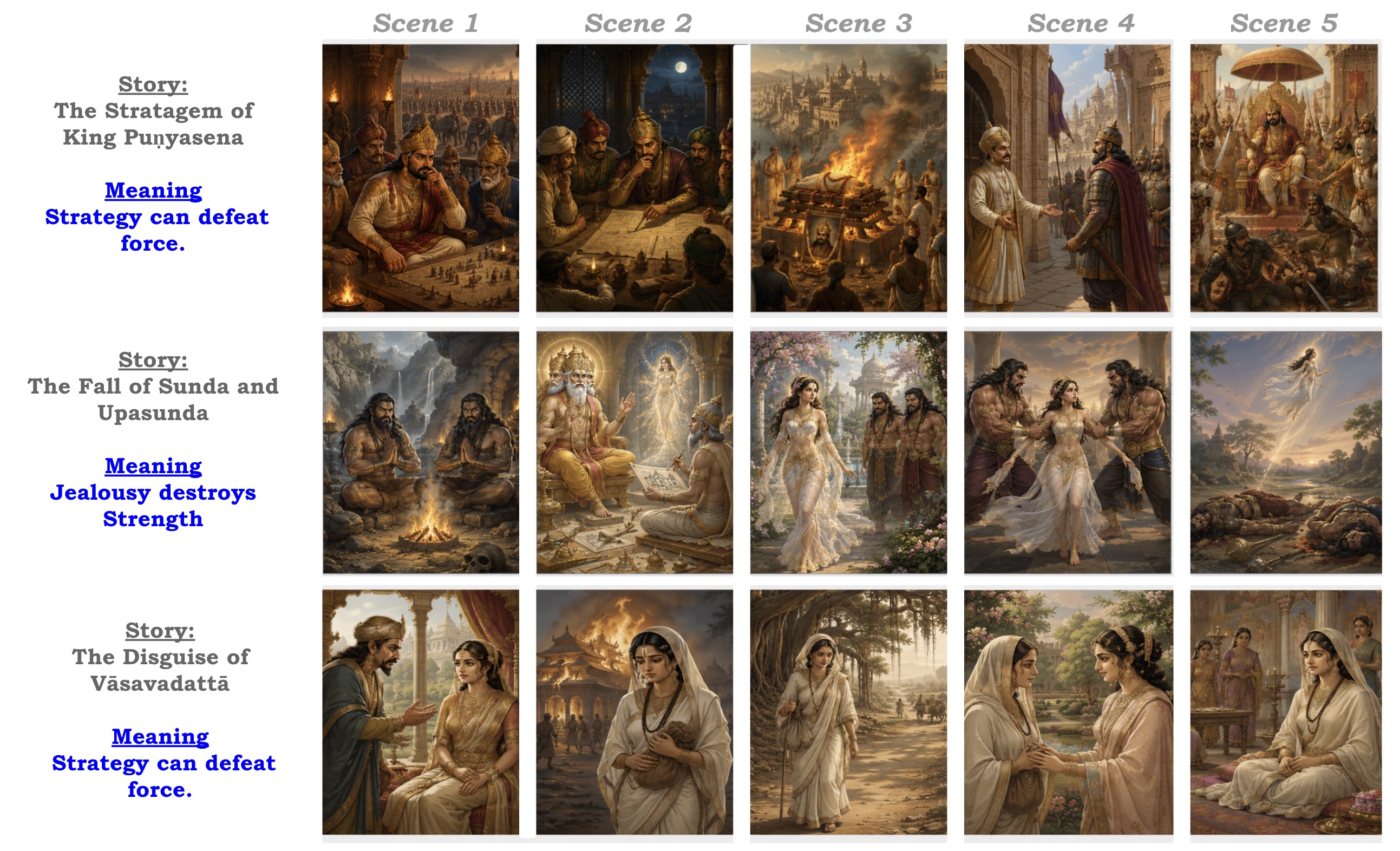}
\caption{\textbf{\textsc{KathaBench-25K} Ocean examples.} Representative stories from \emph{The Ocean of Story}.}
\label{fig:app_dataset_ocean}
\vspace{-4pt}
\end{figure*}

\begin{figure*}[!t]
\centering
\includegraphics[width=\textwidth]{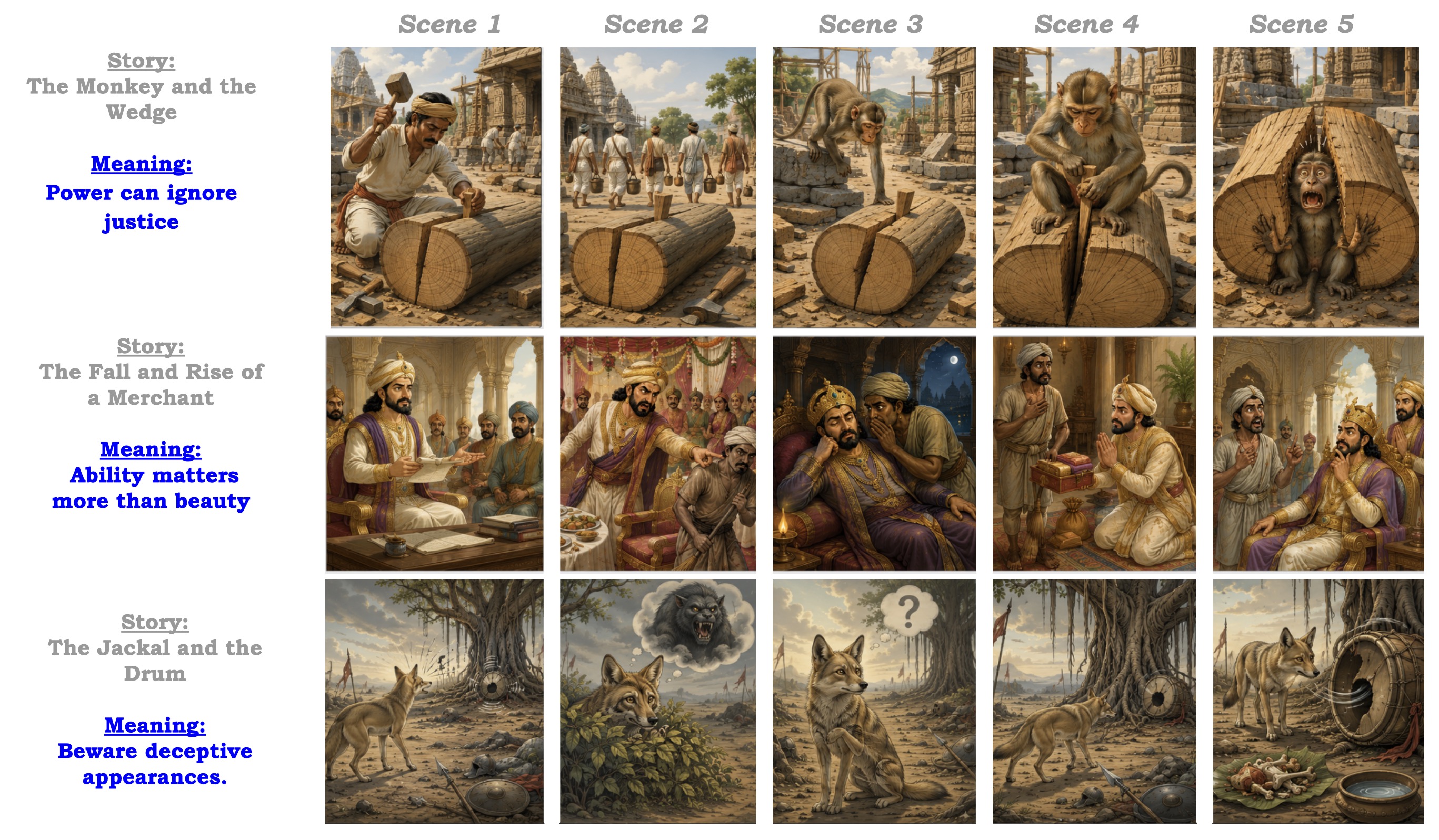}
\caption{\textbf{\textsc{KathaBench-25K} Panchatantra examples.} Representative transition targets from Panchatantra stories.}
\label{fig:app_dataset_panchatantra}
\vspace{-4pt}
\end{figure*}

\begin{figure*}[!t]
\centering
\includegraphics[width=\textwidth]{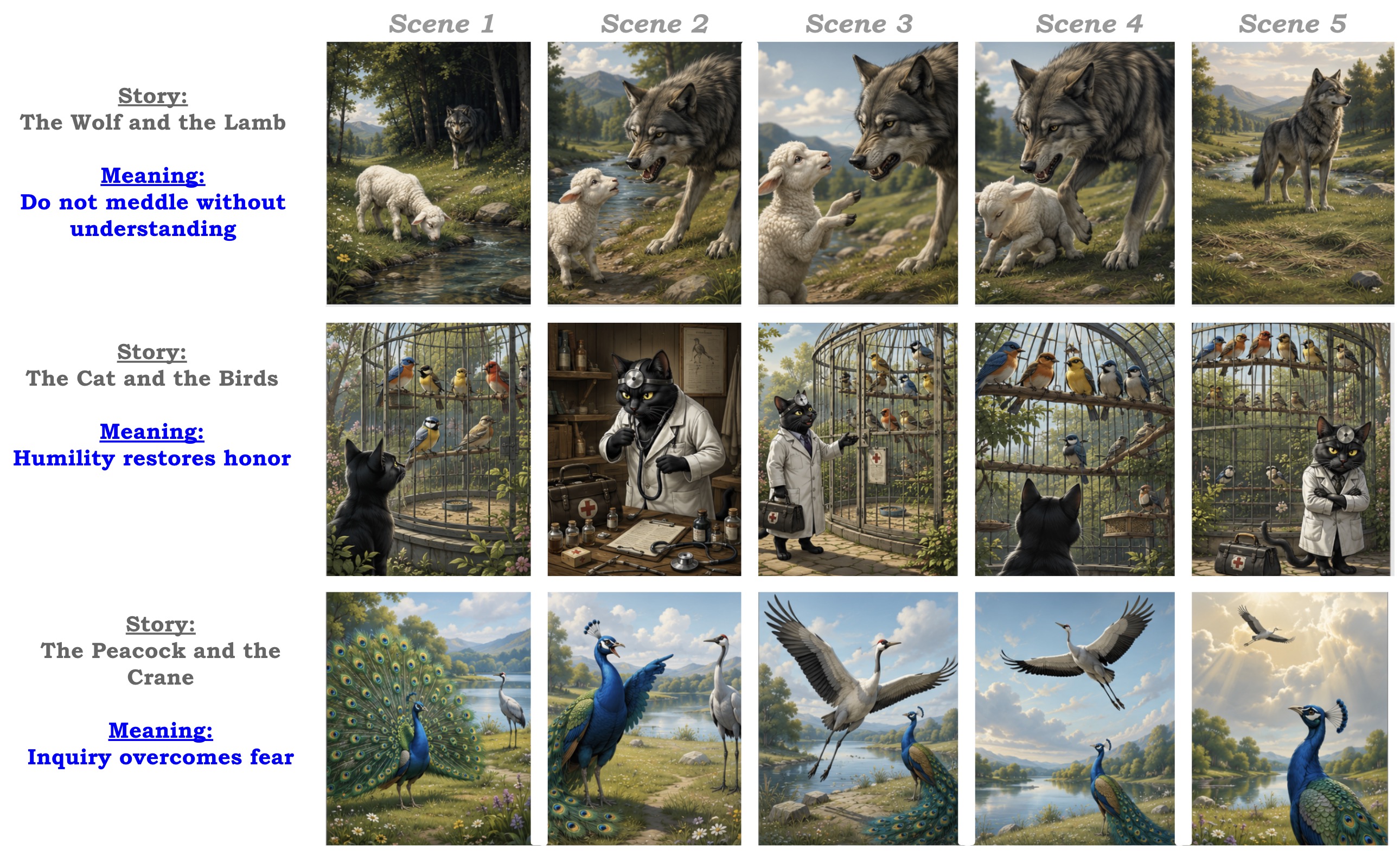}
\caption{\textbf{\textsc{KathaBench-25K} Aesop examples.} Representative transition targets from Aesop's fables.}
\label{fig:app_dataset_aesop}
\vspace{-4pt}
\end{figure*}

Figs.~\ref{fig:app_dataset_ocean}, \ref{fig:app_dataset_panchatantra}, and
\ref{fig:app_dataset_aesop} show representative visual-only \textsc{KathaBench-25K} dataset examples. Each row contains a fixed source story, a short intended meaning, and five target scenes. These examples clarify the kind of transition meaning KathaTrace expects to be recoverable from generated storyboards. Table~\ref{tab:app_dataset_story_coverage} summarizes the story sources and intended meanings used in these qualitative examples.

\begin{table*}[t]
\centering
\scriptsize
\setlength{\tabcolsep}{3.0pt}
\renewcommand{\arraystretch}{0.90}
\caption{\textbf{\textsc{KathaBench-25K} story coverage.} Intended meanings for qualitative examples.}
\label{tab:app_dataset_story_coverage}
\resizebox{\textwidth}{!}{%
\begin{tabular}{lll}
\toprule
\rowcolor{gray!20}
\textbf{Source} & \textbf{Story} & \textbf{Short intended meaning} \\
\midrule
\emph{The Ocean of Story} & The Stratagem of King Puṇyasena & Strategy can defeat force. \\
\emph{The Ocean of Story} & The Fall of Sunda and Upasunda & Jealousy destroys strength. \\
\emph{The Ocean of Story} & The Disguise of Vāsavadattā & Perseverance protects future success. \\
\midrule
Aesop's Fables & The Wolf and the Lamb & Power can ignore justice. \\
Aesop's Fables & The Peacock and the Crane & Ability matters more than beauty. \\
Aesop's Fables & The Cat and the Birds & Beware deceptive appearances. \\
\midrule
Panchatantra & The Monkey and the Wedge & Do not meddle without understanding. \\
Panchatantra & The Jackal and the Drum & Inquiry overcomes fear. \\
Panchatantra & The Fall and Rise of a Merchant & Humility restores honor. \\
\bottomrule
\end{tabular}%
}
\vspace{-4pt}
\end{table*}

\subsection{Qualitative Example Selection}
\label{app:qualitative_selection}

Qualitative examples are selected from the held-out \textsc{KathaBench-25K} split using fixed criteria rather than manual visual preference. We compute STG, dimension-level gaps, validity masks, ambiguity flags, and Semantic Compass repair deltas before selecting examples. Invalid generations and examples marked ambiguous under the text+image condition are not used as evidence of semantic trajectory collapse. Table~\ref{tab:app_qual_selection_protocol} reports the fixed selection groups used for the appendix examples.

\begin{table*}[t]
\centering
\scriptsize
\setlength{\tabcolsep}{3.0pt}
\renewcommand{\arraystretch}{0.90}
\caption{\textbf{\textsc{KathaBench-25K} qualitative selection.} Criteria for appendix examples.}
\label{tab:app_qual_selection_protocol}
\resizebox{\textwidth}{!}{%
\begin{tabular}{lll}
\toprule
\rowcolor{gray!20}
\textbf{Selection group} & \textbf{Criterion} & \textbf{Purpose} \\
\midrule
Dataset examples &
Fixed story source, five-scene plan, and intended meaning &
Shows the benchmark's target narrative structure. \\
High-collapse cases &
Top STG quartile after ambiguity filtering &
Shows visually plausible outputs with weak image-only recoverability. \\
Typical cases &
Near-median STG &
Shows common failures, not only extreme examples. \\
Dimension-specific cases &
Largest valid gap for a transition dimension &
Shows causal, emotional, consequence, or moral-target failures. \\
Repair-success cases &
Largest positive STG reduction after Semantic Compass &
Shows when a localized bridge clarifies missing transition evidence. \\
Repair-boundary cases &
Small or negative repair delta &
Shows limits when the failure is global, ambiguous, or not localizable. \\
\bottomrule
\end{tabular}%
}
\vspace{-4pt}
\end{table*}

\subsection{Same-Story Multi-Model Comparison}
\label{app:same_story_multimodel}

For same-story comparisons, the \textsc{KathaBench-25K} source story, intended meaning, scene plan, number of frames, and recoverability questions are held fixed. Only the generator changes. This controls for story difficulty and makes semantic trajectory failures comparable across models. We use SDXL, FLUX, StoryDiffusion, ConSistory, GPT-4o + GPT-image-1, and Nano Banana as generator families. Closed or API-based systems are reported as references because internal prompt rewriting, safety filtering, model versioning, and decoding details are not fully controllable.

KathaTrace does not score only whether the images look good. It tests whether the intended transition meaning can be recovered from the generated images alone. This separates three failure modes: rendering failure, consistency failure, and semantic trajectory failure. The third case is central to KathaTrace: images may look coherent while omitting the causal, emotional, consequence-bearing, or moral-target transition. Table~\ref{tab:app_same_story_eval_fields} defines the fields used for same-story comparisons, and Table~\ref{tab:app_same_story_model_expected} summarizes the expected qualitative diagnostic patterns.

\begin{table*}[t]
\centering
\scriptsize
\setlength{\tabcolsep}{3.0pt}
\renewcommand{\arraystretch}{0.90}
\caption{\textbf{\textsc{KathaBench-25K} same-story fields.} Diagnostic fields for model comparisons.}
\label{tab:app_same_story_eval_fields}
\resizebox{\textwidth}{!}{%
\begin{tabular}{lll}
\toprule
\rowcolor{gray!20}
\textbf{Field} & \textbf{What it measures} & \textbf{Why it matters} \\
\midrule
Visual quality &
Rendering quality and local visual plausibility &
Checks whether failures are merely low-quality images. \\
Character consistency &
Persistence of main characters across frames &
Separates identity drift from semantic transition loss. \\
Image recoverability &
Recoverability under image-only evidence &
Tests whether the storyboard communicates the transition without source leakage. \\
STG &
$R_{\text{text}} - R_{\text{image}}$ on validity-filtered questions &
Measures text-to-image loss of recoverable transition meaning. \\
Failure type &
Manual or rubric-based diagnostic category &
Identifies missing action, weak causality, emotional drift, consequence loss, or moral-target loss. \\
\bottomrule
\end{tabular}%
}
\vspace{-4pt}
\end{table*}

\begin{table*}[t]
\centering
\scriptsize
\setlength{\tabcolsep}{3.0pt}
\renewcommand{\arraystretch}{0.92}
\caption{\textbf{\textsc{KathaBench-25K} same-story comparison.} Expected qualitative diagnostics by generator.}
\label{tab:app_same_story_model_expected}
\resizebox{\textwidth}{!}{%
\begin{tabular}{llll}
\toprule
\rowcolor{gray!20}
\textbf{Model} &
\textbf{Visual outcome} &
\textbf{Meaning recovered?} &
\textbf{KathaTrace diagnosis} \\
\midrule
SDXL &
Medium--High; scene quality varies &
No / Partial &
Character drift and missing transition evidence. \\
FLUX &
High; strong rendering quality &
Partial &
Good images, but weak causal or moral narrative link. \\
StoryDiffusion &
High; stronger story consistency &
Partial &
Characters are often preserved, but the moral or consequence arc may remain implicit. \\
ConSistory &
High; strong subject consistency &
Partial &
Consistent entities, but weak transition recoverability. \\
GPT-4o + GPT-image-1$^{\ddagger}$ &
High; strong instruction following &
Yes / Partial &
Better scene alignment, but still needs image-only recoverability audit. \\
Nano Banana$^{\ddagger}$ &
High; strong visual instruction following &
Yes / Partial &
Strong scene-level rendering; possible residual semantic drift must be checked. \\
KathaTrace-guided repair &
High; bridge may affect pacing &
Yes &
Localized bridge restores missing transition evidence and improves recoverability. \\
\bottomrule
\end{tabular}%
}
\vspace{-4pt}
\end{table*}

\subsection{KathaTrace Diagnostic and Repair Examples}
\label{app:kathatrace_diagnostic_examples}

\begin{figure*}[!t]
\centering
\includegraphics[width=\textwidth]{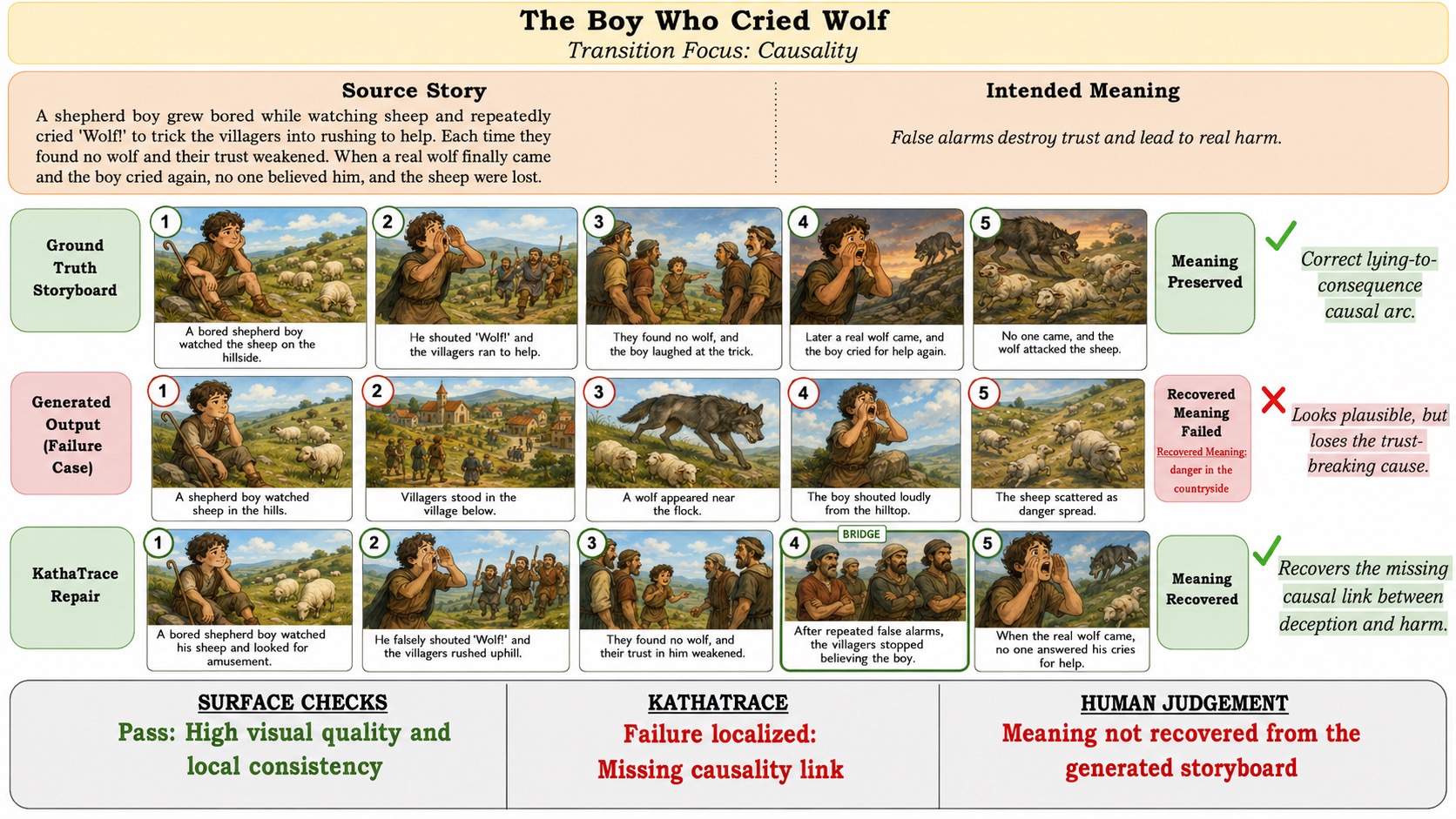}
\caption{\textbf{\textsc{KathaBench-25K} action failure.} Missing wedge-pulling transition in \emph{The Monkey and the Wedge}.}
\label{fig:app_monkey_wedge_action}
\vspace{-4pt}
\end{figure*}

\begin{figure*}[!t]
\centering
\includegraphics[width=\textwidth]{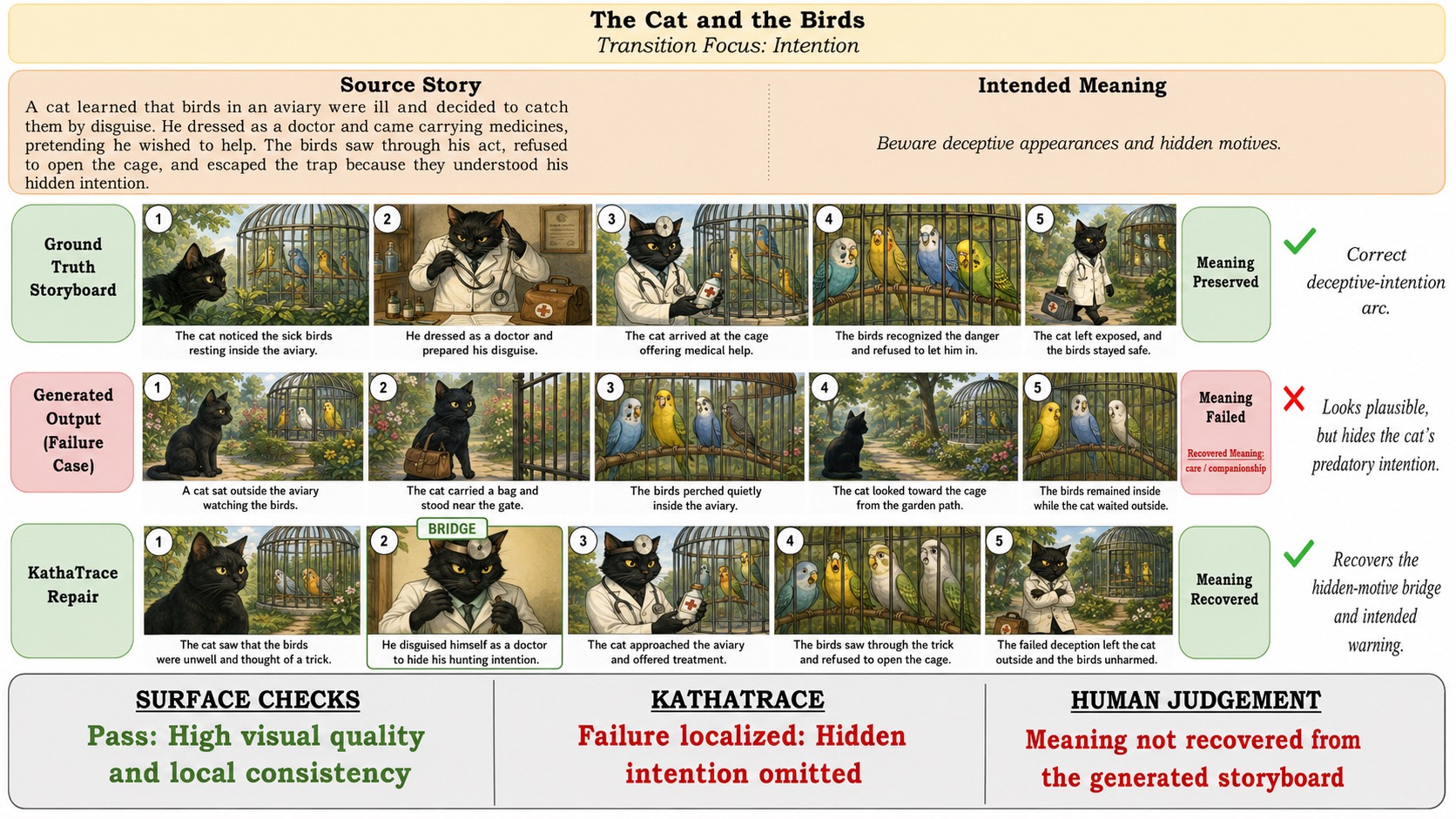}
\caption{\textbf{\textsc{KathaBench-25K} causal failure.} Missing trust-breaking link in \emph{The Boy Who Cried Wolf}.}
\label{fig:app_boy_wolf_causality}
\vspace{-4pt}
\end{figure*}

\begin{figure*}[!t]
\centering
\includegraphics[width=\textwidth]{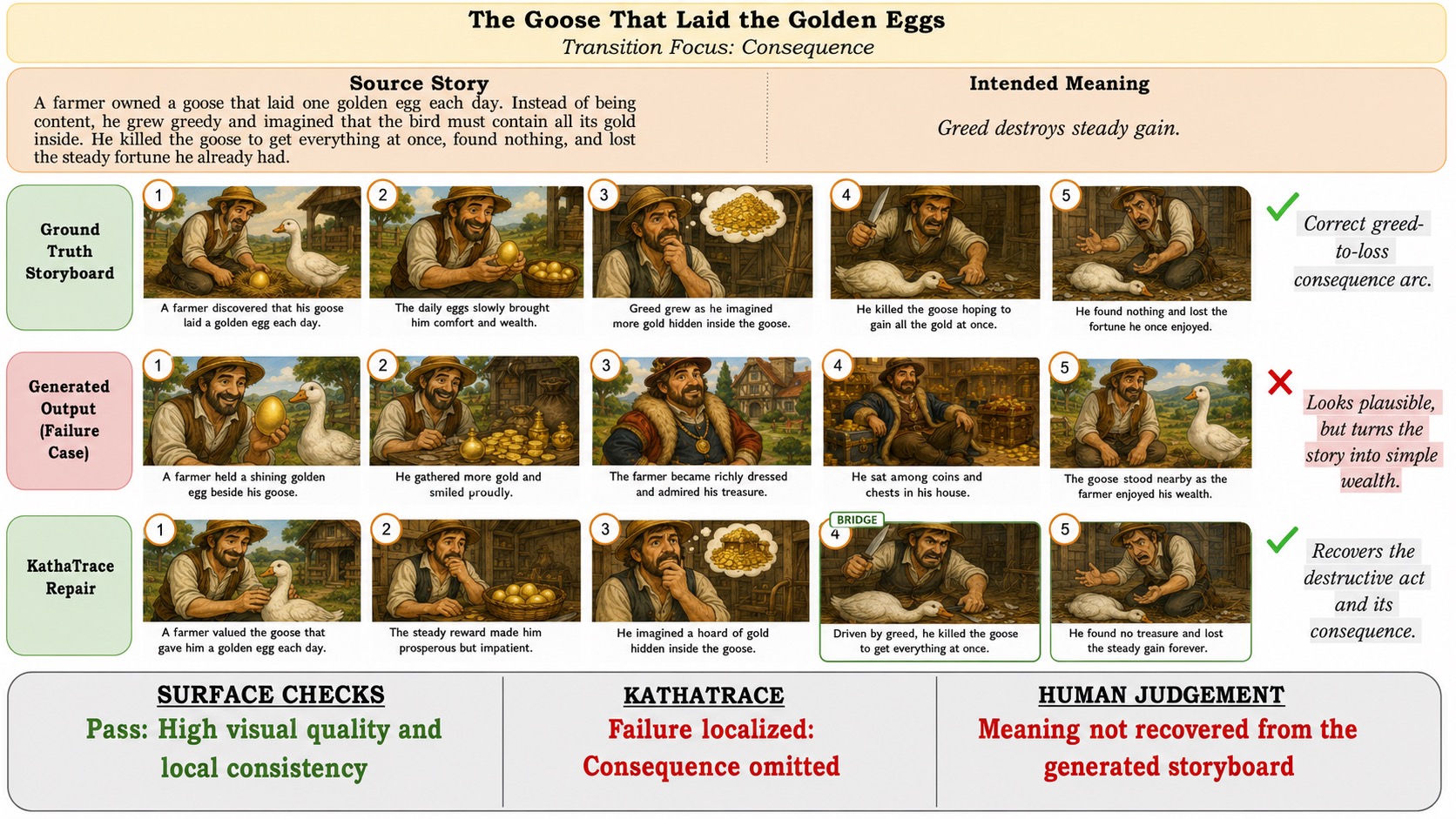}
\caption{\textbf{\textsc{KathaBench-25K} intention-field failure.} Hidden deception in \emph{The Cat and the Birds}.}
\label{fig:app_cat_birds_intention}
\vspace{-4pt}
\end{figure*}

\begin{figure*}[!t]
\centering
\includegraphics[width=\textwidth]{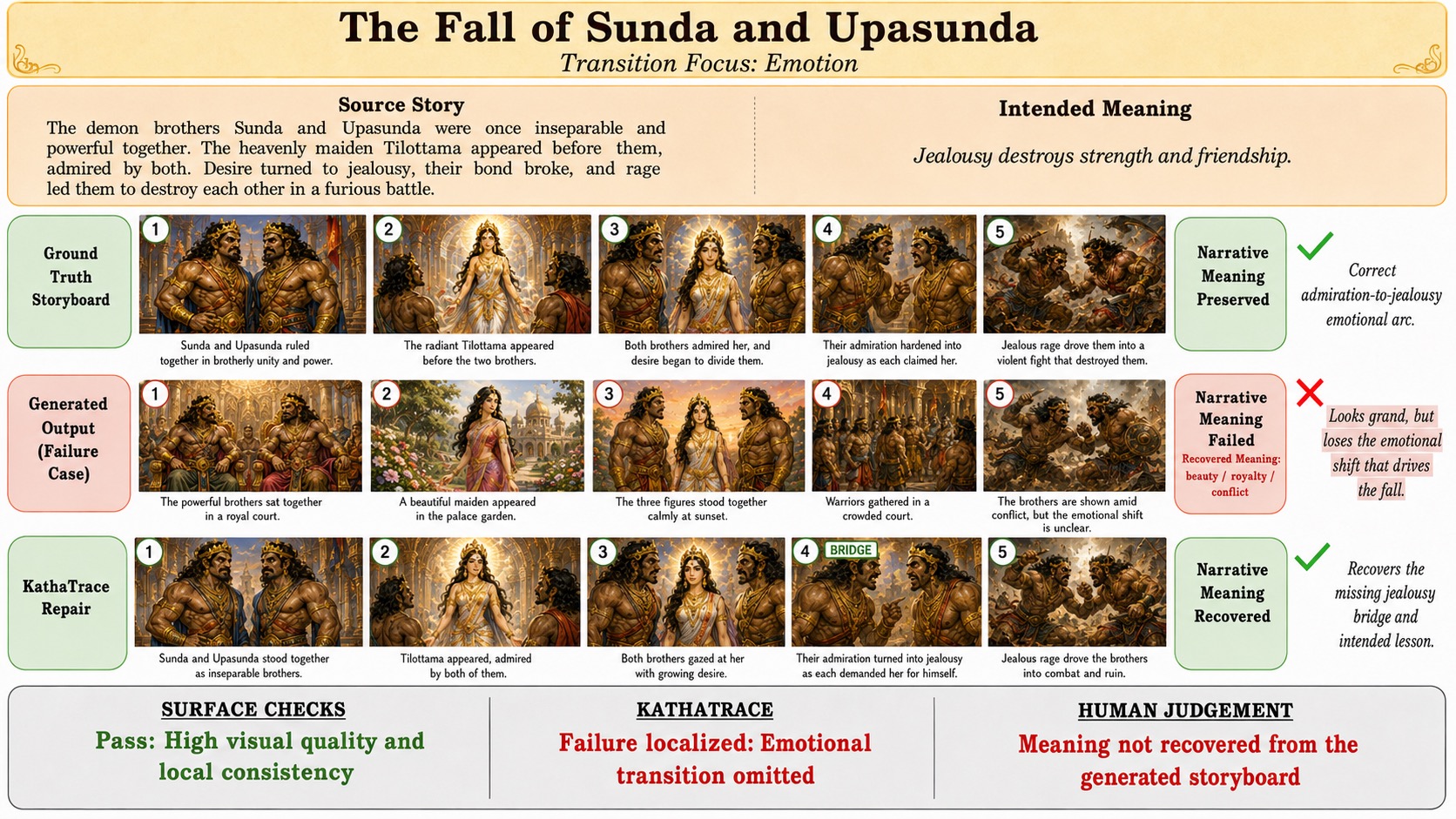}
\caption{\textbf{\textsc{KathaBench-25K} emotional failure.} Missing admiration-to-jealousy turn in \emph{The Fall of Sunda and Upasunda}.}
\label{fig:app_sunda_upasunda_emotion}
\vspace{-4pt}
\end{figure*}

\begin{figure*}[!t]
\centering
\includegraphics[width=\textwidth]{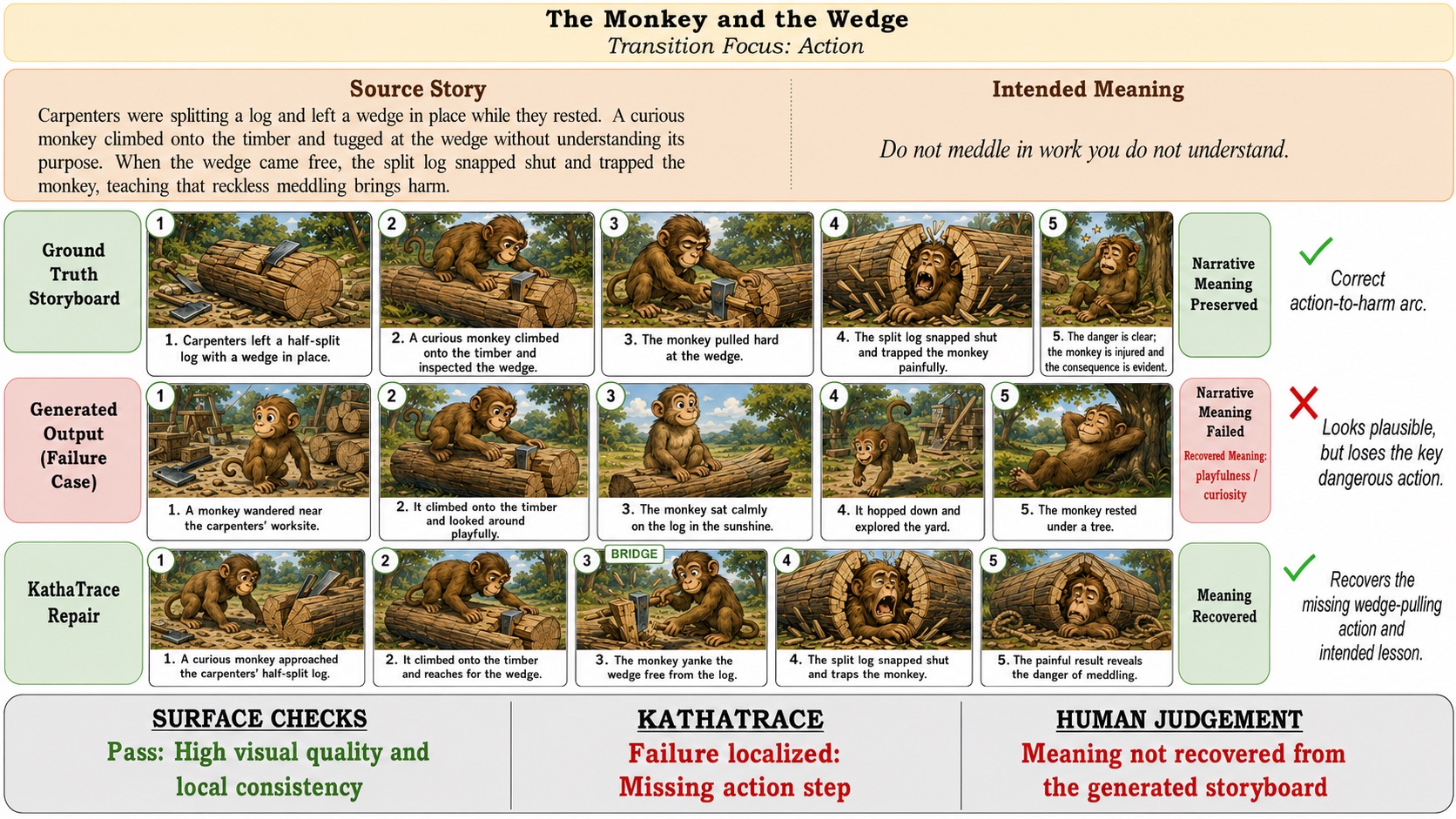}
\caption{\textbf{\textsc{KathaBench-25K} consequence failure.} Missing greed-to-loss transition in \emph{The Goose That Laid the Golden Eggs}.}
\label{fig:app_goose_golden_eggs_consequence}
\vspace{-4pt}
\end{figure*}

Figs.~\ref{fig:app_monkey_wedge_action}--\ref{fig:app_goose_golden_eggs_consequence}
show \textsc{KathaBench-25K} diagnostic examples covering action, causality, source-side intention, emotion, and consequence transitions. Each figure contains the source story, intended meaning, a ground-truth storyboard, a generated failure case, and a KathaTrace diagnosis or repair. The bottom strip compares generic visual/story benchmarks with KathaTrace and human judgment. These examples support the claim that high local coherence can coexist with poor recoverability of transition meaning.

Table~\ref{tab:app_diagnostic_coverage} summarizes the diagnostic coverage of the appendix examples. Intention-like cases are included only as qualitative transition-field examples; they are not part of the released scored QA dimensions.

\begin{table*}[t]
\centering
\scriptsize
\setlength{\tabcolsep}{3.0pt}
\renewcommand{\arraystretch}{0.90}
\caption{\textbf{\textsc{KathaBench-25K} diagnostic coverage.} Qualitative examples by transition focus.}
\label{tab:app_diagnostic_coverage}
\resizebox{\textwidth}{!}{%
\begin{tabular}{llll}
\toprule
\rowcolor{gray!20}
\textbf{Example} & \textbf{Transition focus} & \textbf{Recoverability target} & \textbf{Diagnostic role} \\
\midrule
Main-paper treasure example &
Moral-target &
Recover honesty rather than wealth or luck &
Shows semantic drift despite plausible images. \\
The Watchman and the Cracked Ice &
Causality &
Recover warning as the cause of avoided harm &
Tests warning-to-prevention recoverability. \\
The Beaver and the Broken Dam &
Action visibility &
Recover urgent repair action &
Tests whether a protective action is visually explicit. \\
The Fawn and the Thunder &
Emotion &
Recover fear softening into courage &
Tests emotional transition recoverability. \\
The Heron and the Reed Path &
Intention-like motivation &
Qualitatively recover selfish intent behind guidance &
Shows a hard hidden-motivation case. \\
The Otter and the Cracked Ferry &
Consequence &
Recover ignored warning leading to later loss &
Tests delayed consequence recoverability. \\
\bottomrule
\end{tabular}%
}
\vspace{-4pt}
\end{table*}

\begin{figure*}[!t]
\centering
\begin{minipage}{0.95\textwidth}
    \centering
    \includegraphics[width=\linewidth]{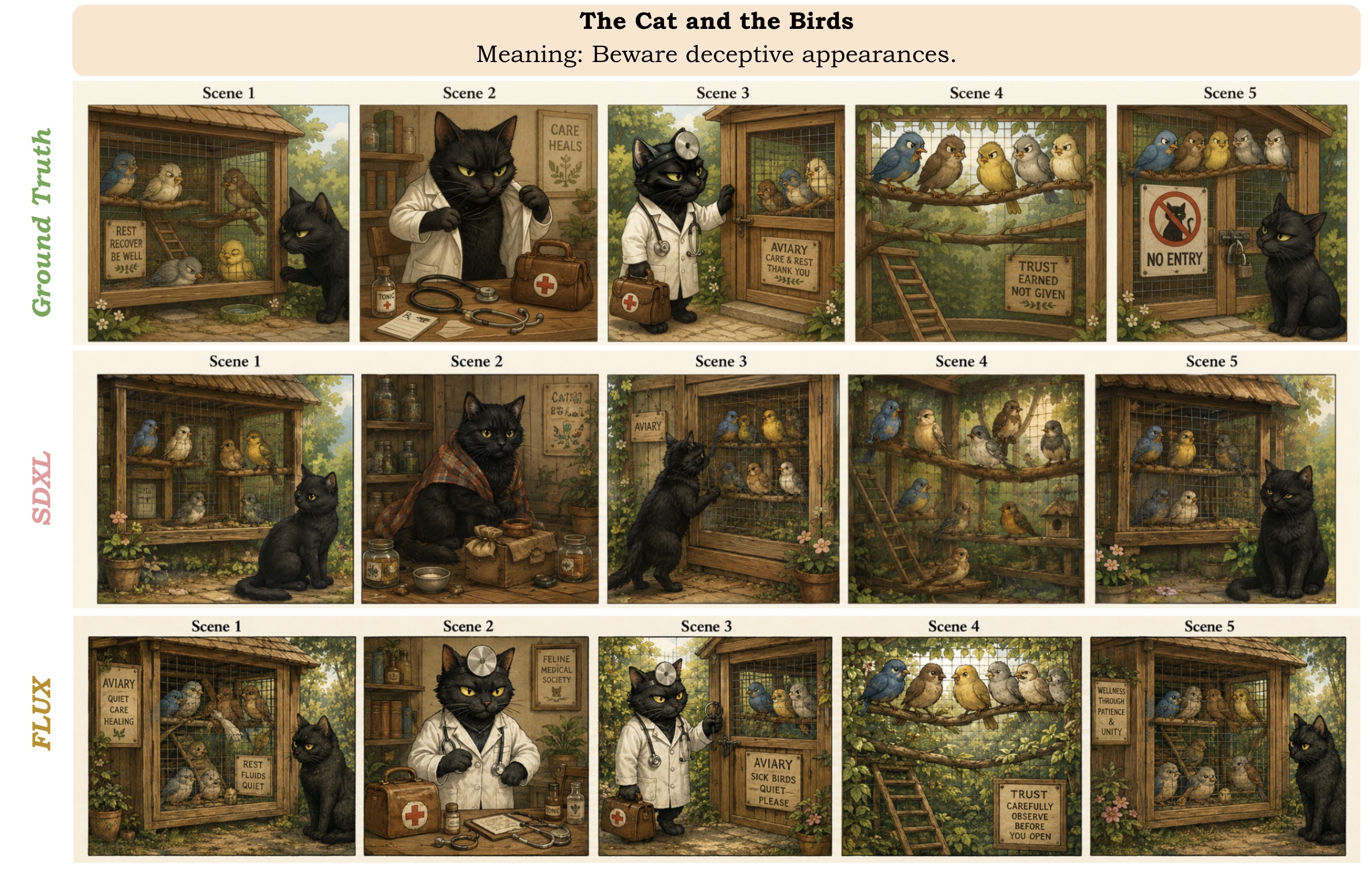}\\[-2pt]
    \includegraphics[width=\linewidth]{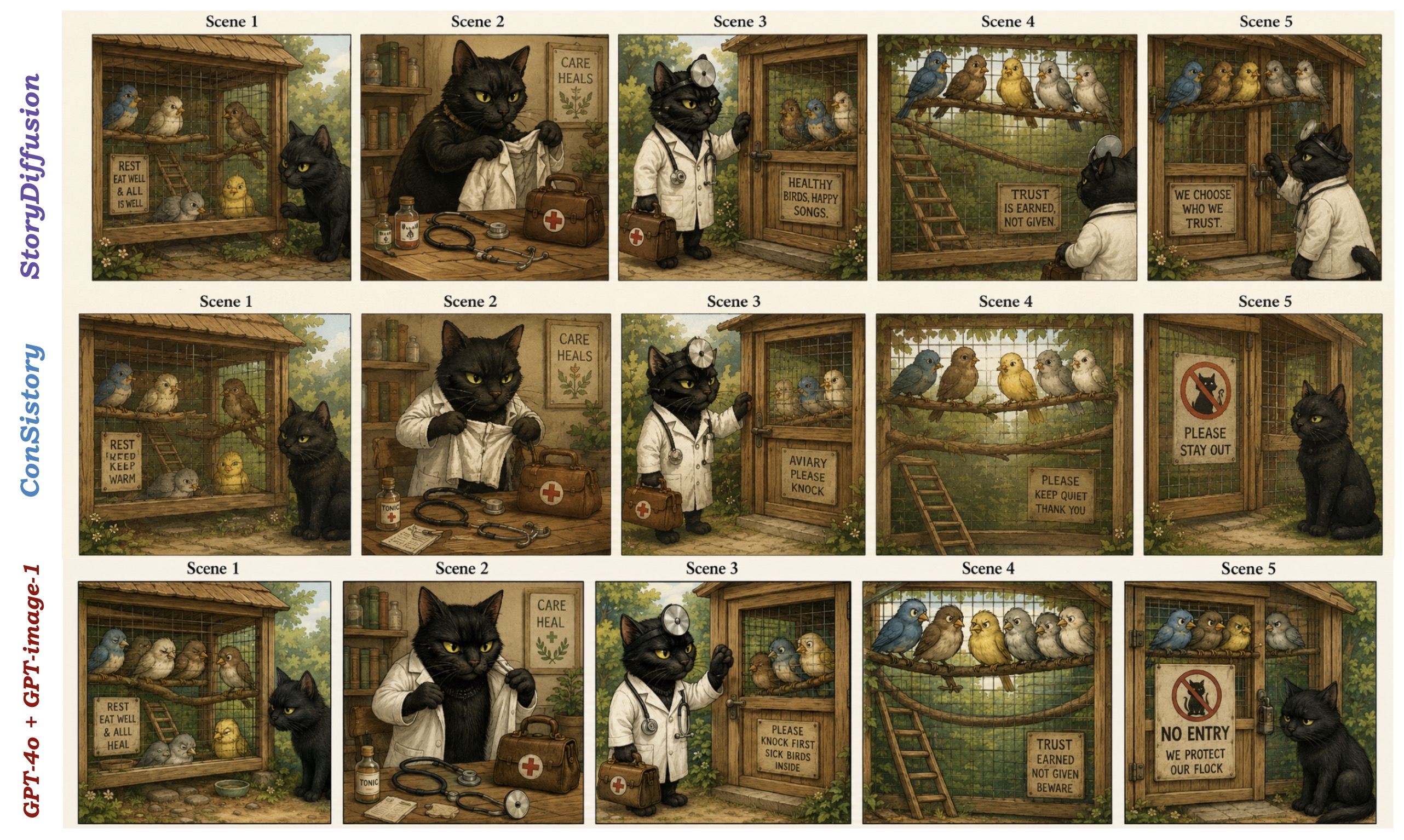}\\[-2pt]
    \includegraphics[width=\linewidth]{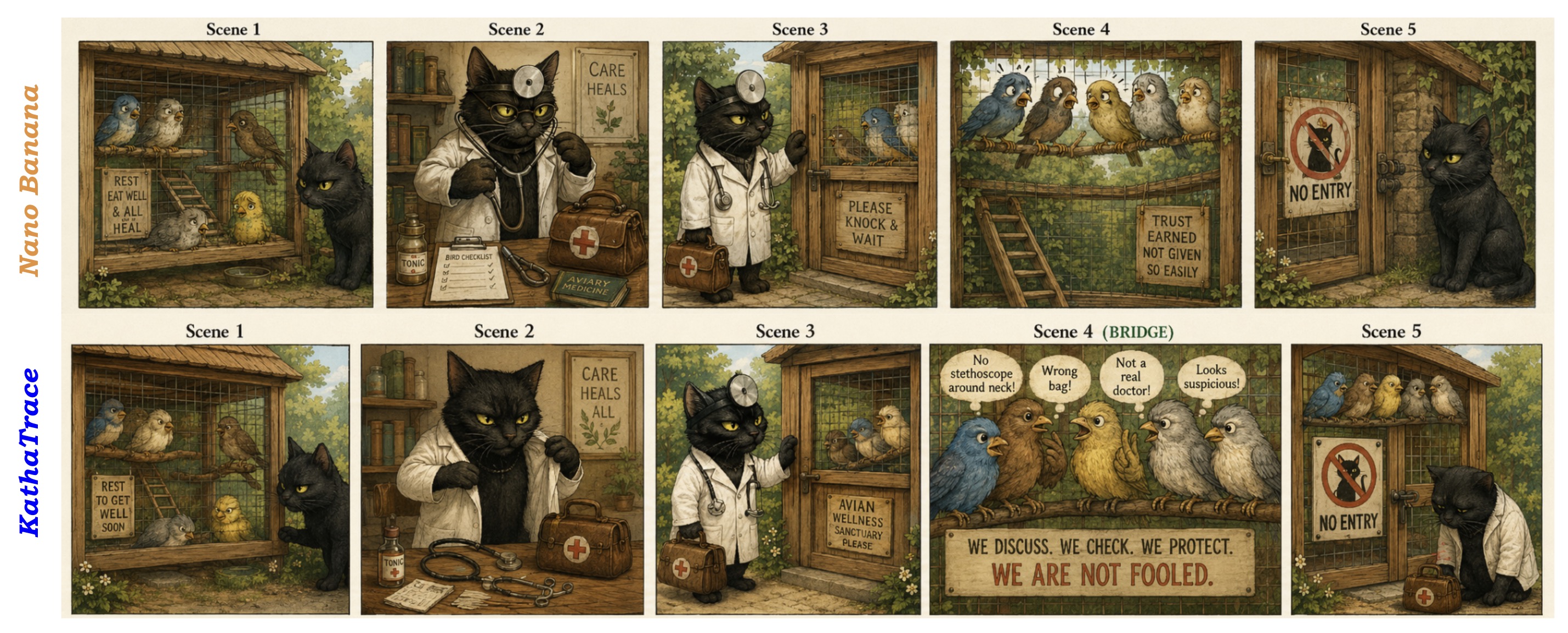}
\end{minipage}
\caption{\textbf{\textsc{KathaBench-25K} same-story comparison.} Generator-agnostic comparison on \emph{The Cat and the Birds}.}
\label{fig:app_cat_birds_generator_agnostic}
\vspace{-4pt}
\end{figure*}

Fig.~\ref{fig:app_cat_birds_generator_agnostic} shows that visually plausible storyboard outputs can differ in how well they preserve the deception-to-refusal transition. The ground-truth storyboard makes the deceptive doctor setup and refusal explicit; weaker generator rows show character drift, implicit deception, or unclear refusal; stronger rows improve scene following but still require image-only recoverability checking. KathaTrace is used here as an evaluator of recoverable narrative meaning, not as a generator.

%%%%%%%%%%%%%%%%%%%%%%%%%%%%%%%%%%%%
\section{Hard Cases for KathaTrace}
\label{seck}

KathaTrace measures whether source-supported transition meaning is recoverable from generated images. It does not assume that every narrative meaning is objective, universal, or visually obvious. Some stories are hard because the key meaning depends on symbolic action, hidden intention, dialogue-like strategy, or delayed reciprocity across non-adjacent scenes.

The examples below define boundary conditions for the protocol. Red dashed boxes mark the visual evidence needed for the intended meaning to remain recoverable. These cases motivate reporting ambiguity rates, strict-gold filtering, and human calibration alongside STG.

\subsection{Symbolic Moral Sacrifice: King Shibi and the Dove}
\label{app:hardcase_shibi}

\begin{figure*}[!t]
\centering
\includegraphics[width=0.92\textwidth]{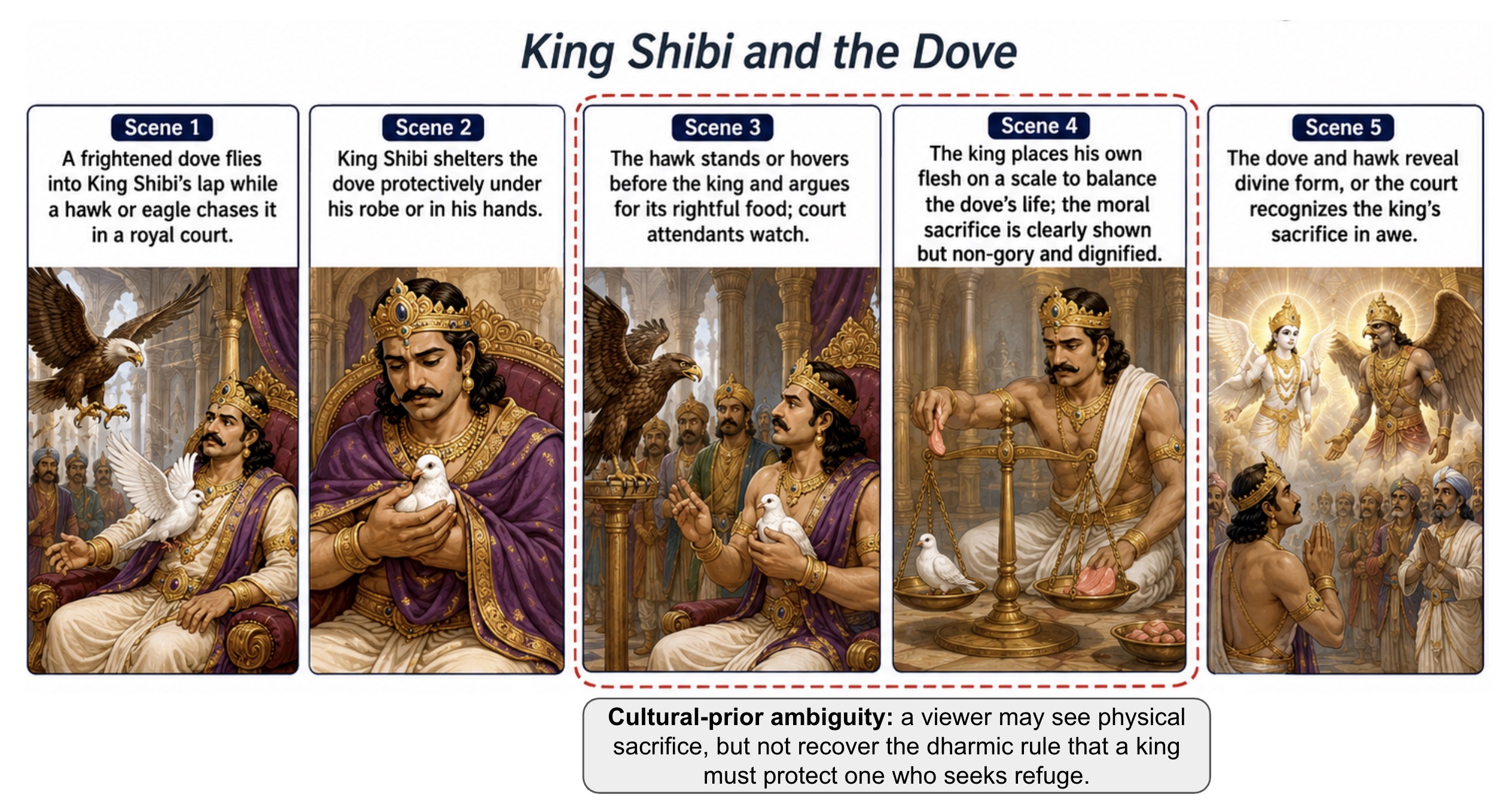}
\caption{\textbf{\textsc{KathaBench-25K} hard case: symbolic sacrifice.} \emph{King Shibi and the Dove}.}
\label{fig:app_hardcase_shibi}
\vspace{-4pt}
\end{figure*}

Fig.~\ref{fig:app_hardcase_shibi} shows a hard case where the central meaning is symbolic rather than merely sequential. Scenes~1--2 can be visually recovered as a rescue: the dove seeks protection and the king shelters it. The harder transition occurs in Scenes~3--4, where the hawk's demand and the king's self-sacrifice establish the moral target. If a storyboard only shows a king, a dove, and a hawk, the meaning may collapse into ``rescue'' or ``kindness'' rather than sacrifice, duty, or justice.

The final divine revelation clarifies the moral frame, but image-only recovery may still require cultural or symbolic knowledge. KathaTrace should therefore treat this example as high-ambiguity unless the storyboard explicitly depicts the weighing scene and the dignity of the sacrifice.

\subsection{Hidden Intention and Verbal Strategy: The Monkey and the Crocodile}
\label{app:hardcase_monkey_crocodile}

\begin{figure*}[!t]
\centering
\includegraphics[width=0.92\textwidth]{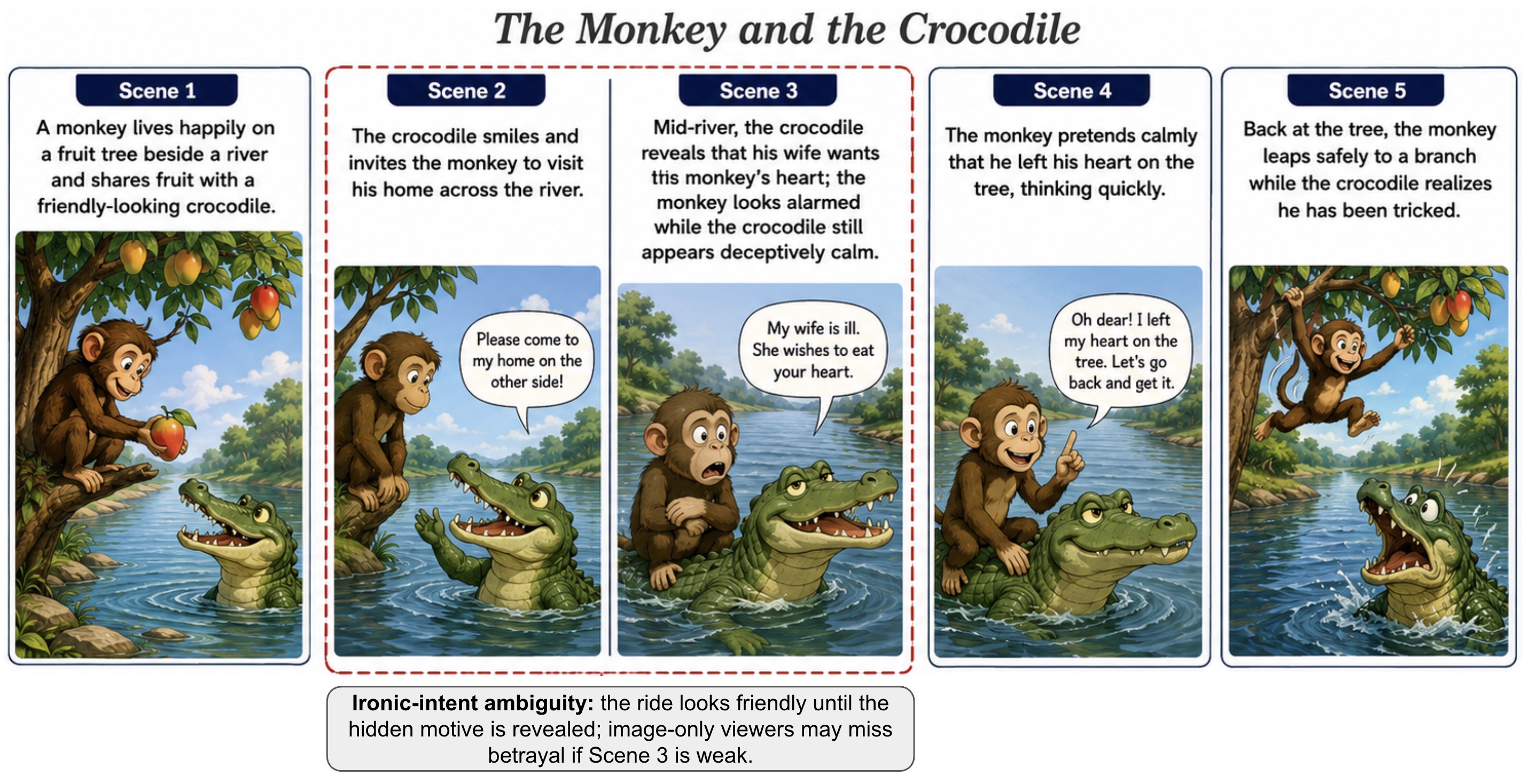}
\caption{\textbf{\textsc{KathaBench-25K} hard case: hidden intention.} \emph{The Monkey and the Crocodile}.}
\label{fig:app_hardcase_monkey_crocodile}
\vspace{-4pt}
\end{figure*}

Fig.~\ref{fig:app_hardcase_monkey_crocodile} shows a hard case where the key transition is mental and dialogic. Scene~1 establishes apparent friendship, but Scenes~2--3 reveal that the crocodile's invitation hides a plan to harm the monkey. A visually coherent storyboard may show the monkey riding the crocodile while failing to communicate deception or danger.

The intended meaning depends on two recoverable facts: the crocodile's friendliness is false, and the monkey survives through quick verbal strategy. Without speech cues, facial cues, or explicit threat evidence, the image-only interpretation may become ``friendship,'' ``travel,'' or ``river adventure.'' This makes the example a boundary case for intention recoverability.

\subsection{Long-Range Reciprocity: The Elephants and the Mice}
\label{app:hardcase_elephants_mice}

\begin{figure*}[!t]
\centering
\includegraphics[width=0.92\textwidth]{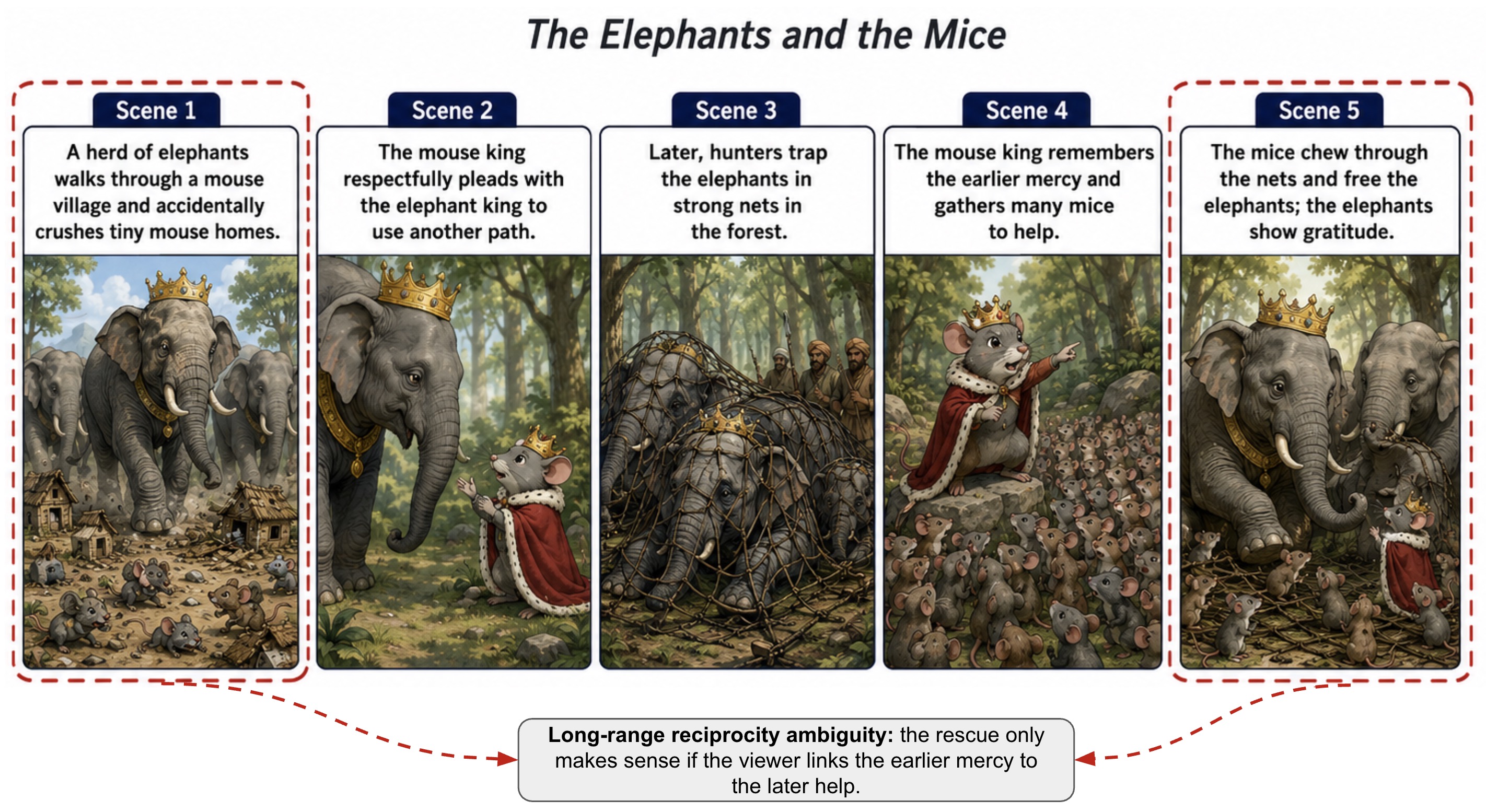}
\caption{\textbf{\textsc{KathaBench-25K} hard case: long-range reciprocity.} \emph{The Elephants and the Mice}.}
\label{fig:app_hardcase_elephants_mice}
\vspace{-4pt}
\end{figure*}

Fig.~\ref{fig:app_hardcase_elephants_mice} shows a hard case where the intended meaning is distributed across the full sequence. Scene~1 introduces harm to the mouse village, Scene~2 shows the mice pleading for mercy, and Scenes~3--5 reverse the power relation when the mice rescue the trapped elephants. If a storyboard preserves only the final rescue, the viewer may recover ``small animals help large animals,'' but not the fuller reciprocity meaning.

This is a boundary case for adjacent-transition scoring. KathaTrace can test whether the early mercy and later rescue are visually present, but the final moral interpretation requires linking distant events. Such examples show why long-range consequence and reciprocity stories should be interpreted with ambiguity rates and human agreement, not STG alone.

%%%%%%%%%%%%%%%%%%%%%%%%%%%%%%%%%%
\subsection{Extensibility Beyond Storyboards}
\label{app:extensibility_beyond_storyboards}

\begin{figure*}[!t]
\centering
\includegraphics[width=\textwidth]{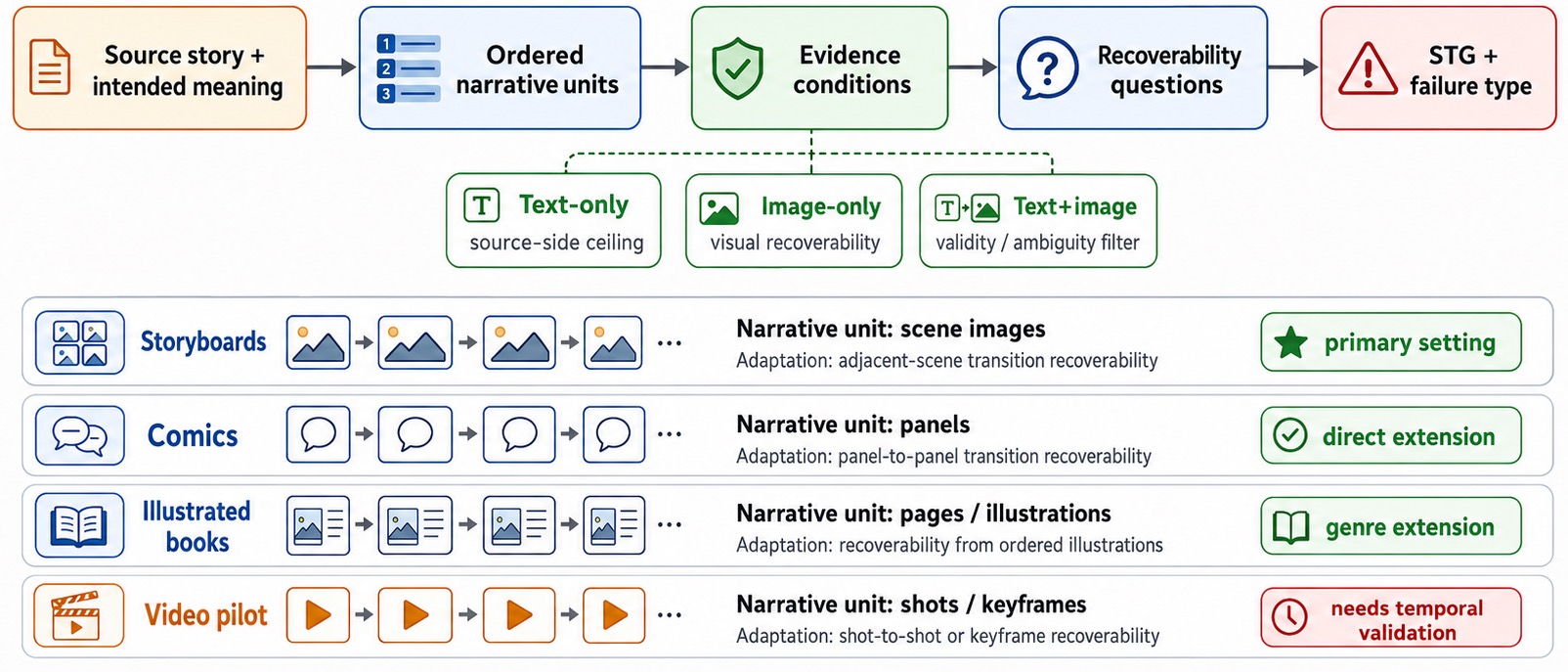}
\caption{\textbf{KathaTrace beyond storyboards.} Ordered narrative units for comics, illustrated pages, and video keyframes.}
\label{fig:app_extensibility_overview}
\vspace{-4pt}
\end{figure*}

KathaTrace is generator-agnostic and format-agnostic. It requires only three inputs: (i) an ordered sequence of narrative units, (ii) source-supported transition targets, and (iii) recoverability questions evaluated under controlled evidence conditions. This makes the protocol directly applicable to comics and illustrated narratives, and partially applicable to video through sampled shots or keyframes. Fig.~\ref{fig:app_extensibility_overview} summarizes these cross-format extensions.

Comics are the closest extension because panels already provide discrete narrative units. A comic page can be evaluated by replacing storyboard scenes with comic panels and applying the same text-only, image-only, and text+image checks to adjacent panel transitions. The core question remains unchanged: whether the transition meaning is recoverable from the visual sequence alone.

\begin{figure*}[!t]
\centering
\includegraphics[width=\textwidth]{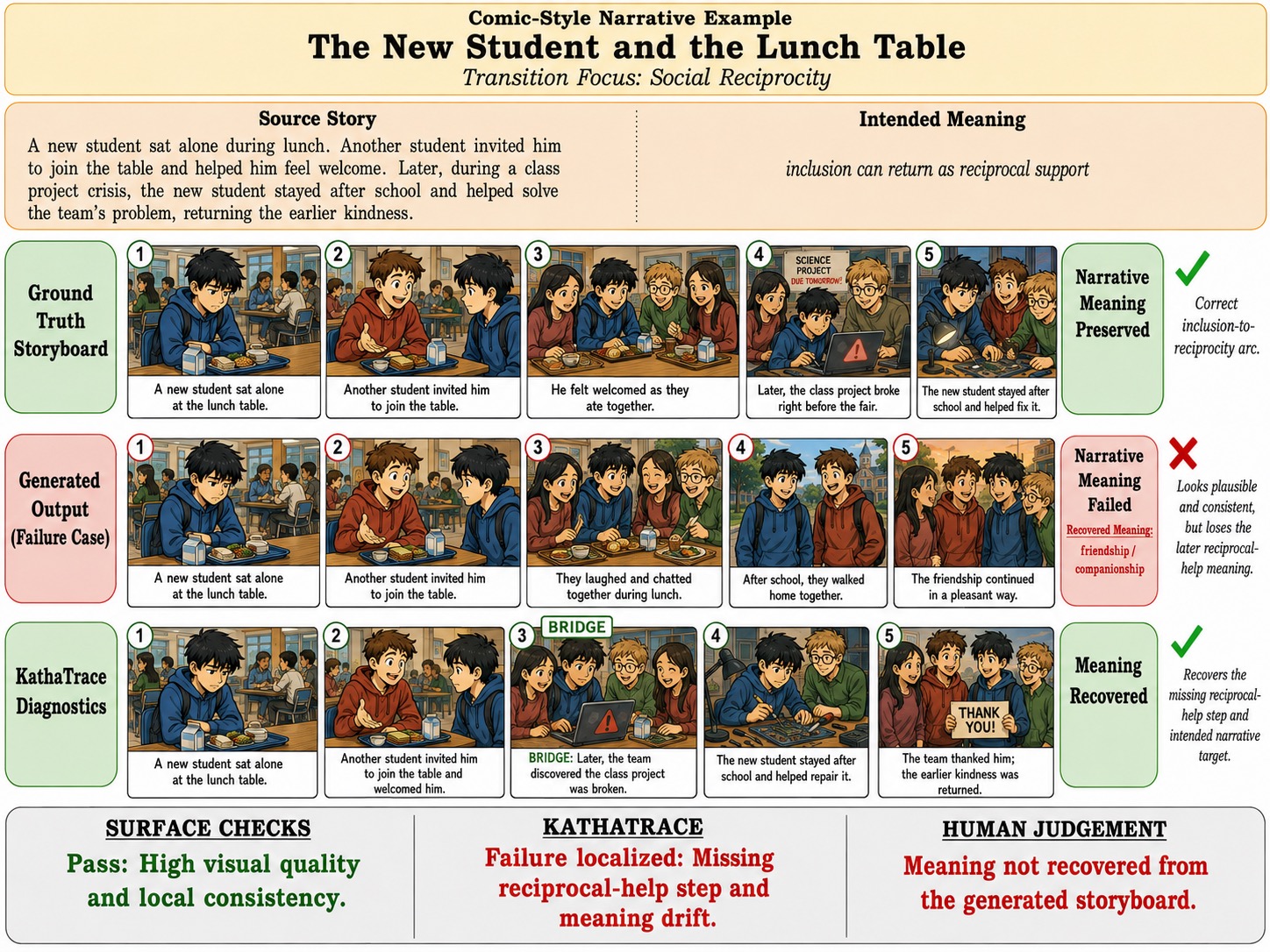}
\caption{\textbf{Comic-style pilot: lunch-table reciprocity.} Cross-format KathaTrace example.}
\label{fig:app_comic_lunch_table}
\vspace{-4pt}
\end{figure*}

\begin{figure*}[!t]
\centering
\includegraphics[width=\textwidth]{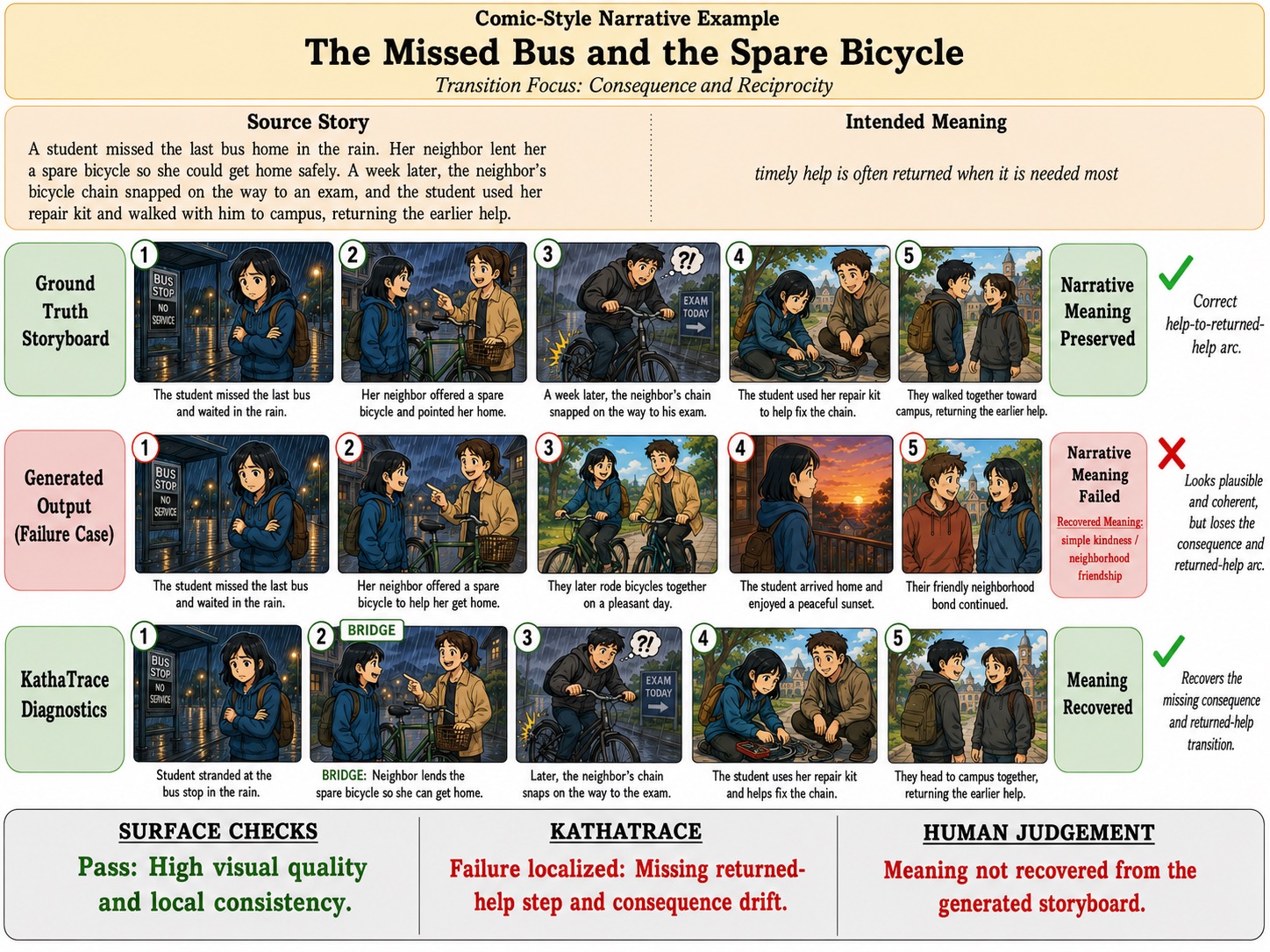}
\caption{\textbf{Comic-style pilot: spare-bicycle reciprocity.} Cross-format KathaTrace example.}
\label{fig:app_comic_spare_bicycle}
\vspace{-4pt}
\end{figure*}

Figs.~\ref{fig:app_comic_lunch_table} and \ref{fig:app_comic_spare_bicycle} show two illustrative comic-style narrative examples. These are cross-format pilot examples rather than released \textsc{KathaBench-25K} benchmark records, and they are included to demonstrate that the same failure-localization and bridge-repair logic transfers naturally to comic panels. In both cases, the generated failure rows remain visually plausible and locally coherent, but omit a meaning-bearing transition; the KathaTrace-guided rows insert or restore the missing bridge step and recover the intended target.

Illustrated books are a similarly natural extension. The protocol can evaluate a sequence of illustrated pages or page-level scene units using the same recoverability conditions, provided that the target transition can be grounded in the source story and tested with fixed accepted answers. The main difference from storyboards is layout granularity: a page may contain one dominant scene or multiple sub-events, so the narrative unit must be defined consistently before evaluation.

Video is a natural but harder extension. A lightweight version can evaluate keyframes or shot-level summaries with the same recoverability protocol. A full video benchmark would additionally require shot segmentation, temporal alignment, motion cues, and careful handling of audio or dialogue. We therefore treat video as future work rather than a fully validated setting in this paper.

More broadly, the protocol can be extended to myths, historical anecdotes, moral tales, procedural stories, school or social narratives, and other ordered visual story forms whenever the intended transition meaning can be grounded in the source and tested with fixed accepted answers. The requirement is not genre-specific; it is whether the transition target remains recoverable from ordered visual evidence.

%%%%%%%%%%%%%%%%%%%%%%%%%%%%%%%%%%
\section{Robustness Analysis}
\label{secm}

This section tests whether the main \textsc{KathaBench-25K} Semantic Trajectory Gap (STG) conclusions remain interpretable under common validity threats. The controls address judge dependence, prompt sensitivity, categorical-label reliability, ordinal-rating instability, moral-target-only effects, source-side leakage, and ambiguity filtering. Held-out Semantic Compass evaluation and length-controlled repair controls are reported in Secs.~\ref{app:heldout_judge_question_ablation} and~\ref{app:length_controlled_bridge_baselines}.

For method $m$ and question $q$, ambiguity is defined using the text+image condition. Let $J(q,(S,\hat{X}_m))$ be the judge answer when both the source story and generated storyboard are available, and let $\mathcal{A}(q)$ be the frozen accepted-answer set. The ambiguity indicator is
\begin{equation}
b_m(q)=
\mathbb{I}\!\left[
\mathrm{match}\!\left(J(q,(S,\hat{X}_m)),\mathcal{A}(q)\right)=0
\right].
\label{eq:robustness_ambiguity_indicator}
\end{equation}
The ambiguity rate is
\begin{equation}
\rho_{\mathrm{amb},m}
=
\frac{1}{|Q|}
\sum_{q\in Q} b_m(q).
\label{eq:robustness_ambiguity_rate}
\end{equation}
Thus, a question is counted as ambiguous or defective for method $m$ when the intended answer is not recoverable even with both text and image evidence. Such questions are excluded from filtered STG so that unclear items are not counted as image-side semantic loss.

Table~\ref{tab:app_stg_validity_controls} summarizes the main robustness controls for STG interpretation.

\begin{table*}[t]
\centering
\scriptsize
\setlength{\tabcolsep}{3.5pt}
\renewcommand{\arraystretch}{0.95}
\caption{\textbf{\textsc{KathaBench-25K} STG robustness.} Validity controls for semantic-gap interpretation.}
\label{tab:app_stg_validity_controls}
\resizebox{\textwidth}{!}{%
\begin{tabular}{p{0.18\linewidth} p{0.28\linewidth} p{0.27\linewidth} p{0.22\linewidth}}
\toprule
\rowcolor{gray!20}
\textbf{Threat} & \textbf{Control} & \textbf{Observed result} & \textbf{Interpretation} \\
\midrule
Single-judge dependence &
Compare individual judges and the ensemble on the same evidence packets. &
The recoverability gap remains across judge families; the calibrated ensemble gives human agreement $77.1\%$ and Spearman $\rho=0.74$. &
STG is not driven by one judge. \\
\rowcolor{gray!5}
Prompt sensitivity &
Compare neutral, strict visual-evidence, and uncertainty-encouraging prompts. &
The final prompt gives human correlation $0.71$ and false acceptance $5.3\%$. &
Prompting reduces unsupported image-only inference. \\
Categorical-label reliability &
Evaluate categorical agreement before adjudication. &
Moral-target Fleiss' $\kappa=0.8453$. &
The categorical labels are reliable enough for benchmark grouping. \\
\rowcolor{gray!5}
Ordinal-rating instability &
Treat 1--5 human ratings as subjective diagnostics only. &
Ordinal Krippendorff $\alpha$ is near zero under compressed score ranges. &
Ordinal ratings are not used as dataset-label reliability evidence. \\
Moral-target-only effect &
Remove moral-target questions from latent-gap computation. &
The non-moral latent set still gives STG $=22.4$. &
The gap is not caused only by moral-target questions. \\
\rowcolor{gray!5}
Source-side leakage &
Scan image-only packets for source stories, prompts, captions, labels, and metadata. &
Detected leakage $=0.0\%$. &
Image-only scoring does not use explicit source-side text or metadata. \\
Human calibration &
Compare VLM decisions with human-gold judgments. &
Agreement $=77.1\%$, Spearman $\rho=0.74$, ECE $=0.081$. &
VLM judges are calibrated proxies, not ground truth. \\
\bottomrule
\end{tabular}%
}
\vspace{-4pt}
\end{table*}

These controls separate dataset-label reliability from subjective perception ratings. Categorical moral-target labels are evaluated with Fleiss' $\kappa$, consensus review, and adjudication. Ordinal 1--5 ratings are used only as human-study diagnostics because their restricted score ranges can produce near-zero chance-corrected agreement even when aggregate trends are stable.

The strongest threat to image-only evaluation is source leakage. If image-only packets contained source stories, prompts, captions, labels, or metadata, image recoverability would be inflated. The leakage scan in Table~\ref{tab:app_stg_validity_controls} finds no detected source-side fields. This does not mean every visual inference is unambiguous; it means the image-only condition does not receive explicit source-side information.

\begin{table*}[t]
\centering
\scriptsize
\setlength{\tabcolsep}{3.2pt}
\renewcommand{\arraystretch}{0.95}
\caption{\textbf{\textsc{KathaBench-25K} evidence robustness.} Effect of ambiguity filtering.}
\label{tab:app_evidence_filtering_robustness}
\resizebox{\textwidth}{!}{%
\begin{tabular}{l c c c c c c c}
\toprule
\rowcolor{gray!20}
\textbf{Setting} &
\textbf{Text} &
\textbf{Image} &
\textbf{Text+Image} &
\textbf{Valid Qs.} $\uparrow$ &
\textbf{Image Rec.} $\uparrow$ &
\textbf{STG} $\downarrow$ &
\textbf{Human Corr.} $\uparrow$ \\
\midrule
Image-only only &
No & Yes & No &
$100.0\%$ & $54.6$ & -- & $0.58$ \\
\rowcolor{gray!5}
Text + image-only, no ambiguity filter &
Yes & Yes & No &
$100.0\%$ & $54.6$ & $26.6$ & $0.63$ \\
Text+image validity filtering &
Yes & Yes & Yes &
$92.7\%$ & $53.4$ & $24.2$ & $0.69$ \\
\rowcolor{blue!12}
Final ambiguity-filtered protocol &
Yes & Yes & Yes &
$89.6\%$ & \best{$54.6$} & \best{$23.5$} & \best{$0.71$} \\
\bottomrule
\end{tabular}%
}
\vspace{2pt}
\raggedright\footnotesize
Recoverability and STG are recomputed on the remaining valid questions after each filtering step, so image recoverability need not change monotonically across rows.
\vspace{-4pt}
\end{table*}

\begin{table}[t]
\centering
\scriptsize
\setlength{\tabcolsep}{3.5pt}
\renewcommand{\arraystretch}{1.05}
\caption{\textbf{\textsc{KathaBench-25K} visual budget.} Interpretation of panel-count controls.}
\label{tab:visual_budget_design}
\resizebox{\linewidth}{!}{
\begin{tabular}{l p{4.4cm} p{4.8cm}}
\toprule
\rowcolor{gray!20}
\textbf{Setting} & \textbf{Visual budget} & \textbf{Diagnostic purpose} \\
\midrule
Canonical storyboard &
Five generated panels aligned to the five released scenes. &
Standard benchmark evaluation. \\
Extra-frame controls &
Six displayed panels after adding one random, non-semantic, or caption-only frame. &
Tests whether gains come merely from increasing sequence length. \\
Bridge repair only &
Six displayed panels after adding one transition-localized bridge panel. &
Tests whether repairing the weakest transition improves recoverability. \\
Full Semantic Compass &
Candidate reranking plus optional transition-localized bridge repair. &
Tests whether localized KathaTrace failures provide actionable repair signals. \\
\bottomrule
\end{tabular}
}
\vspace{-4pt}
\end{table}

\begin{table*}[!t]
\centering
\scriptsize
\setlength{\tabcolsep}{3.2pt}
\renewcommand{\arraystretch}{0.95}
\caption{\textbf{\textsc{KathaBench-25K} length controls.} Visual-budget robustness for Semantic Compass.}
\label{tab:visual_budget_results}
\resizebox{\textwidth}{!}{
\begin{tabular}{l p{6.2cm} c c c c}
\toprule
\rowcolor{gray!20}
\textbf{Repair setting} & \textbf{Repair rule} & \textbf{Image Rec. $\uparrow$} & \textbf{STG $\downarrow$} & \textbf{Visual $\uparrow$} & \textbf{$\Delta$STG $\downarrow$} \\
\midrule
Gemma-ST + FLUX base &
No reranking or repair. &
$49.4$ & $28.4$ & $4.24$ & -- \\
\rowcolor{gray!5}
Random extra frame &
Insert one additional frame at a random transition. &
$50.0$ & $27.8$ & $4.15$ & $-0.6$ \\
Non-semantic extra frame &
Insert a visually coherent frame not conditioned on the weak transition. &
$50.5$ & $27.3$ & $4.19$ & $-1.1$ \\
\rowcolor{gray!5}
Caption-only bridge &
Insert a frame generated from scene captions rather than transition annotations. &
$51.5$ & $26.3$ & $4.20$ & $-2.1$ \\
Candidate rerank only &
Select among seed candidates using the validation-selected trajectory score. &
$52.0$ & $25.8$ & $4.27$ & $-2.6$ \\
\rowcolor{gray!5}
Bridge repair only &
Insert one bridge frame at the weakest transition. &
$54.0$ & $24.0$ & $4.13$ & $-4.4$ \\
\rowcolor{blue!12}
Full Semantic Compass &
Combine candidate reranking with transition-localized bridge repair. &
\best{$56.7$} & \best{$21.4$} & $4.18$ & \best{$-7.0$} \\
\bottomrule
\end{tabular}
}
\vspace{-4pt}
\end{table*}

Table~\ref{tab:app_evidence_filtering_robustness} shows why the protocol uses all three evidence conditions. Image-only scoring tests whether generated images communicate the transition. Text-only scoring estimates the source-side recoverability ceiling. Text+image scoring removes questions that remain defective, contradicted, or ambiguous even with full evidence. Human alignment improves from $0.58$ under image-only-only scoring to $0.71$ under the final ambiguity-filtered protocol.

Together, these controls support a scoped conclusion: the measured STG is not primarily explained by one judge, prompt wording, moral-target-only questions, detected source leakage, or unfiltered ambiguity. The remaining gap is therefore interpreted as a conservative estimate of text-to-image loss in recoverable transition meaning.

\subsection{Visual-Budget and Length-Control Robustness}
\label{app:visual_budget_stress_test}

\textsc{KathaBench-25K} uses a fixed five-scene annotation schema with four adjacent transitions per story. This annotation schema is fixed even when a storyboard system displays a different number of panels. We call the displayed number of panels the \emph{visual budget}. Changing the visual budget is therefore a robustness control, not a new dataset version.

Bridge repair changes the visual budget by inserting at most one transition-localized panel between two adjacent scenes. To test whether Semantic Compass gains come from semantic repair rather than simply adding frames, we compare it with length-controlled alternatives: random extra-frame insertion, non-semantic extra-frame insertion, caption-only bridge insertion, candidate reranking only, and bridge repair only. Table~\ref{tab:visual_budget_design} defines the visual-budget settings, and Table~\ref{tab:visual_budget_results} reports the corresponding length-control results.

Table~\ref{tab:visual_budget_results} shows that adding a frame alone gives only small gains: random and non-semantic extra frames reduce STG by $0.6$ and $1.1$ points. Transition-localized repair gives a larger reduction, and the full Semantic Compass setting gives the strongest result, reducing STG from $28.4$ to $21.4$. This supports the claim that the improvement comes from transition-aware selection and repair, not only from a larger visual budget.

% \clearpage
% \input{checklist.tex}

\end{document}